\documentclass[11pt,colorinlistoftodos]{article}

\usepackage[margin=1in]{geometry}

\usepackage[utf8]{inputenc}
\usepackage[english]{babel}
\usepackage{amsfonts}
\usepackage{amssymb}
\usepackage{enumitem}
\usepackage{dsfont}
\usepackage{bbm}
\usepackage{mathtools}
\usepackage{comment}
\usepackage{xcolor}
\usepackage{caption}
\usepackage{subcaption}
\usepackage{booktabs}	
\usepackage{setspace}
\usepackage{graphicx}
\usepackage[section]{placeins}
\usepackage{fullpage}

\usepackage{soul}
\usepackage[backref=page,breaklinks=true,hidelinks]{hyperref}
\hypersetup{
    colorlinks,
    linkcolor={red!50!black},
    citecolor={blue!50!black},
    urlcolor={blue!80!black}
}
\renewcommand*{\backref}[1]{}
\renewcommand*{\backrefalt}[4]{%
    \ifcase #1 {\footnotesize(Not cited.)}%
    \or        {\footnotesize(Cited on page~#2.)}%
    \else      {\footnotesize(Cited on pages~#2.)}%
    \fi}
\usepackage{amsthm}
\usepackage{thmtools}
\usepackage{natbib}
\usepackage{todonotes}
\usepackage{mleftright}
\usepackage{xfrac}
\usepackage{tikz}
\usepackage{pgfplots}
\pgfplotsset{compat=newest}
\usepgfplotslibrary{fillbetween}
\usepackage[font=small]{caption}

\graphicspath{{Graphs/}}

\presetkeys%
    {todonotes}%
    {inline,backgroundcolor=yellow!20!white}{}

\usepackage[pagewise]{lineno}

\usepackage{version}

\newcommand{\citecomplementarity}{%
    \cite{mclaughlin2024designing}%
}

\newtheorem{asm}{Assumption}
\newtheorem{rem}{Remark}
\newtheorem{prop}{Proposition}
\newtheorem{exm}{Example}

\newcommand{\E}{\mathbb{E}}
\newcommand{\R}{\mathbb{R}}

\let\P\relax
\DeclareMathOperator{\P}{P}
\let\E\relax
\DeclareMathOperator{\E}{E}
\DeclarePairedDelimiter{\Ind}{\mathbbm{1}\lparen}{\rparen}

\renewcommand{\d}{\textnormal{d}}

\newcommand{\safe}{\textnormal{safe}}
\newcommand{\risky}{\textnormal{risky}}
\newcommand{\withheld}{\textnormal{withheld}}
\newcommand{\neutral}{\textnormal{neutral}}
\newcommand{\good}{\textnormal{good}}
\newcommand{\bad}{\textnormal{bad}}
\newcommand{\high}{\uparrow}
\newcommand{\low}{\downarrow}
\newcommand{\rec}{f}
\DeclareMathOperator*{\argmin}{arg\,min}

\newcommand{\cobs}{c_{\text{obs}}}
\newcommand{\chum}{c_h}

\makeatletter
\def\blfootnote{\xdef\@thefnmark{}\@footnotetext}
\makeatother
  
\excludeversion{blind}
\includeversion{nonblind}

    \title{Algorithmic Assistance \\ with Recommendation-Dependent Preferences}
    \begin{nonblind}
    \author{
    Bryce McLaughlin \\ \normalsize Wharton \and Jann Spiess \\ \normalsize Stanford GSB
    }
    \end{nonblind}

    \date{
    First version: August 2022 \\
    This version: October 2025
    }

\begin{document}
    \renewcommand{\sectionautorefname}{Section}
    \renewcommand{\subsectionautorefname}{Section}
    \renewcommand{\footnoteautorefname}{Footnote}

    \maketitle

    \begin{nonblind}
    \blfootnote{
        Bryce McLaughlin (\href{mailto:brycemcl@wharton.upenn.edu}{brycemcl@wharton.upenn.edu}), Wharton School, University of Pennsylvania; Jann Spiess (\href{mailto:jspiess@stanford.edu}{jspiess@stanford.edu}), Graduate School of Business, Stanford University.
        We thank Talia Gillis, Asa Palley, Clare Snyder, and audience members at the 2022 MSOM Conference, the 2022 INFORMS Annual Meeting, the 2023 EC conference, UW--Madison, Stanford, and Columbia for helpful comments and suggestions.
        This manuscript supersedes an earlier version that was accepted and presented at EC'23, with an extended abstract published as:
        McLaughlin, Bryce and Jann Spiess (2023). Algorithmic Assistance with Recommendation-Dependent Preferences. In \emph{Proceedings of the 24th ACM Conference on Economics and Computation (EC'23)}, page 991.
    }
    \end{nonblind}

    \begin{blind}
    \blfootnote{
        This manuscript supersedes an earlier version that was accepted and presented at EC'23, with an extended abstract published as:
        Anonymous (2023). Algorithmic Assistance with Recommendation-Dependent Preferences. In \emph{Proceedings of the 24th ACM Conference on Economics and Computation (EC'23)}, page 991.
    }
    \end{blind}

    \begin{abstract}
        When an algorithm provides risk assessments, we typically think of them as helpful inputs to human decisions, such as when risk scores are presented to judges or doctors. However, a decision-maker may react not only to the information provided by the algorithm. The decision-maker may also view the algorithmic recommendation as a default action, making it costly for them to deviate, such as when a judge is reluctant to overrule a high-risk assessment for a defendant or a doctor fears the consequences of deviating from recommended procedures. To address such unintended consequences of algorithmic assistance, we propose a model of joint human--machine decision-making. Within this model, we consider the effect and design of algorithmic recommendations when they affect choices not just by shifting beliefs, but also by altering preferences. We motivate this assumption from institutional factors, such as a desire to avoid audits, as well as from well-established models in behavioral science that predict loss aversion relative to a reference point. We show that recommendation-dependent preferences create inefficiencies where the decision-maker is overly responsive to the recommendation. As a remedy, we discuss algorithms that strategically withhold recommendations and show how they can improve the quality of final decisions. Concretely, we prove that an intuitive algorithm achieves minimax optimality by sending recommendations only when it is confident that their implementation would improve over an unassisted baseline decision.
    \end{abstract}
    
    \newpage

    \section{Introduction}
    \label{sec:Introduction}

    One important application of algorithms is to turn complex data into simple predictions or recommendations that help decision-makers make better choices, such as risk assessments presented to judges or doctors.
    We typically think of such algorithmic assessments as providing additional information about which choices will lead to better outcomes.
    Yet decision-makers may react to algorithmic input not just by shifting beliefs, but also by changing their preferences.
    In this article, we consider the effect and design of algorithmic advice when it also imposes a cost on the decision-maker whenever they deviate from the recommended action, such as when a judge is reluctant to release a defendant determined to be at risk for re-offense by a criminal risk assessment or
    a doctor fears the consequences of not testing a patient with a high predicted risk of a specific medical condition.
    We show that recommendation dependence creates inefficiencies where the decision-maker is overly responsive to the recommendation, and propose changes to the design of recommendation algorithms.
    
    We model the interaction of a decision-maker with a recommendation algorithm in a principal--agent model of joint human--machine decision-making.
    The principal designs a recommendation algorithm.
    The agent plays the role of the human decision-maker, and chooses between a safe and a risky action based on their private information along with a recommendation provided by the algorithm.
    When the state of the world is good, the risky action is best, while in the bad state, the risky action leads to high loss.
    The agent uses the information available to them to assess the probability that the state is bad, and chooses the risky action only if that predicted probability is low.
    For example, a judge who considers whether to release a defendant on bail (risky action) aims to release only those defendants with low probability of failing to appear or committing a new crime (bad state).
    
    To this model, we add the assumption that recommendations affect decisions not only through the information they provide, but also by setting a reference point against which the agent measures their outcomes.
    We assume that the agent perceives an additional (personal) cost from making an error when deviating from the algorithm's recommendation. 
    Specifically, in our model, there is an additional loss when the agent takes a risky decision against the safe recommendation of the algorithm and the bad state materializes.
    Similarly, there may also be an additional loss from deviating when the agent opts for the safe option relative to a risky recommendation and the good state occurs.
        
    A first motivation for such recommendation-dependent preferences stems from institutional factors, such as when making mistakes that defy recommendations triggers audits or may create backlash.
    For example, a judge may be reluctant to release a defendant in light of a jailing recommendation for fear of repercussions, even if they believe that the defendant represents a lower risk. Similarly, a doctor may prefer to order a test (safe decision) when the algorithm recommends doing so for fear of missing a bad diagnosis against algorithmic advice, leading to an accusation of malpractice.
    We provide one specific model that microfounds such targeted audits. Specifically, we show that it may be optimal for an outside observer to mostly target audits towards mistakes that go against algorithmic recommendations, since they are most indicative of avoidable mistakes.
        
    A second motivation is provided by established models from behavioral science that suggest expected losses impact decision-makers more than commensurate gains, relative to some reference point that we assume here is affected by the algorithmic recommendation. 
    The combination of perceiving decision utility or regret relative to a reference point and experiencing loss aversion against that reference point predicts recommendation dependence when we assume that the reference point is obtained from the recommended action.
    In addition, recommendation dependence also arises naturally from a model of costly defaults, where overriding a default is costly.
    
    Having set up a model of recommendation-dependent preferences, we show that the effect of algorithmic advice generally differs from a reference-independent baseline case.
    Recommendation dependence increases adherence to the algorithmic recommendation.
    This adherence makes decisions less efficient as it reduces the amount of private information that the agent reveals through their chosen action.
    For example, if a judge is worried about repercussions from releasing a defendant that the algorithm recommends jailing, the judge may follow the recommendation even if they have private information that suggests the defendant is not at high risk of committing a new crime or failing to appear.

    Recommendation dependence leads to inefficiencies that can be mitigated (but not completely avoided) by improved recommendation design.
    We first tackle the case where the algorithm explicitly recommends courses of action.
    In this case, we show that recommendation dependence generally changes the optimal recommendation algorithm.
    Optimal algorithmic design in a world of recommendation dependence now balances two forces: On the one hand, a good recommendation algorithm should supply the decision-maker with a maximal amount of information.
    On the other hand, the algorithm should also minimize the distortion that comes from its recommendation.
    For example, if the agent is reluctant to overrule a safe recommendation because of additional costs from making a mistake in this case, then the optimal algorithm may reduce distortions by recommending the safe option less, leading to an optimal algorithm that instead proposes the risky option in some cases where a baseline algorithm would recommend the safe action.
    
    Having shown how recommendation dependence affects the consequences and optimal design of recommendations, we discuss the benefits of allowing the algorithm to give a neutral ``don't know'' recommendation when the algorithm is unsure of the best decision.
    With recommendation-dependent preferences, adding a third option of not providing a recommendation at all has two distinct benefits.
    The first benefit is that it allows the transmission of additional information through the recommendation, signaling an intermediate probability of a bad outcome occurring.
    The second benefit is that not providing a recommendation in some cases also reduces the cost of recommendation dependence, and allows the agent to make optimal decisions in this case.
    Specifically, we show in a simple example that adding such an additional ``don't know'' level within our model can improve decisions relatively more in a world with recommendation-dependent preferences than in a world where the agent's preferences are not affected by the algorithm.

    We then provide prescriptions for training recommendation algorithms in practice. Our above results assume that the designer of the algorithm knows the joint distribution of the outcome with the information available to the algorithm and to the human decision-maker. Yet in practice, only limited baseline information may be available when an algorithm is trained. We therefore consider the case where the designer only observes training data with human decisions that are unassisted by recommendations. In this world, we derive minimax optimal recommendation algorithms that make worst-case assumptions about the private information and preferences of the human decision-maker. We show that a simple triage algorithm guarantees human--algorithm complementarity in this case. This triage algorithm sends recommendations only in the case when their implementation is guaranteed to improve over the human's unassisted baseline actions, and withholds recommendations otherwise. In addition, we show how this idea extends to the case where not all outcomes are observed in the training data, such as in the case when we only learn for defendants who are released from jail whether they commit a new crime or fail to appear.

    We extend our results to algorithms that present a risk score to the human decision-maker, such as when a doctor receives the predicted probability that a patient suffers from a specific condition.
    In this case, recommendation dependence may lead to overadherence to the action that is implicitly suggested by the risk assessment.
    We consider an algorithm that strategically withholds risk scores, and show how such strategic silence can improve overall outcomes.
    While withholding risk assessments destroys valuable information,
    we argue that creating instances without recommendations also reduces distortions.
    For example, a judge may make better decisions in borderline 
    cases if the algorithm strategically withholds uninformative risk assessments and thereby ensures that the human decision-maker uses their private information efficiently.

    We contribute to a cross-disciplinary literature that studies human--AI interaction.
    This includes work where the knowledge of an AI and human decision-makers (or more generally multiple knowledge sources) are combined \citep[e.g.][]{lawrence_judgmental_2006,palley_extracting_2019,steyvers_bayesian_2022,peng_no_2024}, where humans assist an AI \citep[e.g.][]{hampshire_beyond_2020,ibrahim_eliciting_2021,alur_human_2024},
    where an algorithm optimizes advice given to human decision-makers \citep[][]{bastani2021improving,sun_predicting_2022,agarwal_designing_2025,hoong_improving_2025}, when to give advice \citep{noti_learning_2022}, or which instances to delegate \citep{raghu_algorithmic_2019,mozannar_consistent_2021,bondi_role_2022}.
    \cite{hemmer_human-ai_2021, lai_towards_2021,vaccaro_when_2024} provide reviews of the literature on complementarity in human--AI systems.
    Recent contributions to this literature emphasize that the success of human--machine collaboration is dependent on details of context, implementation, and presentation of algorithmic advice \citep[such as][]{bansal_updates_2019,green2019principles,snyder_algorithm_2022,mclaughlin2024designing}, including information about its uncertainty \citep{mcgrath_when_2020,taudien_effect_2022,vodrahalli_uncalibrated_2022}, explanations of black-box classifiers \citep{LakkarajuB20}, or the set of options suggested in a conformal prediction setting \citep{straitouri_improving_2023,toni_towards_2024}.
    Relative to this literature, we make three distinct contributions:
    First, we bring in ideas from behavioral science to explicitly model how algorithms may impact decisions beyond the information they provide.
    Second, we motivate the importance of algorithms strategically withholding recommendations to enable effective human--algorithm collaboration.
    Finally, we show that simple triage-type algorithms present a feasible and effective way of implementing recommendations with recommendation-dependent decision-makers in practice.
    
    The remaining article is organized as follows.
    In \autoref{sec:Model}, we formalize the concept of recommendation-dependent preferences within a model of algorithm-assisted human decisions, for which we provide some micro-foundations in \autoref{sec:Foundations}.
    In \autoref{sec:Implications}, we describe how this recommendation dependence introduces inefficiencies and affects the design of optimal recommendation algorithms.
    As a remedy, we discuss the value of strategically withholding recommendations in \autoref{sec:NonRec}.
    We propose a feasible implementation of effective recommendation algorithms from limited data in \autoref{sec:Implementation}.
    In \autoref{sec:Implicit}, we consider a version of our model where the decision-maker's information also includes the machine's risk prediction,
    before concluding in \autoref{sec:Conclusion}.
    
    \section{A Model of Recommendation-Dependent Preferences}
    \label{sec:Model}

    We model the interaction of a human decision-maker with a recommendation algorithm as a game between an algorithm designer (principal) and the decision-maker (agent).
    The principal designs an algorithm that provides the agent with a recommendation $R$, such as a suggested course of action or a risk score.
    The agent leverages this recommendation to make a decision $A$ about an instance with outcome $Y$.
    Throughout, we focus on the case where outcomes and actions are binary, with outcomes $Y \in \{ \good,\bad\}$ and actions $A \in \{\safe,\risky\}$.
    Principal and agent both want to take the safe decision when faced with a bad outcome, but prefer the risky decision when the outcome is good.
    For example, the agent may be a judge who decides whether to release ($A = \risky$) or jail ($A = \safe$) a defendant, where the defendant, if released on bail, may turn out to commit an offense or fail to appear ($Y = \bad$), or may appear without any new criminal activity ($Y = \good$).
    
    We assume that the agent and the algorithm have access to features $X_h$ and $X_m$, respectively.
    The two signals may be correlated and contain joint information about the instance at hand that encodes any commonly known context and information about the distribution of $Y$.
    In addition, the signal $X_h$ of the human decision-maker may also include details not available to the machine, such as properties of the specific instance only visible in-person, and the machine's signal $X_m$ may likewise encode information not directly accessible to the human decision-maker, such as information deduced from administrative records or large training data.
    
    Jointly, the outcome $Y$ and the signals $X_h$ and $X_m$ follow a distribution $\P$.
    We assume that this joint distribution is common knowledge between the algorithm designer and the human decision-maker, but that the realization of $X_h$ is only observed by the agent and the realization of $X_m$ is only observed by the principal.%
    \footnote{Since the principal never observes $X_h$ and the agent never sees all of $X_m$, it may be unrealistic that the joint distribution $\P$ is fully known to both. 
    In \autoref{sec:Implementation}, we therefore consider second-best solutions when the knowledge of $\P$ is incomplete and has to be inferred from limited training data.
    }
    In the judge example, the distribution $\P$ is over whether the defendant will fail to appear or commit a new crime if released, together with the information the judge and the algorithm have about a defendant.

    The game between the designer of the algorithm (principal) and the human decision-maker (agent) plays out as follows:
    \begin{enumerate}
        \item The algorithm designer (principal) chooses a recommendation algorithm $\rec: \mathcal{X}_m \rightarrow \mathcal{R}$ that maps the machine information $X_m \in \mathcal{X}_m$ to a recommendation $R = \rec(X_m) \in \mathcal{R}$, for a set of available recommendations $\mathcal{R}$.
        \item The outcome and features $(Y,X_h,X_m) \in \{ \good,\bad\} \times \mathcal{X}_h \times \mathcal{X}_m$ are drawn from $\P$.
        \item The human decision-maker (agent) observes the recommendation algorithm $\rec$, the feature $X_h$, and the machine recommendation $R = \rec(X_m)$, then takes a decision $A \in \{ \risky,\safe\}$.%
        \footnote{
        \label{ftn:Agentinfo}
        While our model assumes that the decision-maker has knowledge of the full distribution $\P$, it is sufficient to assume that they only know the distribution of $(Y,X_h,R)$, and not also of the machine information $X_m$. This is because the agent takes the recommendation policy $\rec$ as given and takes a decision given $R = \rec(X_m)$ and $X_h$, for which the distribution of $Y|X_h,R$ is sufficient. The agent could therefore learn all relevant information over time from observing draws $(Y,X_h,R)$.}
        \item The outcome $Y \in \{ \good,\bad\}$ and losses are realized.
    \end{enumerate}
    For example, 
    the designer of the algorithm in the judge example may choose a mapping from the information available to the machine to a recommendation of which action to take ($\mathcal{R} = \{\risky, \safe\}$), which the judge then observes together with the additional information the judge learns from the defendant in the courtroom before deciding on whether to jail or release.

    We assume that the principal aims to minimize expected loss (risk) $\E[\ell(Y,A)]$ for some loss function $\ell$.
    As a crucial deviation from standard (rational) models of human decision-making, we assume that the agent anticipates a decision loss $\ell^*(Y,A,R)$ that deviates from the consequence of the action alone, and can depend on the recommendation.
    We then assume that the agent minimizes expected loss (risk) $\E[\ell^*(Y,A,R)]$, taking the recommendation algorithm $\rec$ as given.%
    \footnote{We assume that principal and agent act \emph{as if} they minimize expected loss according to these loss functions and the distribution $\P$, irrespective of whether the outcomes $Y$ end up being observed.}
    We say that this loss function expresses recommendation-dependent preferences as choices are now evaluated both by their consequences and relative to the given recommendation.
    
    We assume that the principal (strictly) prefers the risky action when the outcome is good and the safe action when the outcome is bad. Hence, there are some $c_I, c_{II} > 0$ so that we can, without loss of generality, write the principal's loss function as 
    \begin{equation}
        \label{eqn:Loss}
        \ell(y,a) =
        \begin{cases}
            c_I, & y=\good, a=\safe, \\
            c_{II}, & y=\bad, a=\risky.
        \end{cases}
        \end{equation}
    The two cases in the principal's loss function cover the two mistakes of choosing the safe option despite the outcome being good (leading to $c_{I} > 0$, type-I error) or the risky decision in a bad case (leading to $c_{II} > 0$, type-II error).
    For the judge's decision, $c_I$ is the cost of jailing a defendant who would not engage in criminal activity, and $c_{II}$ the cost of releasing a defendant who commits a new crime or fails to re-appear.%
    \footnote{Note that loss functions of this type are not typically unique, since only relative costs matter; for example, we could normalize $c_I = 1$ without loss of generality.}

    For most of our article, we assume that the agent experiences the same loss $\ell(Y,A)$ from how action and outcome align as the principal, and additional loss from any mistake made against the recommendation of the algorithm.
    As a baseline, we consider recommendations that are binary and correspond to suggested actions ($\mathcal{R} = \{\risky, \safe\}$). 
    We assume that agent preferences are
     \begin{equation}
        \label{eqn:Recdeppref}
        \ell^*(y,a,r) =
        \ell(y,a)
        +
        \begin{cases}
            \Delta_I, & y=\good, a=\safe, r=\risky, \\
            \Delta_{II}, & y=\bad, a=\risky, r=\safe,
        \end{cases}
    \end{equation}
    Here, $\Delta_I \geq 0$ describes the \textit{additional} loss of the human decision-maker when they play it safe against the machine's recommendation of a risky action, and the risky action would have been optimal.
    Similarly, $\Delta_{II} \geq 0$ quantifies the penalty of taking a risky decision in the bad state when the algorithm recommends the safe action, such as when the judge gets in trouble for releasing a defendant against the recommendation of the algorithm, who then goes on to commit a crime.
    \autoref{tab:losses} summarizes the resulting losses in Panel~(b), and compares them directly to the principal's losses in Panel~(a), which depend on recommendations only through final decisions.
    We discuss formal justifications for this form of losses in \autoref{sec:Foundations}.

    \begin{table}[ht]
        \centering
        \begin{subtable}{.35\textwidth}
        \centering
        \begin{tabular}{lrcc}
         &  &  &  \\
         \toprule
         \multicolumn{2}{l}{Decision} & $\safe$ & $\risky$ \\
        \midrule
        Outcome & $\good$ & $c_I$ & 0 \\
        & $\bad$ & 0 & $c_{II}$ \\
        \bottomrule
        \end{tabular}
        \caption{Welfare losses of the principal}
        \label{tab:losses_principal}
        \end{subtable}%
        \hspace{1cm}%
        \begin{subtable}{.55\textwidth}
        \centering
        \begin{tabular}{lrcccc}
        \toprule
        \multicolumn{2}{l}{Recommendation} & \multicolumn{2}{c}{$\safe$} & \multicolumn{2}{c}{$\risky$} \\
        \cmidrule(lr){3-4}
        \cmidrule(lr){5-6}
        \multicolumn{2}{l}{Decision} & $\safe$ & $\risky$ & $\safe$ & $\risky$ \\
        \midrule
        Outcome & $\good$ & $c_I$ & 0 & $c_I{+}\Delta_I$ & 0 \\
         & $\bad$ & 0 & $c_{II}{+}\Delta_{II}$ & 0 & $c_{II}$ \\
        \bottomrule
        \end{tabular}
        \caption{Decision losses of the agent}
        \label{tab:losses_agent}
        \end{subtable}
        \caption{Losses of principal (left) and agent (right) as a function of the realized outcome $Y \in \{\good,\bad\}$, algorithmic recommendation $R \in \{\safe,\risky\}$, and decision $A \in \{\safe,\risky\}$.}
        \label{tab:losses}
        \end{table}

    In our model, four frictions keep the principal from implementing the first-best action.
    First, the principal only has access to the machine features $X_m$ and not to the human features $X_h$, which creates complementarities that motivate collaborative human--machine decision-making.%
    \footnote{Here, we take it as given that the human decision-maker retains the final decision authority, but the complementarity in information also motivates why the principal may want to delegate the decision to the human decision-maker in the first place.}
    Second, the principal is limited to sending information via simple recommendations from the set $\mathcal{R}$.
    Third, there is partial misalignment between principal and agent due to recommendation dependence.
    And fourth, the principal cannot make transfers and can only influence the agent's decision via the recommendation policy.
    This setup is similar to the sender--receiver game in \cite{kamenica_bayesian_2011} with respect to the last point, but differs since communication is limited to simple recommendations and misalignment is affected by the recommendations themselves.

    \section{Sources of Recommendation Dependence}
    \label{sec:Foundations}
    
    In the previous section, we proposed a model of recommendation-dependent preferences where recommendations distort the trade-offs between choices available to the agent.
    In this section, we provide three examples that yield such recommendation-dependent preferences: first, a derivation from loss aversion or regret relative to a reference point as in behavioral science; second, a foundation in terms of default actions that are costly to override; and third, a model of costly oversight by an observer.
    In addition, we collect additional examples and evidence for recommendation dependence that do not neatly fall into those three categories.

\subsection{Recommendations as Reference Points for Loss Aversion}
\label{subsec:Refdep}

    Behavioral science has developed models of decision-making that express the idea that humans are sensitive to changes in their utility relative to a reference level, particularly a decrease in their utility. In Prospect Theory \citep{kahneman1979prospect,barberis2013thirty}, human decision-makers evaluate their outcome against some reference point and factor the gain or loss relative to this reference into their decision-making process. Similarly, in Regret Theory \citep{bell_regret_1982,loomes_regret_1982,diecidue_regret_2017}, decision-makers evaluate their outcome relative to counterfactual outcomes which would have been observed under alternate courses of action, and factor expected regret into their decision-making process.
    Both frameworks allow the decision-maker to experience losses (regrets) against a reference outcome more than gains (rejoicing). As a result, changing this reference will change the decision-maker's preferences. If the reference depends on the recommendation, the decision-maker's preferences will be recommendation-dependent.

    As an example, we now consider loss aversion relative to a reference point affected by the recommended action, and show how it directly leads to recommendation dependence.
    First, we assume that choices are \emph{evaluated relative to a reference loss}, which we here assume is influenced by the action $R$ recommended by the algorithm. 
    In \autoref{apx:gainloss}, we consider reference points that come from the expected loss of implementing the recommended action or a lottery of the implied losses, in line with gain--loss utility in \cite{koszegi_model_2006}.
    Here, we focus on the case where the reference outcome is the counterfactual loss achieved by the recommendation, that is $\ell(Y,R)$. This example connects our approach to a version of Regret Theory in which the only counterfactual loss considered in the reference is the action suggested by the recommendation. 
    Second, we assume the decision-maker puts \emph{more emphasis on losses} relative to the reference point $\ell(Y,R)$ than on gains.
    Specifically, we assume that loss aversion takes the form of a factor $\lambda > 1$ by which losses are multiplied.
    This means that decision loss from taking action $A$ given recommendation $R$ and observing outcome $Y$ is given by
    \[
        \ell^{\text{LA}}(Y,A,R)
        = \lambda [\ell(Y,A) - \ell(Y,R)]_+ -  \: [\ell(Y,A) - \ell(Y,R)]_-,
    \]
    where by $[\cdot]_+$ and $[\cdot]_-$ we denote the (absolute value of the) positive and negative parts, respectively.
    This loss is equivalent to  recommendation-dependent decision loss from \eqref{eqn:Recdeppref}:
    
    \begin{prop}[Derivation from loss aversion]
    \label{prop:PT}
    Decision-maker choices according to $\ell^{\text{LA}}$ with $\ell_0(Y,R,U) = \ell(Y,R)$ are equivalent to choices according to $\ell^*$ with $\Delta_I = (\lambda - 1) c_I, \Delta_{II} = (\lambda - 1) c_{II}$.
    \end{prop}

    This form of reference dependence represents only one specific choice of modeling the idea that the recommendation becomes a reference point. In \autoref{prop:refdep} in the appendix, we show that a similar result applies when we instead consider reference points set by the expected loss from implementing the recommendation or by a lottery over type-I and type-II losses. Alternatively, we could consider a personal equilibrium that correctly anticipates the distribution of human decisions following a specific recommendation, as in the model in \cite{koszegi_model_2006} for the endogenous formation of reference points.

    Some recent studies have observed humans responding to algorithmic assistance in line with reference effects. \cite{albright2023hidden} finds that the introduction of an explicit release recommendation to an already existing assessment greatly increased releases without adding any additional information.
    \cite{fogliato_who_2022} finds that when an algorithm is introduced earlier in a decision-making process, humans adhere to it more, reducing accuracy.
    We interpret these findings as supporting the idea that algorithms set reference points that human decision-makers anchor on.

    We have so far considered the case where the recommendation becomes the reference point. However, what is considered the reference point for human decisions may be affected by design decisions beyond our model. For example, the saliency of different alternatives and the attention a user pays to them may affect the reference point \citep{bhatia_attention_2019,kibris_theory_2023}. \cite{baucells_reference-point_2011} provides evidence for this theory in a financial context, noting the out-sized effects of the first price an agent saw for an asset and the most recent price on that agent's current willingness to pay.
    The aforementioned \cite{fogliato_who_2022} varies the order in which recommendations are provided to the human decision-maker. This study documents that human decision-makers appear to over-anchor in the machine recommendation when it is given first, leading to inefficient decisions that discount human information. If, however, the human decision-maker is first forced to announce their planned action before the machine recommendation becomes available, then decisions following the machine recommendations exhibit less anchoring effects.
    This finding provides some nuance to our model assumptions: If machine recommendations are provided to the human decision-maker first, they appear to act as reference points and induce anchoring effects. If, however, an alternative action (such as the human plan) becomes the reference point, then our model may not apply anymore as stated, and the principal would instead have to anticipate reference dependence relative to this alternative reference point. This points to interesting design choices outside of the scope of our current model, namely the sequencing, framing, and highlighting of information provision and decisions in efficient human--algorithm collaboration.

    \subsection{Default Actions and Costly Deviation}
    \label{subsec:Default}

    Rather than merely influencing psychological reference points, recommendations may also represent defaults that are expensive to change. For example, saving defaults may be sticky, and are therefore an important policy choice \citep[e.g.][]{choi2004better}. We capture the idea of defaults by assuming that $\mathcal{R} = \{\risky,\safe\}$, and that keeping the given recommendation does not incur any additional loss, while overriding the recommendation%
    \footnote{Alternatively, there could be separate costs of deviating from the risky versus the safe recommendation.}
    comes at a cost $c$:
    \begin{equation}
    \label{eqn:Defaultpref}
        \ell^{\text{default}}(y,a,r) = \ell(y,a) + \Ind{a \neq r} \: c
    \end{equation}
    Then this additional cost implies recommendation-dependent preferences:
    
    \begin{prop}[Costly defaults as recommendation-dependent preferences]
    \label{prop:Costlydefault}
        Assume that $c < \min\{c_I,c_{II}\}$.
        Then the loss function in \eqref{eqn:Defaultpref} is equivalent to a recommendation-dependent loss function of the form \eqref{eqn:Recdeppref}
        with $\Delta_I = \frac{c (c_I + c_{II}) }{c_{II} - c}, \Delta_{II} = \frac{c (c_I + c_{II}) }{c_{I} - c}$.
    \end{prop}

    In \autoref{sec:Implications}, we show how recommendation dependence can lead to inefficient over-adherence to recommendations. This discussion mirrors recent scientific debates and empirical evidence on default nudges in personal finance. While the previous literature had documented benefits to setting higher default savings and repayment rates, recent results suggest that over-adherence to these defaults may inefficiently increase credit card debt and thus have net-negative welfare effects for some consumers \citep[e.g.][]{guttman2023semblance}. This provides an application for better personalized recommendation algorithms that take recommendation dependence into account, such as those we propose in \autoref{sec:Implementation}.

    \subsection{Oversight, Culpability, and Blame}
    \label{subsec:Blamegame}

    The above micro-foundations assume that the individual experiences exogenous costs when deviating from recommendations, either because of loss aversion relative to a reference point or because deviating from defaults is perceived to be costly. In this section, we instead ask whether it is ever optimal to design oversight mechanisms that target decision-makers based on mistakes that go against recommended actions.
    In particular, we present one specific model of targeted audits that leads to recommendation-dependent preferences, and provide sufficient conditions under which they take the simple form from \eqref{eqn:Recdeppref}.

    We formalize the implications of outside scrutiny and assigning blame as a game with a relatively uninformed third-party observer. Specifically, for the case of binary recommendations $\mathcal{R} = \{\risky,\safe\}$, we assume that this outside observer does not know the full distribution $\P$ and only gets access to the recommendation $R$, the realized outcome $Y$, and the human action $A$. The observer assumes that the action $A$ is taken either by an attentive type, $\theta = \text{attentive}$, or by a negligent type, $\theta = \text{negligent}$\footnote{The existence of negligent agents is supported e.g. by \cite{angelova_algorithmic_2023}, who find many judges mistakenly deviate from recommendations in response to spurious noise from other cases.}. The attentive type takes optimal decisions that minimize expected loss, but the exact values $c_I,c_{II},\chum$ of the private costs of the attentive agent are unknown to the observer. The negligent type takes uniformly random decisions.\footnote{Alternatively, we could assume that the negligent type has a probability of taking optimal decisions, but we simplify the exposition by making this probability zero.}
    The observer can audit individual cases after observing the realized $(Y,R,A)$, at a cost of $\cobs \in (0,1)$ to the observer and a cost of $\chum < \min\{c_I,c_{II}\}$ to the human decision-maker. The audit reveals the type $\theta$ of the specific agent; if the agent is found to be of the negligent type, then the observer receives a benefit normalized to 1.%
    \footnote{Since we are only interested in implications for the attentive type, we do not model the negligent type's preferences or punishment. While the attentive type is never found to be negligent, they are still inconvenienced by being audited.}
    We assume that the types themselves are random with $\P(\theta = \text{negligent}) = q \in (0,1)$ and independent of $(Y,X_m,X_h)$, and we focus on the case of a given recommendation policy $\rec$.
    In this game, what is the observer's optimal screening policy $\pi$ with $\pi(Y,R,A) \in \{1, 0\}$ (where 1 corresponds to auditing)?

    Assuming that the agent has a loss function $\ell$ as above, the agent's total loss is $\ell^*(Y,R,A) = \ell(Y,A) + \pi(Y,R,A) \cdot \chum$.
    This loss is generally recommendation-dependent. We now provide sufficient conditions under which the maximin optimal observer policy in the game between observer and agent leads more specifically to a loss function of the form in \eqref{eqn:Recdeppref}.

    \begin{prop}[Maximin optimal audits]
    \label{prop:Minimaxaudits}
        Write $A^*$ for the optimal decision taken by a rational agent with loss $\ell$ after observing the recommendation $R$ and their private information $X_h$.
        We say that the recommendation $R$ is correct for $Y$ when either $R{=}\risky$ and $Y{=}\good$ or $R{=}\safe$ and $Y{=}\bad$.
        We assume that the observer makes only the assumptions that
        \begin{align}
        \label{eqn:Auditminimaxcondition}
            \P(\text{$R$ is correct for $Y$}) &\geq \frac{1}{2}
            &
            \P(A^*=R|\text{$R$ is correct for $Y$}) &\geq \frac{1 + \eta}{2},
        \end{align}
        for some $\eta \in (0,1)$.
        If
        $
            0 < \frac{\cobs - q }{(1-q)\cobs} < \eta,
        $
        then the unique policy that maximizes minimal expected observer utility (where the minimum is taken over all distributions consistent with the observer's assumptions and all $c_I,c_{II}, \chum > 0$) is to audit all mistakes going against the recommendation, that is,
        \(
            \pi(y,r,a) = \Ind{y {=} \bad, a {=} \risky, r {=} \safe} + \Ind{y {=} \good, a {=} \safe, r {=} \risky}.
        \)
        Subject to these audits, the
        attentive agent takes decisions according to the loss function in \eqref{eqn:Recdeppref} with $\Delta_I = \chum = \Delta_{II}$.
    \end{prop}

    The assumptions in \eqref{eqn:Auditminimaxcondition} are quite straightforward: First, the recommendation is more likely to be correct than not.%
    \footnote{This assumption could be seen as a normalization of the recommendation labels: If the recommendation is more likely to be incorrect, it may make sense to swap the labels of the risky vs safe recommendations.}
    Second, a fully rational agent (with loss function $\ell$) is more likely than not to follow correct recommendations. This assumption could be driven either by the agent complying with the recommendation $R$ because it is informative, or the agent having valuable private information about $Y$, or both.%
    \footnote{Indeed, 
        $\P(A^*=R|R) \geq \frac{1 + \eta}{2}, \P(A^* \text{ is correct for }Y|Y) \geq \frac{1 + \eta}{2}$ are each sufficient conditions for this assumption.
    }
    The factor $\eta$ then controls the degree to which decisions are non-trivial.
    Finally, the assumption $0 < \frac{\cobs - q }{(1-q)\cobs} < \eta$ expresses that the cost $\cobs$ of auditing (or the fraction $q$ of inattentive agents) must lie in an intermediate range: large enough (or $q$ small enough) to ensure that auditing additional instances is not generally optimal, but also small enough (or $q$ large enough) relative to the ``decision quality'' $\eta$ to make it worthwhile to audit mistakes that go against the recommendation.

    The intuition behind this result is equally straightforward: If the recommendation is informative about outcomes and the attentive agent is more likely than not to follow the recommendation when it is the right thing to do, then not following the recommendation and making a mistake is a sign of an avoidable mistake by the negligent type.

    Here, we have assumed a fixed recommendation policy. However, the result goes through in the game from \autoref{sec:Model} provided the observer moves first, the designer takes the screening policy as given, and the conditions \eqref{eqn:Auditminimaxcondition} from \autoref{prop:Minimaxaudits} hold for all recommendation algorithms $\rec$ considered by the designer.
    Note here that the designer's objective is unaffected by the presence of the inattentive type, since that type's choices are not changed by the recommendation. 

    This observer game could be adapted in different variants.
    In some cases, only some outcomes are observed. For example, the outside observer may only ever learn about the outcome $Y$ when a risky action $A = \risky$ (such as a release decision) is taken, while $A {=} \safe$ leads to an unobserved outcome.
    In this case, \autoref{prop:Minimaxaudits} simplifies, with the condition only applying to the $A {=} \risky$ action, leading to $\Delta_I {=} 0, \Delta_{II} {=} \chum$.
    As an alternative implementation, the goal of the audit could be to ensure that the agent spends effort to make attentive decisions rather than negligent ones.
    In that case, the algorithm designer and the observer could be the same, since it may be optimal to induce some degree of recommendation dependence in return for attentive decisions.
    
    Many high-stakes contexts use similar reviews in which decision-makers can be held culpable for systematic errors. In medicine, clinicians can lose their license due to medical malpractice if they deviate from the accepted standard of care \citep{bal_introduction_2009}. \cite{froomkin_when_2019} argue that under current interpretations of tort law, algorithmic recommendations will become the standard of care when they exceed the average performance of clinicians. \cite{dai_express:_2025} explore how clinicians' use of AI interacts with liability schemes and show that using AI as the standard of care introduces AI over-use for some patients and under-use for others. In the judicial system, 17 U.S. states currently utilize retention elections where the community votes on whether judges should be reappointed or removed \citep{IAALS_judicial_2022}. Judges admit that they hesitate to make decisions that could result in public backlash and note that risk assessments can be used to legitimize their decisions to the public \citep{esthappan_assessing_2024}.

    \subsection{Further Evidence}

    To conclude our discussion of sources of recommendation-dependent preferences, we examine results from the literature that indicate that human decision-makers respond to recommendations beyond the information they provide.
    Empirical studies evaluating algorithm-assisted decision-making under uncertainty often find that the informative effects of recommendations are dominated by preference-based effects.
    \cite{banker2019algorithm} observes that consumers are willing to adopt recommendations inferior to their own decisions, while \cite{fugener_will_2021} finds that algorithm-assisted wisdom of the crowds underperforms unassisted wisdom of the crowds, indicating information destruction. Both \cite{stevenson_algorithmic_2019} and \cite{Imai2020-ci} find that risk assessments do not reduce incarceration rates or crime, even though judges' decisions changed significantly. One potential explanation for these phenomena comes from \cite{doval_persuasion_2023}, which suggests that judges may be using risk assessments to infer the preferences of the public.

    \section{Implications of Recommendation Dependence}
    \label{sec:Implications}

    Having set up and motivated a model of recommendation-dependent preferences,
    we now discuss how machine recommendations affect human choices beyond their information content.
    We then derive implications for the optimal design and limitations of binary recommendation algorithms.

    A human decision-maker with the same loss function $\ell$ from \eqref{eqn:Loss} as the principal would trade off the costs of type-I and type-II errors, leading to a simple cutoff rule and decisions
    \begin{equation}
    \label{eqn:Optimaldec}
        A^*
        =
        \argmin_{a \in \{ \risky,\safe\}} \E[\ell(Y,a)|X_h, R]
        =
        \begin{cases}
            \risky, & \P(Y{=}\bad|X_h,R) \leq p^* = \frac{c_{I}}{c_{I}+c_{II}},
            \\
            \safe, & \P(Y{=}\bad|X_h,R) > p^*,
        \end{cases}
    \end{equation}
    given the recommendation policy $\rec$ and resulting recommendations $R = \rec(X_m)$ (where we break ties in favor of the $\risky$ decision).
    If the agent has recommendation-dependent preferences, on the other hand, then decisions still follow a cutoff rule, but the cutoffs are now distorted:

    \begin{rem}[Recommendation-dependent thresholds]
        \label{rem:Thresholds}
        Assume that the agent has recommendation-dependent preferences as in \eqref{eqn:Recdeppref}.
        Then the agent's choices given a recommendation policy $\rec: \mathcal{X}_m \rightarrow \mathcal{R}$ and resulting recommendations $R = \rec(X_m)$ are a.s.\ given by 
        \begin{align*}
            A = \begin{cases}
                \risky, & \P(Y{=}\bad|X_h,R) \leq p^R, \\
                \safe, & \P(Y{=}\bad|X_h,R) > p^R,
            \end{cases}
        \end{align*}
        for thresholds $p^\risky = \frac{c_I + \Delta_I}{c_I + c_{II} + \Delta_I} \geq p^* \geq \frac{c_I}{c_I + c_{II} + \Delta_{II}} = p^\safe$, where we assume that ties are broken in favor of the risky action.
    \end{rem}
    
    That is, faced with a recommendation, the agent updates their belief about the outcomes $Y$, and implements the risky action only if the probability of the bad outcome is sufficiently low given the trade-off between type-I and type-II costs for the specific recommendation.

    In a baseline model without recommendation dependence, providing recommendations can only improve agent choices by supplying additional information.
    But in the presence of recommendation dependence, providing recommendations also distorts preferences.
    To show these two forces, we now decompose the net gain or loss from introducing recommendations into an information gain minus possible distortions through recommendation dependence.
    Specifically, we express the change in expected loss relative to the unassisted decision $A_0$ taken by a human with loss function $\ell$,
    \begin{align}
    \label{eqn:A0}
        A_0 = 
        \argmin_{a \in \{ \risky,\safe\}} \E[\ell(Y,a)|X_h]
        =
        \begin{cases}
            \risky, & \P(Y{=}\bad|X_h) \leq p^* = \frac{c_{I}}{c_{I}+c_{II}},
            \\
            \safe, & \P(Y{=}\bad|X_h) > p^*.
        \end{cases}
    \end{align}
    (Note that we assume that these baseline decisions are not affected by recommendation dependence since no recommendations are provided.)
    Our decomposition is then
    \begin{align}
    \label{eqn:Decomposition}
        \textnormal{Net}(\rec) = \E[\ell(Y,A_0) - \ell(Y,A)]
        =
        \underbrace{\E[\ell(Y,A_0) - \ell(Y,A^*)]}_{= \textnormal{IG}(\rec) \geq 0} - \underbrace{\E[\ell(Y,A) - \ell(Y,A^*)]}_{= \textnormal{Dist}(\rec) \geq 0}
\end{align}
    which separates the net impact, $\textnormal{Net}(\rec)$, of recommendation policy $\rec$ into two components:
    First, the recommendation may provide additional information on the risk of a bad outcome
    ($R$ affects the posterior belief $\P(Y{=}\bad|X_h, R)$ in \autoref{rem:Thresholds}),
    which improves optimal decisions $A^*$ of a recommendation-independent agent and leads to a gain $\textnormal{IG}(\rec)$.
    Second, the recommendation distorts the preferences of the decision-maker
    ($R$ affects $p^R$ in \autoref{rem:Thresholds}),
    which distorts decisions from the perspective of the principal and imposes a cost $\textnormal{Dist}(\rec)$.

    Our decomposition shows that a first consequence of recommendation dependence is inefficiently high compliance with the recommendation.
    Indeed, we can write
    \begin{align*}
        \textnormal{Dist}(\rec)
        &=
        \P(R{=}\risky) \:
        \E[\overbrace{\Ind{p^\risky {\geq} p(X_h,\risky) {>} p^*}}^{\text{over-adherence to $R = \risky$}}
            c_{II}
            \overbrace{(p(X_h,\risky) - p^*)}^{\geq 0}
        |R{=}\risky]
        \\
        &\phantom{=}
            +
        \P(R{=}\safe) \:
        \E[
            \underbrace{\Ind{p^\safe {<} p(X_h,\safe) {\leq} p^*}}_{\text{over-adherence to $R = \safe$}}
            c_I
            \underbrace{(p^*{-}p(X_h,\safe))}_{\geq 0}
            | R{=}\safe],
    \end{align*}
    where $p(X_h,R) = \P(Y{=}\bad|X_h,R)$ is the posterior probability of the bad outcome.
    When the risky recommendation is made, the agent uses an inefficiently high (from the perspective of the principal) threshold, leading to over-adherence to the recommendation and excess type-II loss. Similarly, for the safe recommendation, recommendation dependence leads to over-compliance for instances where the probability of the bad action is below the efficient threshold, but above the perturbed threshold of the agent.
    The degree of (over-) compliance and distortion is monotone in $\Delta_I,\Delta_{II}$:

    \begin{rem}[Recommendation dependence increases adherence and inefficiency]
    \label{rem:Adherence}
        Holding the recommendation policy $\rec: \mathcal{X}_m \rightarrow \mathcal{R}$ fixed,
        the probabilities $\P(A {=} R|R{=}\risky)$ and $\P(A {=} R|R{=}\safe)$ of adherence to the recommendation 
        as well as the principal's expected loss $\E[\ell(Y,A)]$
        are all (weakly) increasing in $\Delta_I$ and $\Delta_{II}$.
        Furthermore, as $\Delta_I, \Delta_{II} \rightarrow \infty$, $\P(A {\neq} R, \ell(Y,A) {>}0) \rightarrow 0$.
    \end{rem}

    Recommendation dependence not only reduces the efficiency of decisions; it also implies that a better-informed decision-maker does not necessarily make better decisions:

    \begin{prop}[More information does not imply better performance]
        \label{prop:Badinfo}
        Assume that $\Delta_I > 0$ or $\Delta_{II} > 0$.
        Consider two decision-makers, both with recommendation-dependent preferences.
        The first decision-maker only has access to $X_h^{(1)}$ (less informed) and the other has access to $X_h = (X_h^{(1)},X_h^{(2)})$ (more informed), in addition to the recommendation $R = \rec(X_m)$.
        Then we can choose random variables $(Y,X_m,X_h^{(1)},X_h^{(2)})$ and a joint distribution $\P$ over them such that expected loss is strictly higher for the more informed decision-maker than for the less informed one, for every non-trivial recommendation policy $\rec: \mathcal{X}_m \rightarrow \mathcal{R}$ (that does not always recommend the same action).
        Furthermore, this distribution can be chosen such that the same holds even when recommendation policies are chosen independently to be optimal for each decision-maker.
    \end{prop}

    This result represents a stark contrast to the case of a rational and aligned decision-maker, for whom more information always leads to better decisions. The result is driven by the incidence of inefficiencies in the $\text{Dist}(\rec)$ component: Additional information increases the spread of the posterior $p(X_h,R)$, which may make it more likely that it falls between the efficient threshold $p^*$ and biased decision thresholds $p^R$, thus leading to over-adherence in those cases.

    Based on our decomposition and these results, how should we think of the optimal design of recommendations?
    First, optimal recommendations are not generally the same as optimal algorithmic decisions, even in the absence of recommendation dependence, since good recommendations aim to maximize information gain rather than minimize the loss of implementing algorithmic choices directly.
    Second, the above results show that recommendation-dependent choices can lead to a trade-off between information gain and distortions that changes with the degree of recommendation dependence.
    Specifically, we would expect that increasing recommendation dependence related to one recommendation, such as increasing cost $\Delta_{II}$ from making a mistake against the $R {=} \safe$ recommendation, increases the distortion related to this recommendation and that it should therefore be given less.
    In \autoref{apx:Design}, we discuss optimal recommendations based on threshold rules, 
    and provide sufficient conditions under which this relationship holds in general.
    Here, we provide a simple example that illustrates these main features of recommendation-dependent choices and optimal binary recommendations.

        \begin{exm}[name=Independent uniform signals,label=exm1]
        
        Consider private signals $X_h$ and $X_m$ being drawn independently from a uniform distribution on $[0,1]$.
        Let the outcome $Y$ be deterministic in terms of $X_h$ and $X_m$,
        $Y = \bad$ if and only if $X_h+X_m \geq 1$,
        which is presented in Panel~(a) of \autoref{fig:ex_baseline}.
        When the agent decides by themselves, they need to act based solely on their observed private signal $X_h$. Since $\P(Y{=}\bad|X_h) = X_h$, the agent's optimal actions can be described in terms of the threshold rule
        \(
            A_0 = \risky
        \)
        if and only if $X_h \leq p^*$,
        where the threshold $p^* = \frac{c_{I}}{c_I + c_{II}}$ balances type-I and type-II errors optimally.
        This rule and the resulting expected loss are illustrated in Panel~(b) of \autoref{fig:ex_baseline}.     

        The recommendations that maximize the information gain in the decomposition from \eqref{eqn:Decomposition} in this example are given by $R = \risky$ if and only if $X_m \leq \sfrac{1}{2}$.
        This would be the optimal recommendation in this case if preferences were fully aligned.
        The agent without recommendation dependence would apply the same threshold $p^* = \frac{c_I}{c_I + c_{II}}$, regardless of the recommendation, to the posterior probability $\P(Y{=} \bad|X_h,R)$, leading to the second-best decision $A^*$.
        The resulting distribution of decisions and losses is depicted in Panel~(a) of \autoref{fig:ex-advised}.

        We now consider a decision-maker who perceives additional reference-dependent decision loss $\Delta_{II} > 0$ whenever they take a risky decision $A=\risky$ against a safe recommendation $R=\safe$ when that decision turns out to be a mistake. (We set $\Delta_I = 0$ for simplicity.)
        Recommendation dependence creates a misalignment between human decisions and the preferences of the principal whenever the recommendation $R = \safe$ is given, leading to an over-adherence to that recommendation.
        Specifically, the decision-maker observes an increased (perceived) cost of a type-II error to $c_{II} + \Delta_{II}$ when $R = \safe$,
        leading to decisions as in Panel~(b) of \autoref{fig:ex-advised}.

        We now consider how thresholds should be optimally set in the example.
        For thresholds $\gamma \in [0,1]$, we consider recommendations of the form
        \(
            R = \risky \text{ if and only if } X_m \leq \gamma
        \),
        to which an optimal agent response for $\Delta_I = 0 \leq \Delta_{II}$ is 
        \begin{align*}
            A
            &=
            \begin{cases}
                \risky, & X_h \leq \frac{c_I + (1-\gamma) c_{II}}{c_I+c_{II}} 
                \\
                \safe, & X_h > \frac{c_I + (1-\gamma) c_{II}}{c_I+c_{II}}
            \end{cases}
            \text{ for }
            R = \risky,
            &
            A
            &=
            \begin{cases}
                \risky, & X_h \leq \frac{(1-\gamma) \: c_{I}}{c_I+c_{II}+\Delta_{II}}
                \\
                \safe, & X_h > \frac{(1-\gamma) \: c_{I}}{c_I+c_{II}+\Delta_{II}}
            \end{cases}
            \text{ for }
            R = \safe.
        \end{align*}
        When there is no recommendation dependence, $\Delta_{II} = 0$, the optimal choice of threshold is $\gamma^* = \sfrac{1}{2}$.
        The optimal threshold $\gamma^*_{\Delta_{II}}$ that minimizes the expected loss of the principal generally depends on the degree of recommendation dependence and is increasing in $\Delta_{II}$.%
        \footnote{
            Specifically, 
        $\gamma^*_{\Delta_{II}} = \frac{1}{2}\left(1 + \sfrac{\left(\frac{c_I\Delta_{II}}{c_I+c_{II} + \Delta_{II}}\right)^2}{\left(2c_Ic_{II} + \left(\frac{c_I\Delta_{II}}{c_I+c_{II} + \Delta_{II}}\right)^2 \right)}\right)$.
       } 
        Thus, when $\Delta_{II} > 0$ (and $\Delta_I = 0$),
        the principal optimally shifts the threshold towards giving the safe recommendation less (Panel~(a) of \autoref{fig:ex-adjusted}).
        As a response, the agent slightly adjusts their threshold towards taking the risky action less often in both recommendation regimes (Panel~(b) of \autoref{fig:ex-adjusted}), but still takes the risky action more often than when the threshold $\gamma = \sfrac{1}{2}$ is used.
    \end{exm}

    \begin{figure}[p!]
        \vspace{-4em}
        \centering
        \begin{subfigure}[T]{0.45\textwidth}
            \centering

            \begin{tikzpicture}
                \begin{axis}[
                    axis equal image,
                    xlabel={$X_h$}, xlabel style={yshift=15pt},
                    ylabel={$X_m$}, ylabel style={yshift=-15pt},
                    xmin=0, xmax=1,
                    ymin=0, ymax=1,
                    width=.65\textwidth,
                    height=.65\textwidth,
                    xtick={0,1},
                    ytick={0,1},
                    tick align=outside,
                    enlargelimits=false,
                    grid style=dashed,
                    grid=both,
                    axis on top
                ]
                \fill [blue!10] (axis cs:0,0) -- (axis cs:0,1) -- (axis cs:1,0) -- cycle;
                \fill [red!10] (axis cs:0,1) -- (axis cs:1,1) -- (axis cs:1,0) -- cycle;         

                \node[color=blue!70] at (axis cs:0.3,0.3) {$\good$};
                \node[color=red!70] at (axis cs:0.7,0.7) {$\bad$};

                \addplot[color=black] coordinates {(0,1) (1,0)};
                
                \end{axis}
            \end{tikzpicture}

            \caption{Distribution of human ($X_h$) and machine ($X_m$) signals along with resulting outcomes ($Y =\good$, light blue; $Y = \bad$, light red).}
        \end{subfigure}
        \hspace{10pt}
        \begin{subfigure}[T]{0.45\textwidth}
            \centering

            \begin{tikzpicture}
                \begin{axis}[
                    axis equal image,
                    xlabel={$X_h$}, xlabel style={yshift=15pt},
                    ylabel={$X_m$}, ylabel style={yshift=-15pt},
                    xmin=0, xmax=1,
                    ymin=0, ymax=1,
                    width=.65\textwidth,
                    height=.65\textwidth,
                    xtick={0,1},
                    ytick={0,1},
                    tick align=outside,
                    enlargelimits=false,
                    grid style=dashed,
                    grid=both,
                    axis on top
                ]
                \fill [blue!10] (axis cs:0,0) -- (axis cs:0,1) -- (axis cs:1,0) -- cycle;
                \fill [red!10] (axis cs:0,1) -- (axis cs:1,1) -- (axis cs:1,0) -- cycle;     
                \fill [red!70] (axis cs:0,1) -- (axis cs:0.4,1) -- (axis cs:0.4,0.6) -- cycle;
                \fill [blue!70] (axis cs:0.4,0.6) -- (axis cs:0.4,0) -- (axis cs:1,0) -- cycle;
            
                \addplot[color=black] coordinates {(0,1) (1,0)};
                
                \addplot[color=black, ultra thick] coordinates {(0.4,0) (0.4,1)};

                \node at (axis cs:0.2,0.4) {$\risky$};
                \node at (axis cs:0.7,0.6) {$\safe$};
                \node[color=white] at (axis cs:0.267,0.867) {$c_{II}$};
                \node[color=white] at (axis cs:0.6,0.2) {$c_I$};
                \end{axis}
            \end{tikzpicture}

            \caption{Optimal decision $A_0$ of the human decision-maker acting alone.
            Loss $c_{II}$ is incurred in the top triangle (red) and loss $c_I$ in the bottom triangle (blue).}
        \end{subfigure}
        \caption{Joint distribution of outcome, human signal, and machine signal in \autoref{exm1}, along with the optimal decision of a human decision-maker acting without recommendation.}
        \label{fig:ex_baseline}
    \end{figure}

    \begin{figure}[p!]
        \centering
        \vspace{-1em}
        \begin{subfigure}[T]{0.45\textwidth}
            \centering

            \begin{tikzpicture}
                \begin{axis}[
                    axis equal image,
                    xlabel={$X_h$}, xlabel style={yshift=15pt},
                    ylabel={$X_m$}, ylabel style={yshift=-15pt},
                    xmin=0, xmax=1,
                    ymin=0, ymax=1,
                    width=.65\textwidth,
                    height=.65\textwidth,
                    xtick={0,1},
                    ytick={0,1},
                    tick align=outside,
                    enlargelimits=false,
                    grid style=dashed,
                    grid=both,
                    axis on top,
                    clip=false
                ]
                \useasboundingbox (0,0) rectangle (1,1.1);
                \fill [blue!10] (axis cs:0,0) -- (axis cs:0,1) -- (axis cs:1,0) -- cycle;
                \fill [red!10] (axis cs:0,1) -- (axis cs:1,1) -- (axis cs:1,0) -- cycle;     
                \fill [red!70] (axis cs:0,1) -- (axis cs:0.2,1) -- (axis cs:0.2,0.8) -- cycle;
                \fill [blue!70] (axis cs:0.2,0.8) -- (axis cs:0.2,0.5) -- (axis cs:0.5,0.5) -- cycle;
                \fill [red!70] (axis cs:0.5,0.5) -- (axis cs:0.7,0.5) -- (axis cs:0.7,0.3) -- cycle;
                \fill [blue!70] (axis cs:0.7,0.3) -- (axis cs:0.7,0) -- (axis cs:1,0) -- cycle;
                \addplot[color=black] coordinates {(0,1) (1,0)};                
                \addplot[color=black, ultra thick] coordinates {(0.2,0.5) (0.2,1)};
                \addplot[color=black, ultra thick] coordinates {(0.7,0) (0.7,0.5)};
                \addplot[color=black, ultra thick, dashed] coordinates {(0,0.5) (1,0.5)};
                \node at (axis cs:0.2,0.25) {$\risky$};
                \node at (axis cs:0.7,0.75) {$\safe$};
                \node[color=white] at (axis cs:0.133,0.95) {$c_{II}$};
                \node[color=white] at (axis cs:0.3,0.6) {$c_I$};
                \node[color=white] at (axis cs:0.633,0.45) {$c_{II}$};
                \node[color=white] at (axis cs:0.8,0.1) {$c_I$};
                \end{axis}
            \end{tikzpicture}

            \caption{The machine's recommendation separates the space into two regions (dashed line), one for each recommended action ($\safe$ on top, $\risky$ on the bottom). The human decision-maker decides according to two separate thresholds.}
        \end{subfigure}
        \hspace{10pt}
        \begin{subfigure}[T]{0.45\textwidth}
            \centering

            \begin{tikzpicture}
                \begin{axis}[
                    axis equal image,
                    xlabel={$X_h$}, xlabel style={yshift=15pt},
                    ylabel={$X_m$}, ylabel style={yshift=-15pt},
                    xmin=0, xmax=1,
                    ymin=0, ymax=1,
                    width=.65\textwidth,
                    height=.65\textwidth,
                    xtick={0,1},
                    ytick={0,1},
                    tick align=outside,
                    enlargelimits=false,
                    grid style=dashed,
                    grid=both,
                    clip=false,
                    axis on top
                ]
                \useasboundingbox (0,0) rectangle (1,1.1);
                \fill [blue!10] (axis cs:0,0) -- (axis cs:0,1) -- (axis cs:1,0) -- cycle;
                \fill [red!10] (axis cs:0,1) -- (axis cs:1,1) -- (axis cs:1,0) -- cycle;     
                \fill [red!70] (axis cs:0,1) -- (axis cs:0.12,1) -- (axis cs:0.12,0.88) -- cycle;
                \fill [blue!70] (axis cs:0.12,0.88) -- (axis cs:0.12,0.5) -- (axis cs:0.5,0.5) -- cycle;
                \fill [red!70] (axis cs:0.5,0.5) -- (axis cs:0.7,0.5) -- (axis cs:0.7,0.3) -- cycle;
                \fill [blue!70] (axis cs:0.7,0.3) -- (axis cs:0.7,0) -- (axis cs:1,0) -- cycle;
                \addplot[color=black] coordinates {(0,1) (1,0)};                
                \addplot[color=black] coordinates {(0.2,0.5) (0.2,1)};
                \addplot[color=black, ultra thick] coordinates {(0.12,0.5) (0.12,1)};
                \draw[->, color=black, ultra thick] (axis cs:0.2,.75) -- (axis cs:0.12,.75);
                \addplot[color=black, ultra thick] coordinates {(0.7,0) (0.7,0.5)};
                \addplot[color=black, ultra thick, dashed] coordinates {(0,0.5) (1,0.5)};
                \node at (axis cs:0.2,0.25) {$\risky$};
                \node at (axis cs:0.7,0.75) {$\safe$};
                \node[color=red!70] at (axis cs:0.16,1.08) {$c_{II} + \Delta_{II}$};
                \node[color=white] at (axis cs:0.275,0.6) {$c_I$};
                \node[color=white] at (axis cs:0.633,0.45) {$c_{II}$};
                \node[color=white] at (axis cs:0.8,0.1) {$c_I$};
                \end{axis}
            \end{tikzpicture}

            \caption{
                The decision-maker anticipates additional loss $\Delta_{II} > 0$ from mistakes from choosing the $\risky$ action against a $\safe$ recommendation, leading to a more stringent threshold when $\safe$ is recommended (top area).}
                \label{fig:ex-advised-recdep}
        \end{subfigure}
        \caption{Comparison of machine-assisted decisions without recommendation dependence (left) and with recommendation dependence (right) for \autoref{exm1}.}
        \label{fig:ex-advised}
    \end{figure}

    \begin{figure}[p!]
            \centering
            \vspace{-1em}
        \begin{subfigure}[b]{0.45\textwidth}
            \centering
            \begin{tikzpicture}
                \begin{axis}[
                    axis equal image,
                    xlabel={$X_h$}, xlabel style={yshift=15pt},
                    ylabel={$X_m$}, ylabel style={yshift=-15pt},
                    xmin=0, xmax=1,
                    ymin=0, ymax=1,
                    width=.65\textwidth,
                    height=.65\textwidth,
                    xtick={0,1},
                    ytick={0,1},
                    tick align=outside,
                    enlargelimits=false,
                    grid style=dashed,
                    grid=both,
                    clip=false,
                    axis on top
                ]
                \useasboundingbox (0,0) rectangle (1,1);
                \fill [blue!10] (axis cs:0,0) -- (axis cs:0,1) -- (axis cs:1,0) -- cycle;
                \fill [red!10] (axis cs:0,1) -- (axis cs:1,1) -- (axis cs:1,0) -- cycle;     
                \addplot[color=black] coordinates {(0,1) (1,0)};                
                \addplot[color=black, dashed] coordinates {(0,0.5) (1,0.5)};

                \addplot[color=black, ultra thick, dashed] coordinates {(0,0.6) (1,0.6)};

                \draw[->, color=black, ultra thick] (axis cs:0.5,.5) -- (axis cs:0.5,.6);

                \node at (axis cs:0.2,0.25) {$\risky$};
                \node at (axis cs:0.7,0.75) {$\safe$};
                \end{axis}
            \end{tikzpicture}
            \caption{The machine threshold shifts to reduce the probability of the region with misaligned decision losses, recommending the risky action more.}
        \end{subfigure}
        \hspace{10pt}
        \begin{subfigure}[b]{0.45\textwidth}
            \centering
            \begin{tikzpicture}
                \begin{axis}[
                    axis equal image,
                    xlabel={$X_h$}, xlabel style={yshift=15pt},
                    ylabel={$X_m$}, ylabel style={yshift=-15pt},
                    xmin=0, xmax=1,
                    ymin=0, ymax=1,
                    width=.65\textwidth,
                    height=.65\textwidth,
                    xtick={0,1},
                    ytick={0,1},
                    tick align=outside,
                    enlargelimits=false,
                    grid style=dashed,
                    grid=both,
                    clip=false,
                    axis on top
                ]
                \useasboundingbox (0,0) rectangle (1,1);
                \fill [blue!10] (axis cs:0,0) -- (axis cs:0,1) -- (axis cs:1,0) -- cycle;
                \fill [red!10] (axis cs:0,1) -- (axis cs:1,1) -- (axis cs:1,0) -- cycle;     
                \addplot[color=black] coordinates {(0,1) (1,0)};                
                \addplot[color=black] coordinates {(0.12,0.5) (0.12,1)};
                \addplot[color=black] coordinates {(0.7,0) (0.7,0.5)};
                \addplot[color=black, dashed] coordinates {(0,0.5) (1,0.5)};

                \addplot[color=black, ultra thick] coordinates {(0.096,0.6) (0.096,1)};
                \addplot[color=black, ultra thick] coordinates {(0.66,0) (0.66,0.6)};
                \addplot[color=black, ultra thick, dashed] coordinates {(0,0.6) (1,0.6)};

                \draw[->, color=black, ultra thick] (axis cs:0.176,.8) -- (axis cs:0.096,.8);
                \draw[->, color=black, ultra thick] (axis cs:0.74,.3) -- (axis cs:0.66,.3);

                \node at (axis cs:0.2,0.25) {$\risky$};
                \node at (axis cs:0.7,0.75) {$\safe$};
                \end{axis}
            \end{tikzpicture}
            \caption{In response, the decision-maker adjusts the decision thresholds to the left, but is still more likely overall to take the risky action than in \autoref{fig:ex-advised-recdep}.}
        \end{subfigure}
        \caption{Optimal decision thresholds are adjusted in response to recommendation dependence. The thin lines show the thresholds in \autoref{fig:ex-advised}, while the arrows depict the optimal change in machine threshold (left) and resulting adjustment of conditional decision-maker choices (right).}
        \label{fig:ex-adjusted}
                \vspace{-3em}

    \end{figure}

    The example illustrates how recommendation dependence affects optimal recommendation algorithms. We end this section by discussing how much the optimal design of the recommendation algorithm matters for the principal's loss.
    On the one hand, recommendation dependence ensures that the resulting decisions will perform at least as well as the recommendation algorithm itself:
    
    \begin{rem}[Improvement over the machine]
        \label{rem:Simpleimprovement}
        For $A$ the recommendation-dependent choice following a recommendation $R = \rec(X_m)$,
        we have that $\E[\ell(Y,A)] \leq \E[\ell(Y,R)]$.
    \end{rem}

    This result is driven by the structure of recommendation-dependent choices: since the only inefficiency comes from over-adherence, at worst, the human decision-maker does at least as well as the recommendation itself.
    On the other hand, this result does not guarantee that the assisted decision improves over the unassisted human decisions.
    Indeed, this is generally infeasible:
    
    \begin{prop}[Inefficient recommendations]
        \label{prop:Inefficient}
        Assume that $\Delta_I,\Delta_{II} > 0$.
        Then, providing a recommendation can be worse than not providing a recommendation at all, that is, there are distributions $\P$ for which the loss $\E[\ell(Y,A)]$ is higher for any recommendation policy than the loss $\E[\ell(Y,A_0)]$ without any machine assistance.
    \end{prop}
    
    This result stands in contrast to the case with recommendation-independent preferences ($\Delta_I = 0 = \Delta_{II}$), in which case any (correctly interpreted) recommendation (weakly) improves loss because the human decision-maker uses information efficiently.
    As a result, we cannot generally find binary recommendations that ensure human--machine complementarity, as it is always possible that not providing any recommendations at all would lead to better outcomes.

    How different can algorithms be that optimize for recommendation-dependent preferences?
    Based on our decomposition, the inefficiency from an algorithm that only targets rational choices comes from choices where the agent is uncertain and their posterior is around the optimal decision threshold $p^*$, in which case the recommendation may sway recommendation-dependent choices towards following the recommendation when it is not efficient to do so.
    Optimal recommendations that are aware of this inefficiency would minimize ambiguous cases where the posterior falls between $p^*$ and $p^R$, and instead group cases together for which the agent's posterior is unambiguous.
    Rather than working out the structure of such second-best recommendations, in the next section, we instead consider changes to the recommendations that can improve both the information gain \emph{and} reduce distortions by allowing the recommendation algorithm to withhold recommendations.
        
    \section{Strategic Non-Recommendations}
    \label{sec:NonRec}

    We have shown that recommendation dependence introduces inefficiencies that make the value of the recommendations ambiguous and affect their optimal design.
    When recommendations distort choices, one solution is to strategically withhold recommendations in cases where the decision-maker knows better which decisions to take.
    \cite{shashikumar2021artificial} proposes training a recommendation algorithm to return an ``I don't know'' response and applies the idea in the context of sepsis prediction.
    Within our formal model, we capture this approach by considering recommendations with three values, $\mathcal{R} = \{\risky, \neutral, \safe\}$.
    We assume that preferences still follow \eqref{eqn:Recdeppref}, meaning that there is no additional loss when the neutral recommendation is given.%
    \footnote{This extension is consistent with our microfoundations in \autoref{sec:Foundations} if we assume that the neutral recommendation precludes loss aversion (\autoref{subsec:Refdep}); does not set a default and does not create a cost (or creates equal cost) when choosing an action (\autoref{subsec:Default}); and does not count towards ``correct'' decisions  (\autoref{subsec:Blamegame}).}

    Such a recommendation structure relaxes the restriction that the provided information is binary to allow for three levels, so we would expect it to improve outcomes even in a model without recommendation dependence.
    However, with recommendation dependence, there can be an additional gain: if there is no additional cost from mistakes in the neutral case, as in our case, then allowing for this third level also reduces the cost from recommendation dependence.
    In terms of the decomposition in \eqref{eqn:Decomposition}, introducing the neutral level simultaneously improves the information gain and reduces the distortion.
    Specifically, take a recommendation algorithm $\rec_-$ that only makes recommendations in $\{\risky, \safe\}$,
    and a recommendation algorithm $\rec_+$ that also uses the neutral level. As long as the two recommendation algorithms do not disagree when both make a definite recommendation (that is, $\rec_+(x_m) = \rec_-(x_m)$ whenever $\rec_+(x_m) \neq \neutral$), then moving from two to three recommendation levels both improves information gain and reduces over-adherence:
    \begin{align*}
        \text{IG}(\rec_+) &\geq \text{IG}(\rec_-),
        &
        \text{Dist}(\rec_+) &\leq \text{Dist}(\rec_-).
    \end{align*}
    As a consequence, providing strategic non-recommendations can have a strictly higher benefit in our model relative to a rational baseline where only the information gain increases, as we illustrate in an application to \autoref{exm1}.
    
    \begin{exm}[name=Independent uniform signals,continues=exm1]
        In the example with uniform independent signals $X_h$ and $X_m$, consider algorithmic recommendations
        \[
            R = 
            \begin{cases}
                \risky, & X_m \leq \gamma^{\low}, \\
                \neutral, & \gamma^{\low} < X_m \leq \gamma^{\high},
                \\
                \safe, & X_m > \gamma^{\high}.
            \end{cases}
        \]
        with thresholds $0 \leq \gamma^{\low} \leq \gamma^{\high} \leq 1$.
        Without considering recommendation dependence, adding a neutral option improves decisions by increasing the amount of information about the machine signal $X_m$ preserved in the recommendation $R$. In the baseline case without recommendation dependence, the machine would optimally provide recommendations based on thresholds $\gamma^{\low} = \sfrac{1}{3}, \gamma^{\high} = \sfrac{2}{3}$, equally dividing the signal space to maximize the amount of information in the recommendation. Recommendation dependence changes optimal recommendations by reducing the frequency of situations in which the safe recommendation is given, as this recommendation distorts decisions. Thus, both thresholds increase, as we visualize in \autoref{fig:ex-nonrec}. The resulting reduction in expected loss relative to the binary-recommendation case from \autoref{fig:ex-adjusted} is larger than in the case without recommendation dependence.
        This is because loss is reduced through increased granularity of the machine recommendation and a reduction in the cases for which recommendation dependence distorts choices.
    \end{exm}

    \begin{figure}[h]
        \centering
        \begin{tikzpicture}
            \begin{axis}[
                axis equal image,
                xlabel={$X_h$}, xlabel style={yshift=15pt},
                ylabel={$X_m$}, ylabel style={yshift=-15pt},
                xmin=0, xmax=1,
                ymin=0, ymax=1,
                width=.4\textwidth,
                height=.4\textwidth,
                xtick={0,1},
                ytick={0,1},
                tick align=outside,
                enlargelimits=false,
                grid style=dashed,
                grid=both,
                clip=false,
                axis on top
            ]
            \fill [blue!10] (axis cs:0,0) -- (axis cs:0,1) -- (axis cs:1,0) -- cycle;
            \fill [red!10] (axis cs:0,1) -- (axis cs:1,1) -- (axis cs:1,0) -- cycle;

            \addplot[color=black, dashed] coordinates {(0,0.33) (1,0.33)};
            \addplot[color=black, dashed] coordinates {(0,0.67) (1,0.67)};
            \draw[->, color=black, ultra thick] (axis cs:0.2,.33) -- (axis cs:0.2,.375);
            \draw[->, color=black, ultra thick] (axis cs:0.7,.67) -- (axis cs:0.7,.75);

            \fill [red!70] (axis cs:0,1) -- (axis cs:0.06,1) -- (axis cs:0.06,0.94) -- cycle;
            \fill [blue!70] (axis cs:0.06,0.94) -- (axis cs:0.06,0.75) -- (axis cs:0.25,0.75) -- cycle;
            \fill [red!70] (axis cs:0.25,0.75) -- (axis cs:0.4,0.75) -- (axis cs:0.4,.6) -- cycle;
            \fill [blue!70] (axis cs:0.4,.6) -- (axis cs:0.4,.375) -- (axis cs:.625,.375) -- cycle;
            \fill [red!70] (axis cs:.625,.375) -- (axis cs:.775,.375) -- (axis cs:.775,.225) -- cycle;
            \fill [blue!70] (axis cs:.775,.225) -- (axis cs:.775,0) -- (axis cs:1,0) -- cycle;
            \addplot[color=black] coordinates {(0,1) (1,0)};                
            \addplot[color=black, ultra thick] coordinates {(0.06,0.75) (0.06,1)};
            \addplot[color=black, ultra thick] coordinates {(0.4,.75) (0.4,0.375)};
            \addplot[color=black, ultra thick] coordinates {(0.775,0) (.775,0.375)};
            \addplot[color=black, ultra thick, dashed] coordinates {(0,0.75) (1,0.75)};
            \addplot[color=black, ultra thick, dashed] coordinates {(0,0.375) (1,0.375)};
            \node at (axis cs:0.7,0.875) {$A=\safe$};
            \node at (axis cs:0.2,0.1875) {$A=\risky$};
            \node[anchor=west] at (axis cs:1.1,.1875) {$R=\risky$};
            \node[anchor=west] at (axis cs:1.1,0.5625) {$R=\neutral$};
            \node[anchor=west] at (axis cs:1.1,0.875) {$R=\safe$};
            \node[color=red!70] at (axis cs:0.1,1.05) {$c_{II} + \Delta_{II}$};

            \node[color=white] at (axis cs:0.12,0.81) {$c_I$};

            \node[color=white] at (axis cs:0.35,0.70) {$c_{II}$};
            \node[color=white] at (axis cs:0.475,.45) {$c_I$};  

            \node[color=white] at (axis cs:.725,.325) {$c_{II}$};
            \node[color=white] at (axis cs:.85,0.075) {$c_I$};

            \end{axis}
        \end{tikzpicture}

        \caption{Incorporating a neutral recommendation provides additional information to the human decision-maker, while also limiting the region in which recommendation dependence distorts choices.}
        \label{fig:ex-nonrec}
    \end{figure}

    Having discussed the effect of an additional recommendation option in the example,
    we now show that adding a third option can achieve combined human--machine decisions that are at least as good (from the principal's perspective) as implementing either human or machine decisions alone,
    or implementing even the best recommendation policy with two recommendation levels.
    
    \begin{prop}[Human--machine complementarity]
        \label{prop:Complementarity}
        There are recommendation policies $R = \rec(X_m)$ such that the expected loss (weakly) improves over the best machine-only decision $A_m$, best human-only decision $A_0$, and the decision $A_-$ following the optimal two-level recommendation $R_- \in \{\risky,\safe\}$:
        \(
            \E[\ell(Y,A)] \leq \min\left\{\E[\ell(Y,A_m)], \E[\ell(Y,A_0)], \E[\ell(Y,A_-)]\right\}.
        \)
    \end{prop}

    This result shows that, without imposing any substantive assumptions, a recommendation policy that includes a neutral level can achieve complementarity within our model.
    It stands in contrast to the case of binary recommendations: indeed, \autoref{prop:Inefficient} shows that the same complementarity cannot generally be achieved without adding the neutral level, as a two-level recommendation may perform worse than not making any recommendations at all, \emph{even if} recommendations are chosen to be optimal for a decision-maker with recommendation-dependent preferences.
    
    While this result and the example demonstrate the value of strategic non-recommendations, they leave open the optimal design of such recommendations in general. In \autoref{apx:Design}, we consider the optimal choice of thresholds for recommendations with three levels, and show a high-level result for how they change with the level of recommendation dependence. However, those results depend on infinitesimal approximations, assumptions about the distribution, and the designer's ability to fully know the joint distribution of outcomes as well as (partially private) signals. Instead, we now focus on solutions in the case where the designer has limited information, and show that minimax optimal solutions take an intuitive form in those cases.

    \section{Practical Implementation and Feasible Complementarity}
    \label{sec:Implementation}

    Above, we have shown that the design of recommendation algorithms has to take biased human decisions into account to be effective. Specifically, providing naive algorithmic recommendations can make choices worse rather than better. Our remedy -- namely, optimal recommendations that correctly anticipate recommendation dependence and stay strategically silent -- has two weaknesses for practical implementation.
    First, it assumes the algorithm designer knows the degree of the human decision-maker's recommendation dependence.
    Second, and more problematically, it assumes knowledge by the algorithm designer of the joint distribution of outcomes, machine information, and (private) human information.
        
     In this section, we revisit the setup with binary recommendations and strategic non-recommendations from \autoref{sec:NonRec}, but we consider \emph{feasible} implementations based on limited information available to the designer of the algorithm. Specifically, we assume that the designer of the algorithm may have incomplete knowledge of the distribution $\P$.
     In addition, we also assume that the degrees $\Delta_I,\Delta_{II} \geq 0$ of recommendation dependence are generally unknown to the designer.
    We then consider minimax optimal solutions in this setting.
    The approach adopts a similar strategy to the recent study of feasibly achieving human--AI complementarity in the potential-outcomes model of \citecomplementarity{}, and leads to similar triage solutions.
    However, our setup differs from \citecomplementarity{} in that its general assumptions do not hold in our model (specifically, monotonicity may be violated), and we instead leverage the specific structure of recommendation-dependent preferences to derive minimax optimal algorithms. In addition, we consider results for cases where outcome data is only partially observed (\autoref{prop:Unobservedminimax}) and recommendation dependence is one-sided (\autoref{prop:Onesidedminimax}).
     
     The designer's incomplete information about $\P$ comes in the form of a joint distribution $\overline{\P}$ over outcomes $Y$, machine characteristics $X_m$, as well as unassisted human choices $A_0$, where
    $A_0 = \risky$ if and only if $\P(Y{=}\bad|X_h) \leq p^*$, as in \eqref{eqn:A0}.
     This distribution represents knowledge of training data where no recommendations were provided, but baseline decisions and outcomes were observed. For example, in a medical context, the data may represent machine-readable patient records $X_m$, the decision $A_0$ of a doctor whether to prescribe some medication or order some test, and the actual medical diagnosis $Y$.
     Specifically, we assume here that the designer of the algorithm does not have access to the human-only features $X_h$, but only to the resulting decision $A_0$.
     This represents a realistic constraint where some features are never recorded and not even available at training time, such as the detailed patient answers to questions by a doctor.
     For the human decision-maker, we continue to assume that they know the distribution of $(Y,X_h,R)$ when taking their decision given a recommendation $R=\rec(X_m)$, but we do not require that they also know the distribution of the machine information $X_m$ (see also \autoref{ftn:Agentinfo}).

     Relative to the results in \autoref{sec:NonRec}, knowledge of the distribution $\overline{\P}$ over $(Y,X_m,A_0)$ only provides partial information of the complete distribution $\P$ over $(Y,X_m,X_h)$ (and no insight into the degree $\Delta_I,\Delta_{II}$ of recommendation dependence).
     Let $\mathcal{P}(\overline{\P})$ denote the set of all distributions over $(Y,X_m,X_h)$ such that the implied distribution over $(Y,X_m,A_0)$ is equal to $\overline{\P}$.
     Also let
        \begin{align*}
            \overline{\rec}_m(x_m)
            =
            \underset{a \in \{\risky,\safe\} }{\mathrm{argmin}} \overline{\E}[\ell(Y,a)|X_m=x_m]
            =
            \begin{cases}
                \risky, & \overline{\P}(Y{=}\bad|X_m=x_m) \leq p^*,
                \\
                \safe, & \overline{\P}(Y{=}\bad|X_m=x_m) > p^*
            \end{cases},
        \end{align*}
    denote the optimal algorithmic decision in the absence of human intervention.

     \begin{prop}[Minimax optimal triage algorithm]
     \label{prop:Triageminimax}
        Assume that $\overline{\P}(Y{=}\bad|X_m,A_0) \in (0,1)$ and that $\overline{\E}[\ell(Y, A_0)|X_m] \leq \overline{\E}[\ell(Y, A^\dagger_0)|X_m]$ almost surely, where $A^\dagger_0$ is the inverse decision of $A_0$ (that is, $A^\dagger_0 = \risky$ if and only if $A_0 = \safe$ and $A^\dagger_0 = \safe$ if $A_0 = \risky$).
        Then the algorithm
         \begin{equation}
            \label{eqn:Triage}
             \overline{\rec}(x_m)
             =
             \begin{cases}
                \overline{\rec}_m(x_m), &\overline{\E}[\ell(Y,\overline{\rec}_m(x_m))|X_m=x_m] < \overline{\E}[\ell(Y,A_0)|X_m=x_m], \\
                \neutral, &\text{otherwise}
             \end{cases}
         \end{equation}
         minimizes the worst-case loss
         \(
             \textstyle\sup_{\P \in \mathcal{P}(\overline{\P}), \Delta_I \geq 0, \Delta_{II} \geq 0}
             \E[\ell(Y,A)]
         \)
         over all recommendation algorithms $\rec: \mathcal{X}_m \rightarrow \{ \risky,\neutral, \safe \}$,
         where $A$ denotes the recommendation-dependent choices of a human decision-maker with $\Delta_I,\Delta_{II}$ following recommendations $R = \rec(X_m)$.
     \end{prop}
     
     This minimax algorithm has an intuitive form akin to a triage solution \citep{raghu_algorithmic_2019}:
     It proposes the best machine action $\overline{\rec}_m(X_m)$ whenever it outperforms the baseline human action,
     and otherwise does not send a recommendation to let the human decision-maker take an ``active'' decision in the parlance of \citecomplementarity.
     
    The assumption on $A^\dagger_0$ ensures that the human decision-maker's baseline actions do not misrank instances even after conditioning on the machine information $X_m$, and ensures that the correct benchmark is the actual decision $A_0$ rather than its inverse $A^\dagger_0$. This assumption is not necessary, and is testable since it only depends on the distribution $\overline{\P}$. We drop this assumption to derive a general minimax solution in \autoref{apx:Minimax}, which also considers cases where the human decision-maker can improve decisions by switching from $A_0$ to $A^\dagger_0$ once recommendations are given.
    However, this distinction is not essential for achieving complementarity.
    Indeed, irrespective on whether the assumptions on $\overline{\P}(Y{=}\bad|X_m,A_0)$ and $A^\dagger_0$ in \autoref{prop:Triageminimax} hold, the proposed triage algorithm guarantees human--algorithm complementarity in the sense of \autoref{prop:Complementarity}:

     \begin{prop}[Feasible human--algorithm complementarity]
        \label{prop:Feasiblecomplementarity}
        The algorithm from 
        \autoref{prop:Triageminimax}
        improves over both human and machine decisions,
        \(\E[\ell(Y,A)] \leq \min\{\E[\ell(Y,A_0)], \E[\ell(Y,A_m)]\},\)
        where $A$ denotes the recommendation-dependent choices of a human decision-maker with $\Delta_I,\Delta_{II}$ following recommendations $R = \overline{\rec}(X_m)$, and $A_m$ represents optimal machine-only decisions. 
     \end{prop}

     This dominance statement holds irrespective of whether the additional assumptions in \autoref{prop:Triageminimax} hold.
     It confirms that the addition of the third, neutral recommendation level has inherent value in achieving complementarity, even when the full distribution $\P$ is not known. (Note that, as before, using only the two recommendations $\mathcal{R} = \{\risky,\safe\}$ would not be able to guarantee complementarity by \autoref{prop:Inefficient}.)

     How large is the improvement guaranteed by the simple triage-type algorithm $\overline{\rec}$?
     The algorithm guarantees the tight upper bound
     \(
         \overline{\E}[
            \min\{
                \overline{\E}[\ell(Y,A_0)|X_m],
                \overline{\E}[\ell(Y,\risky)|X_m],
                \overline{\E}[\ell(Y,\safe)|X_m]
            \}
         ]  
     \)
     that can be calculated directly from the limited information in $\overline{\P}$, and chooses, for every value of $X_m$, among the baseline decision of the agent, the risky action, and the safe one.
     We would therefore expect this recommendation algorithm to do particularly well whenever the human decision-maker makes a large number of \emph{predictable} errors, for which providing a recommendation leads to a guaranteed improvement.

     In practice, we may face an additional hurdle when not all outcomes are observed. Specifically, the observability of the outcome $Y$ may itself depend on the human action in the training data.
     For example, we will only learn if a defendant would commit a crime or fail to appear at their trial if a judge chooses to release that defendant. Similarly, we may only find out a true medical diagnosis if the right test is performed. In cases like this, the algorithm designer would only have access to a distribution $\overline{\P}'$ over $(Y',X_m,A_0)$ with
     \(Y' = Y\) for $A_0 = \risky$ and $Y'$ unobserved otherwise.
     Defining $\mathcal{P}'(\overline{\P}')$ analogously to $\mathcal{P}(\overline{\P})$ as the distributions $\P$ that are consistent with $\overline{\P}'$, our proposed minimax solution now takes an even easier form.
     
     \begin{prop}[Minimax recommendation with limited observability]
     \label{prop:Unobservedminimax}
        The recommendation algorithm
         \begin{equation}
             \overline{\rec}'(x_m)
             =
             \begin{cases}
                \safe, &\overline{\P}'(Y'= \bad|X_m=x_m, A_0 = \risky) > p^* = \frac{c_I}{c_I + c_{II}}, \\
                \neutral, &\text{otherwise}
             \end{cases}
         \end{equation}
         minimizes the worst-case loss
         \(
             \textstyle\sup_{\P \in \mathcal{P}'(\overline{\P}'), \Delta_I \geq 0, \Delta_{II} \geq 0}
             \E[\ell(Y,A)]
         \)
         over all recommendation algorithms $\rec: \mathcal{X}_m \rightarrow \{ \risky,\neutral, \safe \}$,
         where $A$ denotes the recommendation-dependent choices of a human decision-maker with $\Delta_I,\Delta_{II}$ who observes recommendations $R = \rec(X_m)$.
     \end{prop}

     This algorithm is now asymmetrical between the risky and the safe recommendations. Specifically, it never recommends the risky action.
     The reason is that the algorithm designer cannot predict the impact of switching from a safe to a risky action, since outcomes were not observed for those instances.
     The minimax optimal recommendation algorithm instead stays silent for a larger number of instances. As a consequence, the machine-assisted human decisions are still guaranteed to outperform the unassisted decisions $A_0$ as well as any \emph{feasible} algorithm that can generally be learned from the limited distribution $\overline{\P}'$.
     Specifically, we still obtain complementarity in the sense that $\E[\ell(Y,A)] \leq \min\{\E[\ell(Y,A_0)], \E[\ell(Y,A_m')]\}$, where $A_m' = \risky$ if and only if $\overline{\P}'(Y'= \bad|X_m=x_m, A_0 = \risky) \leq p^*$.
     Again, the availability of the neutral option is essential for achieving complementarity in this sense.%
     \footnote{
        Here, there may be better (infeasible) algorithmic decisions that could be learned from additional information about the distribution of $(Y,X_m)$ beyond $\overline{\P}'$, so we do not find the same strong sense of complementarity from \autoref{prop:Feasiblecomplementarity} above.}

     Above, we have assumed no knowledge of the degrees of recommendation dependence, $\Delta_I$ and $\Delta_{II}$. However, in a case of partially unobserved outcomes $Y'$, it may be reasonable to also make assumptions about the partial absence of recommendation dependence. Specifically, if $Y$ is never observed when the safe action is taken (for example, if we never observe criminal behavior, $Y = \bad$, for a defendant who is jailed, $A = \safe$), then we may also hypothesize that the human decision-maker is less concerned with making a mistake against the recommendation in this case (e.g. by jailing an innocent person, $A = \safe, Y = \good$, against the release recommendation of the algorithm, $R = \risky$). This may be particularly applicable when recommendation dependence comes from institutional factors related to observed conduct and outcomes as in \autoref{subsec:Blamegame} (such as a fear of being blamed or persecuted for releasing a defendant who was deemed risky and commits a crime). We therefore now assume that $\Delta_I = 0$, so that unobserved outcomes do not lead to any distortions. In this case, we propose a minimax optimal recommendation algorithm that avoids recommending the safe decision (which would lead to distortions), and instead recommends risky decisions for those instances with a low probability of a bad outcome.

    \begin{prop}[Minimax recommendation with one-sided recommendation dependence]
     \label{prop:Onesidedminimax}
        The recommendation algorithm
         \begin{equation}
        \label{eqn:Onesidedtriage}
         \overline{\rec}''(x_m)
         =
         \begin{cases}
            \risky, & \overline{\P}'(Y'=\bad|X_m=x_m, A_0 = \risky) \leq p^*, \\
            \neutral, &\text{otherwise}
         \end{cases}
     \end{equation}
         minimizes worst-case loss
         $
             \textstyle\sup_{\P \in \mathcal{P}'(\overline{\P}'), \Delta_I = 0, \Delta_{II} \geq 0} \E[\ell(Y,A)]
         $
         over recommendation algorithms $\rec: \mathcal{X}_m \rightarrow \{ \risky,\neutral, \safe \}$,
         where $A$ denotes the recommendation-dependent choices of a human decision-maker following recommendations $R = \rec(X_m)$.
     \end{prop}

    Unlike the case where $\Delta_I > 0$, a minimax optimal recommendation algorithm may now also include recommending the risky option, since doing so does not induce any inefficiency. At the same time, the minimax optimal algorithms in this case of one-sided recommendation dependence are not unique, and also include the minimax algorithm $\overline{\rec}'$ from \autoref{prop:Unobservedminimax}, which we elaborate further on in \autoref{prop:Onesidedminimaxgeneral} of \autoref{apx:Minimax}. However, $\overline{\rec}''$ dominates  $\overline{\rec}'$ in the sense that it always achieves weakly lower expected loss by avoiding the safe action and thus not causing any recommendation dependence. It is worth noting that some risk assessments, such as the one detailed by \cite{albright2023hidden}, are explicitly implemented in such a one-sided manner when the decision-maker takes the \safe\ action too often. \autoref{prop:Onesidedminimax} justifies this choice further.

     The above solutions show how our framework can be applied in realistic cases where knowledge of the joint distribution is limited. These solutions are intuitive; they recommend actions in cases where these actions are sure to improve human decisions, while staying silent for ambiguous cases in which human performance is better or cannot be estimated from the data. These solutions are also readily implementable; they can be implemented on training data by solving an empirical risk minimization problem to estimate the relevant conditional expectations.
     They are also extendable to other assumptions; for example, if we also have additional knowledge about the degree of recommendation dependence, then we can solve the resulting optimization problem to come up with a minimax optimal algorithm.
     We conclude this section by solving for the minimax optimal recommendation in \autoref{exm1} only given knowledge of $\overline{\P}$.

        \begin{figure}
        \centering
        \centering
        \begin{subfigure}[b]{0.45\textwidth}
            \centering
            \begin{tikzpicture}
            \begin{axis}[
                axis equal image,
                xlabel={$X_h$}, xlabel style={yshift=15pt},
                ylabel={$X_m$}, ylabel style={yshift=-15pt},
                xmin=0, xmax=1,
                ymin=0, ymax=1,
                width=.75\textwidth,
                height=.75\textwidth,
                xtick={0,1},
                ytick={0,1},
                tick align=outside,
                enlargelimits=false,
                grid style=dashed,
                grid=both,
                clip=false,
                axis on top
            ]
            \fill [blue!10] (axis cs:0,0) -- (axis cs:0,1) -- (axis cs:1,0) -- cycle;
            \fill [red!10] (axis cs:0,1) -- (axis cs:1,1) -- (axis cs:1,0) -- cycle;

            \addplot[color=black, dashed] coordinates {(0,0.33) (1,0.33)};
            \addplot[color=black, dashed] coordinates {(0,0.67) (1,0.67)};
            \draw[->, color=black, ultra thick] (axis cs:0.2,.33) -- (axis cs:0.2,.24);
            \draw[->, color=black, ultra thick] (axis cs:0.7,.67) -- (axis cs:0.7,.76);

            \fill [red!70] (axis cs:0,1) -- (axis cs:0.096,1) -- (axis cs:0.096,0.904) -- cycle;
            \fill [blue!70] (axis cs:0.096,0.904) -- (axis cs:0.096,0.76) -- (axis cs:0.24,0.76) -- cycle;
            \fill [red!70] (axis cs:0.24,0.76) -- (axis cs:0.448,0.76) -- (axis cs:0.448,0.552) -- cycle;
            \fill [blue!70] (axis cs:0.448,.552) -- (axis cs:0.448,.24) -- (axis cs:.76,.24) -- cycle;
            \fill [red!70] (axis cs:.76,.24) -- (axis cs:.856,.24) -- (axis cs:.856,0.144) -- cycle;
            \fill [blue!70] (axis cs:.856,.144) -- (axis cs:.856,0) -- (axis cs:1,0) -- cycle;
            \addplot[color=black] coordinates {(0,1) (1,0)};                
            \addplot[color=black, ultra thick] coordinates {(0.096,0.76) (0.096,1)};
            \addplot[color=black, ultra thick] coordinates {(0.448,.76) (0.448,0.24)};
            \addplot[color=black, ultra thick] coordinates {(0.856,0) (.856,0.24)};
            \addplot[color=black, ultra thick, dashed] coordinates {(0,0.76) (1,0.76)};
            \addplot[color=black, ultra thick, dashed] coordinates {(0,0.24) (1,0.24)};
            \node at (axis cs:0.7,0.9) {$A=\safe$};
            \node at (axis cs:0.3,0.1) {$A=\risky$};
            \node[anchor=west] at (axis cs:1.05,.125) {$R=\risky$};
            \node[anchor=west] at (axis cs:1.05,0.5) {$R=\neutral$};
            \node[anchor=west] at (axis cs:1.05,0.875) {$R=\safe$};
            \node[color=red!70] at (axis cs:0.075,1.05) {$c_{II}$};

            \node[color=white] at (axis cs:0.15,0.8) {$c_I$};

            \node[color=white] at (axis cs:0.38,0.69) {$c_{II}$};
            \node[color=white] at (axis cs:0.55,0.34) {$c_I$};  

            \node[color=red!70] at (axis cs:.8,.29) {$c_{II}$};
            \node[color=white] at (axis cs:.91,0.04) {$c_I$};

            \end{axis}
        \end{tikzpicture}
            \caption{The minimax optimal algorithm uses recommendations conservatively. Actions shown for a recommendation-independent decision-maker.}
        \end{subfigure}
        \hspace{10pt}
        \begin{subfigure}[b]{0.45\textwidth}
            \centering 
            \begin{tikzpicture}
                \begin{axis}[
                    axis equal image,
                    xlabel={$p^*$}, xlabel style={yshift=15pt},
                    ylabel={$X_m$}, ylabel style={yshift=-15pt},
                    xmin=0, xmax=1,
                    ymin=0, ymax=1,
                    width=.75\textwidth,
                    height=.75\textwidth,
                    xtick={0,1},
                    ytick={0,1},
                    tick align=outside,
                    enlargelimits=false,
                    grid style=dashed,
                    grid=both,
                    clip=false,
                    axis on top,
                    domain=0:1,
                    samples=200
                    ]
                    \addplot[color=black, dashed] coordinates {(0,0.33) (1,0.33)};
                    \addplot[color=black, dashed] coordinates {(0,0.67) (1,0.67)};
                    \draw[->, color=black, ultra thick] (axis cs:0.5,.33) -- (axis cs:0.5,.25);
                    \draw[->, color=black, ultra thick] (axis cs:0.5,.67) -- (axis cs:0.5,.75);

                    \addplot[name path=lowercurve,black, thick, dashed] {x - x^2};
                    \node[anchor=south] at (axis cs:0.5,0.025) {$R=\risky$};
                    \node[anchor=south] at (axis cs:0.5,0.425) {$R=\neutral$};
                    \node[anchor=south] at (axis cs:0.5,0.825) {$R=\safe$};
    
                    \addplot[name path=uppercurve,black, thick, dashed] {x + (1 - x)^2};

                    \path[name path=bottomedge] (axis cs:0,0) -- (axis cs:1,0);

                    \path[name path=topedge] (axis cs:0,1) -- (axis cs:1,1);

                    \addplot [
                    red!30,
                    fill opacity=0.5
                    ] fill between [
                    of=lowercurve and bottomedge
                    ];

                    \addplot [
                    blue!30,
                    fill opacity=0.5
                    ] fill between [
                    of=topedge and uppercurve
                    ];
                \end{axis}
            \end{tikzpicture}
            \caption{The further $p^*$ is from $0.5$, the more conservatively recommendations are given. Unbalanced errors limit cases the agent's action can be safely improved.}
        \end{subfigure}
        \caption{Illustration of the minimax optimal recommendation for $p^* = .4$ (left) and the thresholds at which recommendations are sent as a function of $p^*$ (right) for \autoref{exm1}.}
        \label{fig:exm_minmax}
    \end{figure}

     \begin{exm}[name=Independent uniform signals,continues=exm1]
        In the example with uniform independent signals $X_h$ and $X_m$,
        we can show %
        that
        \begin{align*}
            \overline{\E}\left[\ell(Y,A_0)|X_m = x_m\right]
            =
            \E\left[\ell(Y,A_0)|X_m = x_m\right] %
            &= \max\left\{c_I((1-x_m) - p^*), c_{II} (x_m  - (1-p^*))\right\}.
        \end{align*}
        Here, $\E\left[\ell(Y,A_0)|X_m\right]\leq \E[\ell(Y,A_0^\dagger)|X_m]$ almost surely, so \autoref{prop:Triageminimax} gives the optimal minimax recommendation algorithm. By the definition of $\overline{\rec}_m$ as the optimal algorithmic decision,
        $$\overline{\E}\left[\ell\left(Y,\overline{\rec}_m(x_m)\right)\middle|X_m = x_m\right] = \E\left[\ell\left(Y,\overline{\rec}_m(x_m)\right)\middle|X_m = x_m\right] = \min\left\{c_I(1-x_m),c_{II}x_m\right\}.$$
        Comparing expected losses, the minimax optimal recommendation algorithm is
        $$R = \begin{cases}
            \risky, & X_m < p^* - (p^*)^2\\
            \neutral, &p^* - (p^*)^2 \leq X_m \leq p^* + (1-p^*)^2\\
            \safe, & X_m > p^* + (1-p^*)^2.
        \end{cases}$$
        \autoref{fig:exm_minmax} illustrates this algorithm, where Panel (a) shows the solution for our main example, where $p^* = .4$, and Panel (b) shows how the optimal thresholds vary with $c_I, c_{II}$ through $p^* = \frac{c_I}{c_I + c_{II}}$.
    \end{exm}

    \section{Extension to Implicit Recommendations with Strategic Silence}
    \label{sec:Implicit}

    So far, we have considered the explicit design of recommendations, where the only information the decision-maker receives from the algorithm is a discrete recommendation that explicitly suggests a course of action.
    In many applications, the human decision-maker may get access to a risk score provided by the algorithm.
    In this section, we therefore extend our model to assume that the information available to the decision-maker consists of their signal $X_h$ and a continuous machine prediction of the bad outcome occurring, such as the prediction $\P(Y{=}\bad|X_m)$.
    Recommendations are implicitly given by a risk score or withheld, $\mathcal{R} = [0,1] \cup \{ \withheld \}$.
    For example, a judge may receive an algorithmic prediction of a defendant committing a crime or failing to appear, and a doctor may obtain a risk score that expresses the probability that a patient has some medical condition.
    Throughout, we maintain the assumption that the agent takes the recommendation policy $\rec$ as given and understands the joint distribution of outcomes $Y$, private information $X_h$, and recommendations $R= \rec(X_m)$.

    We consider the consequences of recommendation-dependent preferences when recommendations come from machine risk scores.
    Such a recommendation may be explicit, such as when a judge obtains a probability score along with an explicit recommendation based on a probability threshold.
    Alternatively, the recommendation could be implicit, for example, when a doctor interprets a high-risk assessment as a recommendation to test.
    The former case could be captured by our model of explicit recommendations by assuming that the machine assessment becomes part of the signal $X_h$ available to the decision-maker. But in that case, our above results suggest that it is optimal from the perspective of the principal not to add any explicit recommendations, as they only distort decisions.
    Here, we instead focus on the latter case, where recommendation dependence is relative to the recommendation implicit in the machine's risk score.
    We consider a specific form of additional decision loss related to implicit recommendations given by        \begin{align*}
            \ell^*(Y,A,R)
            &=
            \ell(Y,A)
            +
            \begin{cases}
                \delta_I(\widehat{Y}), & Y{=}\good, A{=}\safe \\
                \delta_{II}(\widehat{Y}), & Y{=}\bad, A{=}\risky
            \end{cases}
            =
            \begin{cases}
                c_I + \delta_I(\widehat{Y}), & Y{=}\good, A{=}\safe \\
                c_{II}+ \delta_{II}(\widehat{Y}), & Y{=}\bad, A{=}\risky
            \end{cases}
        \end{align*}
        where $\widehat{Y} = R$ for $R \in [0,1]$ and $\widehat{Y} = \P(Y=\bad|R{=}\withheld)$ for $R = \withheld$,%
        \footnote{We could alternatively write $\widehat{Y} = \P(Y{=}\bad|R)$ for the \emph{implied} risk score, related to the sufficient-statistic approach of \cite{agarwal_designing_2025}.
        If $R$ represents risk scores $R = \P(Y{=}\bad|X_m)$ whenever $R \neq \withheld$, then the two approaches coincide.}
        where $\delta_I, \delta_{II}: [0,1] \rightarrow [0,\infty)$ fulfill $\delta_I(1) = 0 = \delta_{II}(0)$ with $\delta_I$ monotonically decreasing and $\delta_{II}$ monotonically increasing.
    Here, $\delta_I(\widehat{Y})$ and $\delta_{II}(\widehat{Y})$ represent additional (perceived) losses that come from reference effects through the implied risk assessment $\widehat{Y}$ when the decision-maker makes an error.
    We assume that these additional losses are larger the less likely the chosen action is according to the risk score (and are zero if the risk score implies that the chosen action is optimal).

    For example, we could recover losses similar to \autoref{sec:Model} if we assume that $\delta_I,\delta_{II}$ express recommendation dependence relative to the implied machine decision $A_m = \risky$ for $\widehat{Y} < p^* = \frac{c_I}{c_I + c_{II}}$ and $A_m = \safe$ for $\widehat{Y} > p^*$, in which case
    \begin{align}
        \label{eqn:Implicit}
        \delta_I(\widehat{Y})
        &= \Delta_I \: \Ind{\widehat{Y} < p^*},
        &
        \delta_{II}(\widehat{Y})
        &= \Delta_{II} \: \Ind{\widehat{Y} > p^*}.
    \end{align}
    In contrast to previous sections, the setup above also allows the magnitude of the predicted probability to matter for reference effects.
    For example, if we choose
    \begin{align}
        \label{eqn:Continuous}
        \delta_I(\widehat{Y})
        &= \Delta_I \: (1-\widehat{Y}),
        &
        \delta_{II}(\widehat{Y})
        &= \Delta_{II} \: \widehat{Y},
    \end{align}
    then the additional cost is proportional to the predicted probability of the corresponding adverse outcome: if the probability assessment suggests a high probability of the bad outcome occurring, then the cost of taking the risky action and encountering a bad outcome is higher than if the prediction suggests a low probability of the bad outcome.
    Despite the decision-maker now having access to a continuous algorithmic risk assessment, recommendation-dependent preferences still lead to inefficient choices because of over-adherence to the recommendation implicit in the probability assessment, and can lead to outcomes that are worse than a decision-maker deciding by themselves without any risk score or recommendation.
    The results from \autoref{sec:Implications} still apply.
    Specifically, if $\Delta_{II}$ is large and the machine prediction suggests a substantial probability of the bad outcome occurring, then the decision-maker will choose the safe action too often.

    When recommendations are directly tied to machine predictions, we may not be able to change recommendations explicitly. Instead, we consider in this section the merits of withholding the machine risk prediction itself to reduce distortions through recommendation dependence at the cost of a loss of information.
    Specifically, we assume that the machine assessment is now given by
    \begin{equation}
        \label{eqn:ImplicitThreshold}
        R
        =
        \begin{cases}
            \P(Y{=}\bad|X_m), & \P(Y{=}\bad|X_m) \notin [p^\low,p^\high],\\
            \withheld, & \P(Y{=}\bad|X_m) \in [p^\low,p^\high].
        \end{cases}
    \end{equation}
    That is, the algorithm withholds a score when it is intermediate (and thus may have limited helpful information about the optimal action).%
    \footnote{Such simple threshold rules are not necessarily optimal. However, more complex policies may not be understood by human decision-makers, and such threshold rules represent a natural starting point.}
    In our setup, we assume that the withheld risk score affects the decision-maker's preferences equivalently to a risk assessment $\P(Y{=}\bad|\P(Y{=}\bad|X_m) \in [p^\low,p^\high])$.%
    \footnote{We could alternatively assume that there is no recommendation dependence when the risk prediction is withheld, but this assumption may be unrealistic when a withheld risk score signals a particularly high or low risk score.}
    The risk assessment thus represents a coarsening of the full prediction $\P(Y{=}\bad|X_m)$ that loses information about variations in risk scores between $p^\low$ and $p^\high$ \citep[similar to][]{hoong_improving_2025}.
    Despite losing information, withholding information strategically in this way can improve outcomes in the presence of recommendation dependence.

    Having discussed the idea that withholding the score strategically can improve outcomes, note that an analog of \autoref{prop:Complementarity} holds for the case of continuous risk scores.
    Specifically, we now provide conditions under which we can find a scoring rule of the form \eqref{eqn:ImplicitThreshold} that always (weakly) improves over machine-only and human-only decisions.
    To formulate our result, we call a risk value $\widehat{Y} \in [0,1]$ \emph{recommendation-neutral} if it does not imply any recommendation dependence, that is, if $\frac{\delta_I(\widehat{Y})}{c_I} = \frac{\delta_{II}(\widehat{Y})}{c_{II}}$.
    For example, $\widehat{Y} = p^* = \frac{c_I}{c_I + c_{II}}$ is recommendation-neutral for the specification \eqref{eqn:Implicit} and also for \eqref{eqn:Continuous} if $\Delta_I / c_I = \Delta_{II} / c_{II}$.

    \begin{prop}[Human--machine complementarity from destroying information]
    \label{prop:Delete}
    If $p^* = \frac{c_I}{c_I + c_{II}}$ is recommendation-neutral, then there is a risk score of the form \eqref{eqn:ImplicitThreshold} (with $p^\low \leq p^* \leq p^\high$) that (weakly) improves over the best machine-only decision,
    \(
        \E[\ell(Y,A)] \leq \E[\ell(Y,A_m)].
    \)
    If the overall probability $\P(Y{=}\bad)$ is recommendation-neutral, then there is a risk score of the form \eqref{eqn:ImplicitThreshold} that (weakly) improves over the best human-only decision,
    \(
        \E[\ell(Y,A)] \leq \E[\ell(Y,A_0)].
    \)
    In both cases, the risk score can be chosen to weakly improve over always providing $\P(Y{=}\bad|X_m)$ (i.e., not withholding the risk score).
    \end{prop}
    
    We provide a simple example of such an improvement in \autoref{apx:Continuous}.

    This approach adapts the idea of Bayesian persuasion \citep{kamenica_bayesian_2011} to our context: by changing the structure of the information and coarsening the signal strategically, the designer of the algorithm can improve outcomes through increasing the alignment between their goal and the misaligned choices of the decision-maker. However, unlike the baseline Bayesian persuasion case, the signal structure affects the preferences themselves through the implied recommendations.

    We note that unlike the setting from \autoref{sec:NonRec}, where adding a neutral option \textit{added} information, this modification of the risk assessment strictly \textit{decreases} the information given by the machine.
    In the rational baseline of no recommendation dependence ($\Delta_I = 0 =\Delta_{II}$), this modification would strictly worsen outcomes.
    Yet in the recommendation-dependent case, there is room for net improvements through (strategic) silence about the risk score.

    \section{Conclusion}
    \label{sec:Conclusion}

    When we provide a decision-maker with a recommendation, they may not only react to its information content, but also see it as a default action that affects their preferences.
    In this article, we illustrate in a simple example and with general results how recommendation-dependent preferences create inefficiencies and affect the design of optimal recommendations.
    Our model suggests practically implementable modifications that reduce distortions by strategically altering or even withholding recommendations for instances where they may otherwise hurt more than they help.
    With our work, we aim to provide an example of the integration of more realistic models of human behavior into the design of algorithms, and hope that it can contribute to improving human--AI interaction in critical applications.

    Our model leaves room for relevant extensions.
    First, we have assumed throughout that the decision-maker interprets recommendations correctly. However, in practice, the decision-maker may have a hypothesis about the recommendation that may not be fully accurate, or the decision-maker may have limited cognitive capacity to work with complex signals.
    An extension to a limited understanding by the decision-maker may provide more realistic prescriptions for those cases.

    Second, we have assumed throughout that the decision-maker and the algorithm designer agree on the baseline costs of making errors, and only differ with respect to recommendation-dependent losses of the decision-maker. If the baseline preferences are already misaligned, recommendation dependence may improve decisions by increasing adherence to the preferred action of the algorithm designer, even if it comes at the cost of reducing revealed information.

    Finally, reference points are hardly influenced by recommendations alone, and the sequencing and framing of human--machine interactions may have first-order effects on the efficiency of human choices. One remedy to the inefficiencies we discuss in this article could, for example, be interventions that elicit information from the human decision-maker first to avoid anchoring on machine recommendations. Specifically, when human decision-makers have valuable information and the quality of algorithmic predictions is limited, our theory suggests that anchoring the decision-maker in human rather than algorithmic reference points may improve overall decision quality.

    \bibliography{_behav}

\begin{thebibliography}{}

\bibitem[Agarwal, Moehring, and Wolitzky(2025)]{agarwal_designing_2025}
Agarwal, Nikhil, Alex Moehring, and Alexander Wolitzky (2025).
\newblock Designing {Human}-{AI} {Collaboration}: {A} {Sufficient}-{Statistic} {Approach}.

\bibitem[Albright(2023)]{albright2023hidden}
Albright, Alex (2023).
\newblock The hidden effects of algorithmic recommendations.

\bibitem[Alur, Raghavan, and Shah(2024)]{alur_human_2024}
Alur, Rohan, Manish Raghavan, and Devavrat Shah (2024).
\newblock Human {Expertise} in {Algorithmic} {Prediction}.

\bibitem[Angelova, Dobbie, and Yang(2023)]{angelova_algorithmic_2023}
Angelova, Victoria, Will Dobbie, and Crystal Yang (2023).
\newblock Algorithmic {Recommendations} and {Human} {Discretion}.
\newblock Technical Report w31747, National Bureau of Economic Research, Cambridge, MA.

\bibitem[Bal(2009)]{bal_introduction_2009}
Bal, B.~Sonny (2009).
\newblock An {Introduction} to {Medical} {Malpractice} in the {United} {States}.
\newblock {\em Clinical Orthopaedics and Related Research}, 467(2):339--347.

\bibitem[Banker and Khetani(2019)]{banker2019algorithm}
Banker, Sachin and Salil Khetani (2019).
\newblock Algorithm overdependence: How the use of algorithmic recommendation systems can increase risks to consumer well-being.
\newblock {\em Journal of Public Policy \& Marketing}, 38(4):500--515.

\bibitem[Bansal et~al.(2019)]{bansal_updates_2019}
Bansal, Gagan, Besmira Nushi, Ece Kamar, Daniel~S. Weld, Walter~S. Lasecki, and Eric Horvitz (2019).
\newblock Updates in {Human}-{AI} {Teams}: {Understanding} and {Addressing} the {Performance}/{Compatibility} {Tradeoff}.
\newblock {\em Proceedings of the AAAI Conference on Artificial Intelligence}, 33(01):2429--2437.

\bibitem[Barberis(2013)]{barberis2013thirty}
Barberis, Nicholas~C (2013).
\newblock Thirty years of prospect theory in economics: A review and assessment.
\newblock {\em Journal of Economic Perspectives}, 27(1):173--96.

\bibitem[Bastani, Bastani, and Sinchaisri(2021)]{bastani2021improving}
Bastani, Hamsa, Osbert Bastani, and Wichinpong~Park Sinchaisri (2021).
\newblock Improving human decision-making with machine learning.
\newblock {\em arXiv:2108.08454}.

\bibitem[Baucells, Weber, and Welfens(2011)]{baucells_reference-point_2011}
Baucells, Manel, Martin Weber, and Frank Welfens (2011).
\newblock Reference-{Point} {Formation} and {Updating}.
\newblock {\em Management Science}, 57(3):506--519.

\bibitem[Bell(1982)]{bell_regret_1982}
Bell, David~E. (1982).
\newblock Regret in {Decision} {Making} under {Uncertainty}.
\newblock {\em Operations Research}, 30(5):961--981.

\bibitem[Bhatia and Golman(2019)]{bhatia_attention_2019}
Bhatia, Sudeep and Russell Golman (2019).
\newblock Attention and reference dependence.
\newblock {\em Decision}, 6(2):145--170.

\bibitem[Bondi et~al.(2022)]{bondi_role_2022}
Bondi, Elizabeth, Raphael Koster, Hannah Sheahan, Martin Chadwick, Yoram Bachrach, Taylan Cemgil, Ulrich Paquet, and Krishnamurthy Dvijotham (2022).
\newblock Role of {Human}-{AI} {Interaction} in {Selective} {Prediction}.
\newblock {\em Proceedings of the AAAI Conference on Artificial Intelligence}, 36(5):5286--5294.

\bibitem[Choi et~al.(2004)]{choi2004better}
Choi, James~J, David Laibson, Brigitte~C Madrian, and Andrew Metrick (2004).
\newblock For better or for worse: Default effects and 401 (k) savings behavior.
\newblock In {\em Perspectives on the Economics of Aging}, pages 81--126. University of Chicago Press.

\bibitem[Dai and Singh(2025)]{dai_express:_2025}
Dai, Tinglong and Shubhranshu Singh (2025).
\newblock {Artificial} {Intelligence} on {Call}: {The} {Physician}'s {Decision} of {Whether} to {Use} {AI} in {Clinical} {Practice}.
\newblock {\em Journal of Marketing Research}, page 00222437251332898.

\bibitem[Diecidue and Somasundaram(2017)]{diecidue_regret_2017}
Diecidue, Enrico and Jeeva Somasundaram (2017).
\newblock Regret theory: {A} new foundation.
\newblock {\em Journal of Economic Theory}, 172:88--119.

\bibitem[Doval and Smolin(2023)]{doval_persuasion_2023}
Doval, Laura and Alex Smolin (2023).
\newblock Persuasion and {Welfare}.
\newblock arXiv:2109.03061.

\bibitem[Esthappan(2024)]{esthappan_assessing_2024}
Esthappan, Sino (2024).
\newblock Assessing the {Risks} of {Risk} {Assessments}: {Institutional} {Tensions} and {Data} {Driven} {Judicial} {Decision}-{Making} in {U}.{S}. {Pretrial} {Hearings}.
\newblock {\em Social Problems}, page spae060.

\bibitem[Fogliato et~al.(2022)]{fogliato_who_2022}
Fogliato, Riccardo, Shreya Chappidi, Matthew Lungren, Paul Fisher, Diane Wilson, Michael Fitzke, Mark Parkinson, Eric Horvitz, Kori Inkpen, and Besmira Nushi (2022).
\newblock Who {Goes} {First}? {Influences} of {Human}-{AI} {Workflow} on {Decision} {Making} in {Clinical} {Imaging}.
\newblock In {\em 2022 {ACM} {Conference} on {Fairness}, {Accountability}, and {Transparency}}, {FAccT} '22, pages 1362--1374.

\bibitem[Froomkin, Kerr, and Pineau(2019)]{froomkin_when_2019}
Froomkin, A.~Michael, Ian Kerr, and Joelle Pineau (2019).
\newblock When {AIs} {Outperform} {Doctors}: {Confronting} the {Challenges} of a {Tort}-{Induced} over-{Reliance} on {Machine} {Learning}.
\newblock {\em Arizona Law Review}, 61:33.

\bibitem[Fügener et~al.(2021)]{fugener_will_2021}
Fügener, Andreas, Jörn Grahl, Alok Gupta, and Wolfgang Ketter (2021).
\newblock Will {Humans}-in-{The}-{Loop} {Become} {Borgs}? {Merits} and {Pitfalls} of {Working} with {AI}.
\newblock {\em SSRN 3879937}.

\bibitem[Green and Chen(2019)]{green2019principles}
Green, Ben and Yiling Chen (2019).
\newblock The principles and limits of algorithm-in-the-loop decision making.
\newblock {\em Proceedings of the ACM on Human-Computer Interaction}, 3(CSCW):1--24.

\bibitem[Guttman-Kenney et~al.(2023)]{guttman2023semblance}
Guttman-Kenney, Benedict, Paul~D Adams, Stefan Hunt, David Laibson, Neil Stewart, and Jesse Leary (2023).
\newblock The semblance of success in nudging consumers to pay down credit card debt.
\newblock Technical report, National Bureau of Economic Research.

\bibitem[Hampshire et~al.(2020)]{hampshire_beyond_2020}
Hampshire, Robert~C., Shan Bao, Walter~S. Lasecki, Andrew Daw, and Jamol Pender (2020).
\newblock Beyond safety drivers: {Applying} air traffic control principles to support the deployment of driverless vehicles.
\newblock {\em PLOS ONE}, 15(5):e0232837.

\bibitem[Hemmer et~al.(2021)]{hemmer_human-ai_2021}
Hemmer, Patrick, Max Schemmer, Michael Vössing, and Niklas Kühl (2021).
\newblock Human-{AI} {Complementarity} in {Hybrid} {Intelligence} {Systems}: {A} {Structured} {Literature} {Review}.
\newblock In {\em {PACIS} 2021 {Proceedings}}.

\bibitem[Hoong and Dreyfuss(2025)]{hoong_improving_2025}
Hoong, Ruru and Bnaya Dreyfuss (2025).
\newblock Improving {AI}-{Assisted} {Decision}-{Making} {Through} {Calibrated} {Coarsening}.

\bibitem[IAALS(2022)]{IAALS_judicial_2022}
IAALS (2022).
\newblock Judicial {Performance} {Evaluation} 2.0.
\newblock {\em Institute for the Advancement of the American Legal System}.

\bibitem[Ibrahim, Kim, and Tong(2021)]{ibrahim_eliciting_2021}
Ibrahim, Rouba, Song-Hee Kim, and Jordan Tong (2021).
\newblock Eliciting {Human} {Judgment} for {Prediction} {Algorithms}.
\newblock {\em Management Science}, 67(4):2314--2325.

\bibitem[Imai et~al.(2020)]{Imai2020-ci}
Imai, Kosuke, Zhichao Jiang, James Greiner, Ryan Halen, and Sooahn Shin (2020).
\newblock {Experimental Evaluation of Algorithm-Assisted Human Decision-Making: Application to Pretrial Public Safety Assessment}.
\newblock {\em arXiv:2012.02845}.

\bibitem[Kahneman and Tversky(1979)]{kahneman1979prospect}
Kahneman, Daniel and Amos Tversky (1979).
\newblock Prospect theory: An analysis of decision under risk.
\newblock {\em Econometrica}, 47(2):263--292.

\bibitem[Kamenica and Gentzkow(2011)]{kamenica_bayesian_2011}
Kamenica, Emir and Matthew Gentzkow (2011).
\newblock Bayesian {Persuasion}.
\newblock {\em American Economic Review}, 101(6):2590--2615.

\bibitem[K\H{o}szegi and Rabin(2006)]{koszegi_model_2006}
K\H{o}szegi, B. and M.~Rabin (2006).
\newblock A {Model} of {Reference}-{Dependent} {Preferences}.
\newblock {\em The Quarterly Journal of Economics}, 121(4):1133--1165.

\bibitem[Kıbrıs, Masatlioglu, and Suleymanov(2023)]{kibris_theory_2023}
Kıbrıs, Özgür, Yusufcan Masatlioglu, and Elchin Suleymanov (2023).
\newblock A theory of reference point formation.
\newblock {\em Economic Theory}, 75(1):137--166.

\bibitem[Lai et~al.(2021)]{lai_towards_2021}
Lai, Vivian, Chacha Chen, Q.~Vera Liao, Alison Smith-Renner, and Chenhao Tan (2021).
\newblock Towards a {Science} of {Human}-{AI} {Decision} {Making}: {A} {Survey} of {Empirical} {Studies}.
\newblock {\em arXiv:2112.11471}.

\bibitem[Lakkaraju and Bastani(2020)]{LakkarajuB20}
Lakkaraju, Himabindu and Osbert Bastani (2020).
\newblock {"How do {I} fool you?": Manipulating User Trust via Misleading Black Box Explanations}.
\newblock In Markham, Annette~N., Julia Powles, Toby Walsh, and Anne~L. Washington, editors, {\em {AIES} '20: {AAAI/ACM} Conference on AI, Ethics, and Society, New York, NY, USA, February 7-8, 2020}, pages 79--85.

\bibitem[Lawrence et~al.(2006)]{lawrence_judgmental_2006}
Lawrence, Michael, Paul Goodwin, Marcus O'Connor, and Dilek Önkal (2006).
\newblock Judgmental forecasting: {A} review of progress over the last 25 years.
\newblock {\em International Journal of Forecasting}, 22(3):493--518.

\bibitem[Loomes and Sugden(1982)]{loomes_regret_1982}
Loomes, Graham and Robert Sugden (1982).
\newblock Regret {Theory}: {An} {Alternative} {Theory} of {Rational} {Choice} {Under} {Uncertainty}.
\newblock {\em The Economic Journal}, 92(368):805.

\bibitem[McGrath et~al.(2020)]{mcgrath_when_2020}
McGrath, Sean, Parth Mehta, Alexandra Zytek, Isaac Lage, and Himabindu Lakkaraju (2020).
\newblock When {Does} {Uncertainty} {Matter}? {Understanding} the {Impact} of {Predictive} {Uncertainty} in {ML} {Assisted} {Decision} {Making}.
\newblock {\em arXiv:2011.06167}.

\bibitem[McLaughlin and Spiess(2024)]{mclaughlin2024designing}
McLaughlin, Bryce and Jann Spiess (2024).
\newblock Designing algorithmic recommendations to achieve human-ai complementarity.
\newblock {\em arXiv preprint arXiv:2405.01484}.

\bibitem[Mozannar and Sontag(2021)]{mozannar_consistent_2021}
Mozannar, Hussein and David Sontag (2021).
\newblock Consistent {Estimators} for {Learning} to {Defer} to an {Expert}.
\newblock arXiv:2006.01862.

\bibitem[Noti and Chen(2022)]{noti_learning_2022}
Noti, Gali and Yiling Chen (2022).
\newblock Learning {When} to {Advise} {Human} {Decision} {Makers}.

\bibitem[Palley and Soll(2019)]{palley_extracting_2019}
Palley, Asa~B and Jack~B Soll (2019).
\newblock Extracting the wisdom of crowds when information is shared.
\newblock {\em Management Science}, 65(5):2291--2309.

\bibitem[Peng, Garg, and Kleinberg(2024)]{peng_no_2024}
Peng, Kenny, Nikhil Garg, and Jon Kleinberg (2024).
\newblock A {No} {Free} {Lunch} {Theorem} for {Human}-{AI} {Collaboration}.
\newblock arXiv:2411.15230.

\bibitem[Raghu et~al.(2019)]{raghu_algorithmic_2019}
Raghu, Maithra, Katy Blumer, Greg Corrado, Jon Kleinberg, Ziad Obermeyer, and Sendhil Mullainathan (2019).
\newblock The {Algorithmic} {Automation} {Problem}: {Prediction}, {Triage}, and {Human} {Effort}.
\newblock {\em arXiv:1903.12220}.

\bibitem[Shashikumar et~al.(2021)]{shashikumar2021artificial}
Shashikumar, Supreeth~P, Gabriel Wardi, Atul Malhotra, and Shamim Nemati (2021).
\newblock {Artificial intelligence sepsis prediction algorithm learns to say “I don’t know”}.
\newblock {\em NPJ Digital Medicine}, 4(1):1--9.

\bibitem[Snyder, Keppler, and Leider(2022)]{snyder_algorithm_2022}
Snyder, Clare, Samantha Keppler, and Stephen Leider (2022).
\newblock Algorithm {Reliance} {Under} {Pressure}: {The} {Effect} of {Customer} {Load} on {Service} {Workers}.
\newblock {\em SSRN 4066823}.

\bibitem[Stevenson and Doleac(2019)]{stevenson_algorithmic_2019}
Stevenson, Megan~T and Jennifer~L Doleac (2019).
\newblock Algorithmic {Risk} {Assessment} in the {Hands} of {Humans}.
\newblock {\em SSRN 3489440}.

\bibitem[Steyvers et~al.(2022)]{steyvers_bayesian_2022}
Steyvers, Mark, Heliodoro Tejeda, Gavin Kerrigan, and Padhraic Smyth (2022).
\newblock Bayesian modeling of human–{AI} complementarity.
\newblock {\em Proceedings of the National Academy of Sciences}, 119(11):e2111547119.

\bibitem[Straitouri et~al.(2023)]{straitouri_improving_2023}
Straitouri, Eleni, Lequn Wang, Nastaran Okati, and Manuel~Gomez Rodriguez (2023).
\newblock Improving {Expert} {Predictions} with {Conformal} {Prediction}.
\newblock arXiv:2201.12006.

\bibitem[Sun et~al.(2022)]{sun_predicting_2022}
Sun, Jiankun, Dennis~J. Zhang, Haoyuan Hu, and Jan~A. Van~Mieghem (2022).
\newblock Predicting {Human} {Discretion} to {Adjust} {Algorithmic} {Prescription}: {A} {Large}-{Scale} {Field} {Experiment} in {Warehouse} {Operations}.
\newblock {\em Management Science}, 68(2):846--865.

\bibitem[Taudien et~al.(2022)]{taudien_effect_2022}
Taudien, Anna, Andreas F{\"u}gener, Alok Gupta, and Wolfgang Ketter (2022).
\newblock {The Effect of AI Advice on Human Confidence in Decision-Making}.
\newblock In {\em Proceedings of the 55th Hawaii International Conference on System Sciences}.

\bibitem[Toni et~al.(2024)]{toni_towards_2024}
Toni, Giovanni~De, Nastaran Okati, Suhas Thejaswi, Eleni Straitouri, and Manuel Gomez-Rodriguez (2024).
\newblock Towards {Human}-{AI} {Complementarity} with {Prediction} {Sets}.
\newblock arXiv:2405.17544.

\bibitem[Vaccaro, Almaatouq, and Malone(2024)]{vaccaro_when_2024}
Vaccaro, Michelle, Abdullah Almaatouq, and Thomas Malone (2024).
\newblock When combinations of humans and {AI} are useful: {A} systematic review and meta-analysis.
\newblock {\em Nature Human Behaviour}, 8(12):2293--2303.

\bibitem[Vodrahalli, Gerstenberg, and Zou(2022)]{vodrahalli_uncalibrated_2022}
Vodrahalli, Kailas, Tobias Gerstenberg, and James Zou (2022).
\newblock Uncalibrated {Models} {Can} {Improve} {Human}-{AI} {Collaboration}.
\newblock arXiv:2202.05983.

\end{thebibliography}

    \newpage
    \appendix
    \numberwithin{prop}{section}
    \numberwithin{asm}{section}
    \numberwithin{equation}{section}

    \section{Microfoundation from Gain--Loss Preferences}
    \label{apx:gainloss}

    In \autoref{subsec:Refdep}, we have shown that loss aversion relative to a reference point set by the \emph{actual} loss $\ell(Y,R)$ from implementing the recommendation leads to recommendation-dependent preferences of the form in \eqref{eqn:Recdeppref}.
    We now extend this idea to gain--loss preferences relative to a reference point set by expected losses or the distribution of errors from implementing the recommendation.

    We adopt a standard model of gain--loss utility from \cite{koszegi_model_2006} with additively separable gain--loss utility stemming from different components of outcomes.
    Specifically, we assume that the decision-maker thinks of type-I and type-II errors separately, experiencing losses
        \begin{align*}
        \ell_I(Y,A) &= \Ind{Y{=}\good,A{=}\safe} \: c_I,
        &
        \ell_{II}(Y,A) &= \Ind{Y{=}\bad,A{=}\risky} \: c_{II}.
    \end{align*}
    We write $L_I, L_{II}$ for reference losses for $\ell_I(Y,A), \ell_{II}(Y,A)$, where $L_I, L_{II}$ can be random variables that can be correlated with $Y,R,X_h$.
    This setup will later capture reference points related to the recommendation that can take the form of expected losses, a lottery over type-I and type-II losses, or realized losses.

    Following \cite{koszegi_model_2006}, we assume that reference-dependent loss is calculated separately across the two components and added up, with total reference-dependent loss from taking the decision $A$ relative to outcomes $Y$ and reference losses $L_I, L_{II}$ taking the form
    \begin{align*}
        \ell(Y,A|L_I,L_{II})
        &=
        \begin{cases}
            \lambda (\ell_I(Y,A) - L_I), &  \ell_I(Y,A) > L_I \\
            \ell_I(Y,A) - L_I, &  \ell_I(Y,A) \leq L_I
        \end{cases}
        +
        \begin{cases}
            \lambda (\ell_{II}(Y,A) - L_{II}), &  \ell_{II}(Y,A) > L_{II} \\
            \ell_{II}(Y,A) - L_{II}, &  \ell_{II}(Y,A) \leq L_{II}
        \end{cases}
    \end{align*}
    for $\lambda \geq 1$.
    This loss expresses that the decision-maker evaluates each loss component against the relevant reference point.
    If the loss from the decision $A$ exceeds the reference loss, it is experienced more strongly according to $\lambda$.
    Since this logic is applied separately for type-I and type-II errors, the agent still experiences partial loss aversion when the decision creates a worse type-I loss, while leading to less type-II loss.

    We now consider three natural reference points affected by the recommendation $R$:
    \begin{enumerate}
        \item The reference point is set by the realized loss from the recommendation, $L_I  = \ell_I(Y,R), L_{II} = \ell_{II}(Y,R)$. This expresses regret relative to the counterfactual course of action of adopting the recommendation.
        \item The reference point is set by the expected loss from the recommendation given the agent's information, $L_I  = \E[\ell_I(Y,R)|X_h,R], L_{II} =  \E[\ell_{II}(Y,R)|X_h,R]$. This expresses Prospect-Theory-type reference-dependent loss relative to the expected loss of adopting the recommendation.
        \item The reference point is set by a lottery over errors from adopting the recommendation, that is,
        $L_I  = \ell_I(Y',R), L_{II} = \ell_{II}(Y',R)$ with $Y' \stackrel{d}{=} Y | R,X_h$ and $Y'$, $Y$ independent given $R,X_h$. This implements the lottery-type reference points in \cite{koszegi_model_2006}.
    \end{enumerate}
    As an extension, we could also consider a reference point of the latter kind that is set in personal or preferred personal equilibrium following \cite{koszegi_model_2006}, where we would consider the distribution over losses from a recommendation-assisted human decision rather than a direct implementation of the recommendation as a reference point.

    In case 1., we directly find that
    \begin{align*}
        \ell(Y,A|L_I,L_{II}) = \ell^{\text{LA}}(Y,A,R)
    \end{align*}
    as in \autoref{prop:PT}, so that the result still applies and yields recommendation-dependent preferences with $\Delta_I = (\lambda - 1) c_I, \Delta_{II} = (\lambda - 1) c_{II}$.
    The second and third cases similarly yield recommendation-dependent choices:

    \begin{prop}[Reference dependence implies recommendation dependence]
    \label{prop:refdep}
        For reference points set as in 2.\ or 3.\ above, an agent minimizing expected reference-dependent loss $\E[\ell(Y,A|L_I,L_{II})]$ behaves according to recommendation-dependent preferences of the form \eqref{eqn:Recdeppref} with $\Delta_I = \Delta_I(\lambda),\Delta_{II} = \Delta_{II}(\lambda) \geq 0$ where $\Delta_I(\lambda), \Delta_{II}(\lambda)$ are strictly increasing in $\lambda \geq 1$ with $\Delta_I(1) = 0 = \Delta_{II}(1)$.
        For the special case of $c_I = 1 = c_{II}$, we have that
        $\Delta_I(\lambda) = \sqrt{\lambda} -  1= \Delta_{II}(\lambda)$.
    \end{prop}

    In particular, both specifications 2.\ and 3.\ imply the same $\Delta_I, \Delta_{II}$.

    \section{Structure of Optimal Recommendations}
    \label{apx:Design}

    In this section, we discuss the structure of optimal discrete recommendations.
    We first consider binary recommendations ($\mathcal{R} = \{\safe,\risky\}$), before moving to recommendations with a neutral level ($\mathcal{R} = \{\safe,\risky,\neutral\}$).
    Throughout, we assume that human and machine signals can be written as a combination of a jointly known context and independent private signals.

    \begin{asm}[Separable signals]
        \label{asm:independent}
        We have $X_m = (Z_0,Z_m), X_h = (Z_0,Z_h)$ with $Z_0, Z_m, Z_h$ independent.
    \end{asm}

    Next, we assume that, conditional on the common context $Z_0$, the private information of human and machine can each be summarized by a scalar-valued index.

    \begin{asm}[Scalar index representation]
        \label{asm:index}
        There are measurable scalar-valued functions $\phi, \phi_h,\phi_m$ such that a.s.\ $\P(Y{=}\bad|Z_h,Z_m;Z_0) = \phi(\phi_h(Z_h;Z_0), \phi_m(Z_m;Z_0);Z_0)$.
    \end{asm}

    This assumption means that the signals $Z_0, \phi_h(Z_h;Z_0), \phi_m(Z_m;Z_0)$ are sufficient statistics for $Y$.
    This assumption allows us to express optimal strategies of the principal and the agent in terms of these simple indices only.
    Finally, we restrict the relationship of these two indices and the probability of a bad outcome to be monotonic, meaning that a larger value of the index corresponds to a larger probability of the bad outcomes.

    \begin{asm}[Monotonicity]
        \label{asm:monotonicity}
        The function $\phi(\cdot,\cdot;Z_0)$ is monotonically increasing in both arguments, given $Z_0$.
    \end{asm}

    This assumption allows us to relate the ordinal information in the indices to a ranking of probabilities.
    Together, these three assumptions imply that both optimal decision and optimal recommendations can be written as threshold rules, conditional on the common context $Z_0$.
    We start with a general result on optimal decisions given the recommendation algorithm, where for simplicity we continue to resolve ties in favor of the risky decision.

    \begin{prop}[Threshold decisions]
        \label{prop:Agent-Threshold}
        Under Assumptions~\ref{asm:independent}--\ref{asm:monotonicity}, and given any recommendation policy $R = \rec(X_m)$, the agent's optimal decision is almost surely equal to
        \[
            A
            =
            \begin{cases}
                \risky, & \P(Y{=}\bad|X_h) \leq h^R(Z_0),
                \\
                \safe, & \P(Y{=}\bad|X_h) > h^R(Z_0)
            \end{cases}
        \]
        for some threshold functions $h^{\risky}(Z_0)$ and $h^{\safe}(Z_0)$ that vary only with the common context $Z_0$.
    \end{prop}

    This result says that the human decision after receiving a recommendation has a similar structure to unassisted decisions: the agent compares the best prediction of the bad outcome occurring using their information $(X_h)$, and takes the risky decision only if that probability is low.
    However, the probability threshold to decide between risky and safe actions now depends on the recommendation $R$ (as well as the common context $Z_0$, which may be required to interpret the recommendation).
    This is in contrast to the unassisted case, for which the threshold is simply $p^* = \frac{c_I}{c_I + c_{II}}$.

    While the above representation holds for any recommendation policy, we now specifically consider recommendations that can similarly be written as a threshold rule of the best machine prediction $\P(Y{=}\bad|X_m)$,
    that is,
    \begin{equation}
        \label{eqn:Principal-Threshold}
        R
        =
        \begin{cases}
            \risky, & \P(Y{=}\bad|X_m) \leq m(Z_0),
            \\
            \safe, & \P(Y{=}\bad|X_m) > m(Z_0).
        \end{cases}
    \end{equation}
    The class of these recommendations includes recommending the decision that the algorithm would take, in which case the threshold would simply be $p^* = \frac{c_I}{c_I + c_{II}}$.%
    \footnote{While such threshold rules are optimal for decisions, they are not generally optimal for recommendations, and we may theoretically be able to do better by allowing for more complex mapping between machine information and recommendation.
    However, we think that simple threshold rules are realistic restrictions in many cases and may be better understood by a human decision-maker than more complex rules. We therefore focus on optimal thresholds. Solving for optimal recommendation rules more generally (under realistic transparency restrictions) could be a promising direction for future research.}
    As a consequence, we can describe recommendation algorithms and resulting decisions in terms of the thresholds they imply on $\P(Y{=}\bad|X_h)$ and $\P(Y{=}\bad|X_m)$, respectively. 

    In \eqref{eqn:Principal-Threshold}, machine private information $Z_m$ affects the outcomes given by the machine by changing the best prediction $\P(Y{=}\bad|X_m) = \P(Y{=}\bad|Z_m,Z_0)$ of the bad outcome occurring, which is compared to the threshold $m(Z_0)$.
    Hence, the machine recommendation only changes with the information $Z_m$ through the prediction $\P(Y{=}\bad|Z_m,Z_0)$.
    In the case where there is no common context $Z_0$, this would imply that machine recommendations are given by putting a fixed threshold on $\P(Y{=}\bad|Z_m)$, as is the case in \autoref{exm1}.
    At the same time, we allow the common information $Z_0$ to affect the threshold to account for differences in the joint distribution of $(Y,Z_m,Z_h)$ that are known to the designer and human decision-maker.
    
    We now show that the insights from the example generalize to other cases for which the above assumptions hold.
    We start by noting that the optimal thresholds for the agent taking the risky decision are higher for a risky recommendation and lower for a safe recommendation, with the threshold the agent would choose absent a recommendation being in-between the two.

    \begin{prop}[Optimal agent thresholds]
        \label{prop:Agent-Thresholdordering}
        Assume that Assumptions~\ref{asm:independent}--\ref{asm:monotonicity} hold, and that the principal's threshold policy $m(Z_0)$ is optimal.
        Then, for any $\Delta_I,\Delta_{II} \geq 0$, we can choose thresholds in \autoref{prop:Agent-Threshold} such that
        \[
            h^{\risky}(Z_0)
            \geq
            p^*
            \geq
            h^{\safe}(Z_0)
        \]
        where we note that $p^* = \frac{c_I}{c_I + c_{II}}$ is the threshold the agent could choose if they chose an action directly, without a recommendation.
    \end{prop}

    Next, we consider how the optimal policy of the agent changes as the degree of recommendation dependence changes.
    As in the example, we find that an increasing level of recommendation dependence leads to thresholds that make the recommended action more likely to be taken.
    Also, decreasing the threshold of the algorithm means that the agent thresholds both increase to compensate for a lower implied probability of the bad outcome occurring.

    \begin{prop}[Change in optimal agent thresholds]
        \label{prop:Agent-Thresholdcomparative}
        Assume that Assumptions~\ref{asm:independent}--\ref{asm:monotonicity} hold.
        Then we can choose thresholds in \autoref{prop:Agent-Threshold} across values of $\Delta_I,\Delta_{II} \geq 0$ such that:
        \begin{enumerate}
            \item Assuming the principal follows threshold policy as in \eqref{eqn:Principal-Threshold} with some fixed threshold $m(Z_0)$ that only depends on $Z_0$, then $h^{\risky}(Z_0)$ can be chosen such that it (weakly) increases in $\Delta_I$ and $h^{\safe}(Z_0)$ can be chosen such that it (weakly) decreases in $\Delta_{II}$.
            \item Furthermore, $h^{\risky}(Z_0), h^{\safe}(Z_0)$ can be chosen so that they both (weakly) decrease in $m(Z_0)$.
        \end{enumerate}
    \end{prop}

    We now turn to changes in the optimal algorithmic recommendation itself.
    A natural starting point for giving recommendations is to have the algorithm recommend the optimal action it would take if it were to make the decision itself.
    However, as the example above shows, this optimal decision would not generally correspond to an optimal recommendation.
    Furthermore, the optimal recommendation itself depends on the degree of recommendation dependence.

    \begin{prop}[Optimal algorithmic decision vs optimal algorithmic recommendation]
    \label{prop:DecisionvsRec}
        The optimal threshold $m^*_{\Delta_I,\Delta_{II}}(Z_0)$ for \eqref{eqn:Principal-Threshold} is not generally the same as $p^* = \frac{c_I}{c_I + c_{II}}$, which is the threshold in \eqref{eqn:Principal-Threshold} that leads to a loss-minimizing decision of the algorithm if the algorithm were to be implemented directly.
        Furthermore, the optimal threshold $m^*_{\Delta_I,\Delta_{II}}(Z_0)$ generally depends on $\Delta_I,\Delta_{II}$.
    \end{prop}

    As the main result of this section, we now consider how the optimal threshold of the algorithm itself depends on the level of recommendation dependence.
    In order to simplify the derivation of some of these comparative statics, we make the additional assumption that human and machine information are continuously distributed with full support, that the function $\phi(\cdot,\cdot;Z_0)$ is continuously differentiable and strictly positively increasing, and that the optimal threshold in the reference-independent case is unique with well-behaved expected loss around the optimum.

    \begin{asm}[Continuously distributed signals and differentiable outcome probabilities]
        \label{asm:smoothness}
        Conditional on the common context $Z_0$, $\phi_h(Z_h;Z_0)$ and $\phi_m(Z_m;Z_0)$ are a.s.\ continuously distributed on $\R$ (that is, their measures are absolutely continuous with respect to Lebesgue measure) with positive density,
        and $\phi(\cdot,\cdot;Z_0)$ is continuously differentiable and strictly monotonically increasing, given $Z_0$.
        Furthermore, almost surely we have that the optimal threshold $m^*(Z_0) = \argmin_m \E[\ell(Y,A)|Z_0]$ for the reference-independent case $\Delta_I = 0 = \Delta_{II}$ is unique with $\left. \frac{\partial^2}{\partial^2 m} \E[\ell(Y,A)|Z_0] \right|_{m = m^*(Z_0)} > 0$.
    \end{asm}

    As suggested by the example, we would generally expect that the optimal threshold $m^*(Z_0)$ decreases in $\Delta_I$ and increases in $\Delta_{II}$, that is, the recommendation to which the agent adheres too much should be given less.
    While there are pathological cases in which the comparative statics can move in the opposite direction, that statement holds under regularity assumptions in a neighborhood around the benchmark $\Delta_I = 0 = \Delta_{II}$ without recommendation dependence.

    \begin{prop}[Threshold monotonicity]
    \label{prop:Localmonotonicity}
        Assume that Assumptions~\ref{asm:independent}--\ref{asm:smoothness} hold.
        Then the optimal threshold $m^*_{\Delta_I,\Delta_{II}}(Z_0)$ is almost surely continuously differentiable for small $\Delta_I, \Delta_{II} \geq 0$, with $\frac{\partial}{\partial \Delta_I} m^*_{\Delta_I,\Delta_{II}}(Z_0) < 0$ and $\frac{\partial}{\partial \Delta_{II}} m^*_{\Delta_I,\Delta_{II}}(Z_0) > 0$.
    \end{prop}

    In particular, increasing the decision loss when the bad outcome materializes leads to a recommendation that is more likely to recommend the risky decision.
    The reason is that increased recommendation dependence in the case of a safe recommendation (higher $\Delta_{II}$) means that the decision-maker does not make the risky decision enough.
    As an optimal response, the algorithm recommends the safe action less, thereby shifting away from the inefficient decision region.

    We finish this discussion by considering the design of recommendations when a third option is available ($\mathcal{R} = \{\risky,\safe,\neutral\}$).
    We again invoke our assumptions from above and consider machine recommendations
    \begin{equation}
        \label{eqn:thirdoption-thresholds}
        R = \rec(X_m)
        =
        \begin{cases}
            \risky, & \P(Y{=}\bad|X_m) \leq m^{\low}(Z_0), \\
            \neutral, & m^{\low}(Z_0) < \P(Y{=}\bad|X_m) \leq m^{\high}(Z_0),
            \\
            \safe, & \P(Y{=}\bad|X_m) > m^{\high}(Z_0)
        \end{cases}
    \end{equation}
    based on simple thresholds on the machine prediction.
    We note that the complementarity result from \autoref{prop:Complementarity} still applies if we restrict recommendations to take this form.
    As in the case of simple binary recommendations, optimal thresholds are still monotonic in the strength of recommendation dependence in a neighborhood around the benchmark case without recommendation dependence, under the same assumptions.
    
    \begin{prop}[Threshold monotonicity with non-recommendation]
    \label{prop:Thirdmonotonicity}
        Assume that Assumptions~\ref{asm:independent}--\ref{asm:smoothness} hold.%
        \footnote{Here, we interpret the assumption on the second derivative of the expected loss function in \autoref{asm:smoothness} to mean that the Hessian matrix at the unique optimal thresholds $m^{\low *}(Z_0), m^{\high *}(Z_0)$ without recommendation dependence is positive definite.}
        Then the optimal thresholds $m^{\low *}_{\Delta_I,\Delta_{II}}(Z_0), m^{\high *}_{\Delta_I,\Delta_{II}}(Z_0)$ are almost surely continuously differentiable for small $\Delta_I, \Delta_{II} \geq 0$, with $\frac{\partial}{\partial \Delta_I} m^{\low *}_{\Delta_I,\Delta_{II}}(Z_0) < 0$ and $\frac{\partial}{\partial \Delta_{II}} m^{\high *}_{\Delta_I,\Delta_{II}}(Z_0) > 0$.
    \end{prop}

        \section{General Minimax Algorithm}
        
        \label{apx:Minimax}

        \begin{prop}[General minimax optimal triage algorithm]
     \label{prop:Triageminimaxgeneral}
        Assume that $\overline{\P}(Y{=}\bad|X_m,A_0) \in (0,1)$ almost surely.
        Write
        \begin{align*}
             \overline{\rec}(x_m)
             &=
             \begin{cases}
                \overline{\rec}_m(x_m), &\overline{\E}[\ell(Y,\overline{\rec}_m(x_m))|X_m=x_m] < \overline{\E}[\ell(Y,A_0)|X_m=x_m], \\
                \neutral, &\text{otherwise}
             \end{cases},
             \\
             \overline{\rec}^\dagger(x_m)
             &=
             \begin{cases}
                \overline{\rec}_m(x_m), &\overline{\E}[\ell(Y,\overline{\rec}_m(x_m))|X_m=x_m] < \overline{\E}[\ell(Y,A^\dagger_0)|X_m=x_m], \\
                \neutral, &\text{otherwise}
             \end{cases}
         \end{align*}
         and let
         \begin{equation}
            \label{eqn:Recstar}
             \overline{\rec}^*
             =
             \begin{cases}
                \overline{\rec},
                &
                \begin{aligned}
                &\overline{\E}[\Ind{\overline{\rec}(X_m) {\neq} \neutral} \ell(Y,\overline{\rec}(X_m)) + \Ind{\overline{\rec}(X_m) {=} \neutral} \ell(Y,A_0)]
                \\
                &\leq
                 \overline{\E}[\Ind{\overline{\rec}^\dagger(X_m) {\neq} \neutral} \ell(Y,\overline{\rec}^\dagger(X_m)) + \Ind{\overline{\rec}^\dagger(X_m) {=} \neutral} \ell(Y,A^\dagger_0)]
                \end{aligned}
                \\
                \overline{\rec}^\dagger, &\text{otherwise}
             \end{cases}.
        \end{equation}
        Then the algorithm $\overline{\rec}^*$
         minimizes the worst-case loss
         \(
             \textstyle\sup_{\P \in \mathcal{P}(\overline{\P}), \Delta_I \geq 0, \Delta_{II} \geq 0}
             \E[\ell(Y,A)]
         \)
         over all recommendation algorithms $\rec: \mathcal{X}_m \rightarrow \{ \risky,\neutral, \safe \}$,
         where $A$ denotes the recommendation-dependent choices of a human decision-maker with $\Delta_I,\Delta_{II}$ who observes recommendations $R = \rec(X_m)$.
     \end{prop}

    This recommendation algorithm is a version of the one in \autoref{prop:Triageminimax}, only that it considers two alternatives for the case of not making a recommendation: either the actual human decision or its inverse. 
    Depending on which yields a better worst-case loss based on information about the (known) distribution $\overline{\P}$, it picks the better of the two. (The assumption in \autoref{prop:Triageminimax} rules out the latter as a contender.)
     The resulting solution is non-trivial, as neither sending the machine decision ($\rec(x_m) \equiv \overline{\rec}_m(x_m)$) nor not making any recommendations ($\rec(x_m) \equiv \neutral$) would generally be minimax optimal. Both algorithms guarantee complementarity in the sense of \autoref{prop:Complementarity}:
 
    \begin{prop}[Feasible human--algorithm complementarity]
        \label{prop:Generalfeasiblecomplementarity}
        The recommendation algorithms from 
        \autoref{prop:Triageminimax} and from \autoref{prop:Triageminimaxgeneral}
        improve over both human and machine decisions,
        \[\max\{\E[\ell(Y,\overline{A})],\E[\ell(Y,\overline{A}^*)]\} \leq \min\{\E[\ell(Y,A_0)], \E[\ell(Y,A_m]\},\]
        where $\overline{A},\overline{A}^*$ denote the recommendation-dependent choices of a human decision-maker with $\Delta_I,\Delta_{II}$ who observes recommendations $R = \overline{\rec}(X_m)$ and $R = \overline{\rec}^*(X_m)$, respectively. 
     \end{prop}

    In setting of \autoref{prop:Onesidedminimax}, there is a whole family of minimax algorithms that include $\overline{\rec}', \overline{\rec}''$:

     \begin{prop}[Minimax recommendation with one-sided recommendation dependence]
     \label{prop:Onesidedminimaxgeneral}
        The recommendation algorithms
         \begin{equation}
        \label{eqn:Onesidedtriagegeneral}
         \overline{\rec}_{p^\low, p^\high}(x_m)
         =
         \begin{cases}
            \safe, & \overline{\P}'(Y'=\bad|X_m=x_m, A_0 = \risky) > p^\high, \\
            \risky, & \overline{\P}'(Y'=\bad|X_m=x_m, A_0 = \risky) \leq p^\low, \\
            \neutral, &\text{otherwise} \\
         \end{cases}
     \end{equation}
     for any $p^\low \leq p^\high$ with $p^\low = p^*$ and/or $p^\high = p^*$
         minimize worst-case loss
         $
             \textstyle\sup_{\P \in \mathcal{P}'(\overline{\P}'), \Delta_I = 0, \Delta_{II} \geq 0} \E[\ell(Y,A)]
         $
         over recommendation algorithms $\rec: \mathcal{X}_m \rightarrow \{ \risky,\neutral, \safe \}$,
         where $A$ denotes the recommendation-dependent choices of a human decision-maker following recommendations $R = \rec(X_m)$.
     \end{prop}

     We obtain $\overline{\rec}'$ from $p^\high = p^*, p^\low < 0$ and  $\overline{\rec}''$ from  $p^\low = p^*, p^\high \geq 1$.

     \section{Example for Strategic Silence}

     \label{apx:Continuous}

     \begin{exm}[name=Independent signal with symmetric losses,label=exm2]
        As in \autoref{exm1}, we consider private signals $X_h$ and $X_m$ that are drawn independently from a uniform distribution on $[0,1]$.
        But unlike in \autoref{exm1}, we now assume $Y$ is stochastic conditional on $X_h$ and $X_m$,
        \[\P(Y {=} \bad|X_m,X_h) = \frac{X_m+X_h}{2}.\]
        The probability of the bad outcome occurring is illustrated in Panel~(a) of \autoref{fig:ex-continuous}.

        Assuming symmetric error costs $c_I = 1 = c_{II}$,
        the optimal decision given both signals $X_m$ and $X_h$ is $A=\risky$ if $X_h + X_m \leq 1$ and $A=\safe$ otherwise.
        This optimal decision is illustrated in Panel~(b) of \autoref{fig:ex-continuous}.
        In this example, the machine prediction of the bad outcomes is $\P(Y {=} \bad|X_m) = \frac{1+2 X_m}{4}$.
        Since the machine signal $X_m$ can be recovered from the machine prediction $\widehat{Y}$, a human decision-maker without recommendation dependence takes the optimal decision.
        With recommendation-dependent preferences as in \eqref{eqn:Implicit},
        the human decision-maker instead chooses
        \[
            A = \begin{cases}
                \risky, & X_h \leq 1 - X_m - \frac{\Delta_{II}}{2+\Delta_{II}} \: \Ind{X_m > \sfrac{1}{2}},
                \\
                \safe, & X_h > 1 - X_m - \frac{\Delta_{II}}{2+\Delta_{II}} \: \Ind{X_m > \sfrac{1}{2}},
            \end{cases},
        \]
        which is illustrated in Panel~(c) of \autoref{fig:ex-continuous}.
        From the perspective of the principal, this choice creates inefficiencies where the risky decision is taken too little, especially for high values of $\Delta_{II}$.

        In the example, now consider the risk score
        \[
            \widehat{Y}
            =
            \begin{cases}
                \P(Y{=}\bad|X_m), & X_m \notin [\sfrac{1}{2}{-}\epsilon,\sfrac{1}{2}{+}\epsilon],\\
                \withheld, & X_m \in [\sfrac{1}{2}{-}\epsilon,\sfrac{1}{2}{+}\epsilon],
            \end{cases}
        \]
        where $\P(Y{=}\bad|X_m) = \frac{1+2X_m}{4}$ and we assume that the agent interprets the withheld risk score as $\widehat{Y} = \P(Y{=}\bad|X_m \in [\sfrac{1}{2}{-}\epsilon,\sfrac{1}{2}{+}\epsilon]) = \sfrac{1}{2}$.
        As a consequence, there is no recommendation dependence when the score is withheld, and the decision-maker takes actions as in Panel~(d) of \autoref{fig:ex-continuous}.
    Withholding information around $X_m=\sfrac{1}{2}$ eliminates recommendation dependence for $X_m \in [\sfrac{1}{2}{-}\epsilon,\sfrac{1}{2})$ (although decisions are still not first best), while also leading to inefficient decisions for $X_m \in (\sfrac{1}{2},\sfrac{1}{2}{+}\epsilon]$.
    For small $\epsilon$, the gain from reducing recommendation dependence outweighs the cost from withholding information.
\end{exm}

    \begin{figure}[p]
        \centering
        \begin{subfigure}[T]{0.48\textwidth}
            \centering

            \begin{tikzpicture}
                \begin{axis}[
                    axis equal image,
                    xlabel={$X_h$}, xlabel style={yshift=15pt},
                    ylabel={$X_m$}, ylabel style={yshift=-15pt},
                    xmin=0, xmax=1,
                    ymin=0, ymax=1,
                    width=.75\textwidth,
                    height=.75\textwidth,
                    xtick={0,1},
                    ytick={0,1},
                    tick align=outside,
                    enlargelimits=false,
                    grid style=dashed,
                    grid=both,
                    axis on top,
                    clip=false
                ]
                \useasboundingbox (0,0) rectangle (1.2,1);

                \colorlet{mypurple}{red!50!blue}
                \addplot[color=mypurple!70, ultra thick] coordinates {(0,1) (1,0)};
                \node[color=mypurple!70,anchor=south west] at (axis cs:0.5,0.5) {$\sfrac{1}{2}$};

                \addplot[color=red!70,ultra thick] coordinates {(1,0.5) (0.5,1)};
                \node[color=red!70,anchor=south west] at (axis cs:0.75,0.75) {$\sfrac{3}{4}$};

                \addplot[color=blue!70,ultra thick] coordinates {(0.5,0) (0,0.5)};
                \node[color=blue!70,anchor=south west] at (axis cs:0.25,0.25) {$\sfrac{1}{4}$};
                \end{axis}
            \end{tikzpicture}

            \caption{The square represents the uniform distribution over signals $X_h$ and $X_m$, with the lines illustrating the probability $\P(Y {=} \bad|X_h,X_m)$ at different levels.}
            \label{fig:sec_mod}
        
        \end{subfigure}
        \hspace{5pt}
        \begin{subfigure}[T]{0.48\textwidth}
            \centering

            \begin{tikzpicture}
                \begin{axis}[
                    axis equal image,
                    xlabel={$X_h$}, xlabel style={yshift=15pt},
                    ylabel={$X_m$}, ylabel style={yshift=-15pt},
                    xmin=0, xmax=1,
                    ymin=0, ymax=1,
                    width=.75\textwidth,
                    height=.75\textwidth,
                    xtick={0,1},
                    ytick={0,1},
                    tick align=outside,
                    enlargelimits=false,
                    grid style=dashed,
                    grid=both,
                    axis on top,
                    clip=false
                ]
                \useasboundingbox (0,0) rectangle (1.2,1);

                \fill [blue!10] (axis cs:0,0) -- (axis cs:0,1) -- (axis cs:1,0) -- cycle;
                \fill [red!10] (axis cs:0,1) -- (axis cs:1,1) -- (axis cs:1,0) -- cycle;  

                \addplot[color=black] coordinates {(0,1) (1,0)};                

                \node at (axis cs:0.25,0.25) {$\risky$};
                \node at (axis cs:0.75,0.75) {$\safe$};
                \end{axis}
            \end{tikzpicture}

            \caption{
            The optimal decision from knowing both signals $X_h$ and $X_m$ (as well as the decision taken by a machine-assisted human decision-maker without recommendation dependence) is to take the $\risky$ action in the lower left quadrant where the outcome is more likely to be $\good$ than $\bad$, and the $\safe$ action otherwise.}
        \end{subfigure}

        \begin{subfigure}[T]{0.48\textwidth}
            \centering

            \begin{tikzpicture}
                \begin{axis}[
                    axis equal image,
                    xlabel={$X_h$}, xlabel style={yshift=15pt},
                    ylabel={$R = X_m$},
                    xmin=0, xmax=1,
                    ymin=0, ymax=1,
                    width=.75\textwidth,
                    height=.75\textwidth,
                    xtick={0,1},
                    ytick={0,1},
                    tick align=outside,
                    enlargelimits=false,
                    grid style=dashed,
                    grid=both,
                    axis on top,
                    clip=false
                ]
                \useasboundingbox (0,0) rectangle (1.2,1);

                \fill [blue!10] (axis cs:0,0) -- (axis cs:0,1) -- (axis cs:1,0) -- cycle;
                \fill [red!10] (axis cs:0,1) -- (axis cs:1,1) -- (axis cs:1,0) -- cycle;  
                
                \fill [blue!70] (axis cs:0.375,0.5) -- (axis cs:0,0.875) -- (axis cs:0,1) -- (axis cs:0.5,0.5) -- cycle;

                \addplot[color=black] coordinates {(0,1) (1,0)};                
                \addplot[color=black, ultra thick] coordinates {(0.375,0.5) (0,0.875)};
                \addplot[color=black, ultra thick] coordinates {(0.5,0.5) (1,0)};
                \addplot[color=black, dashed,thick] coordinates {(0,0.5) (1,0.5)};
                \node at (axis cs:0.25,0.25) {$A=\risky$};
                \node at (axis cs:0.75,0.75) {$A=\safe$};
                \end{axis}
            \end{tikzpicture}

            \caption{Recommendation-dependence with $\Delta_{II} > 0$ leads to excess safe action, which in turn produces excess loss from type-II errors (dark blue region) from the perspective of the principal.
            }
        \end{subfigure}
        \hspace{5pt}
        \begin{subfigure}[T]{0.48\textwidth}
        \centering

        \begin{tikzpicture}
            \begin{axis}[
                axis equal image,
                xlabel={$X_h$}, xlabel style={yshift=15pt},
                ylabel={$X_m$}, ylabel style={yshift=-15pt},
                xmin=0, xmax=1,
                ymin=0, ymax=1,
                width=.75\textwidth,
                height=.75\textwidth,
                xtick={0,1},
                ytick={0,1},
                tick align=outside,
                enlargelimits=false,
                grid style=dashed,
                grid=both,
                axis on top,
                clip=false
            ]
            \useasboundingbox (0,0) rectangle (1.2,1);

            \fill [blue!10] (axis cs:0,0) -- (axis cs:0,1) -- (axis cs:1,0) -- cycle;
            \fill [red!10] (axis cs:0,1) -- (axis cs:1,1) -- (axis cs:1,0) -- cycle;  
            
            \fill [blue!70] (axis cs:0.275,0.6) -- (axis cs:0,0.875) -- (axis cs:0,1) -- (axis cs:0.4,0.6) -- cycle;
            \fill [red!70] (axis cs:0.4,0.6) -- (axis cs:.5,.5) -- (axis cs:0.5,0.6) -- cycle;
            \fill [blue!70] (axis cs:0.5,0.5) -- (axis cs:.5,.4) -- (axis cs:0.6,0.4) -- cycle;

            \addplot[color=black] coordinates {(0,1) (1,0)};                
            \addplot[color=black, ultra thick] coordinates {(0.275,0.6) (0,0.875)};
            \addplot[color=black, ultra thick] coordinates {(0.5,0.6) (0.5,0.4)};
            \addplot[color=black, ultra thick] coordinates {(0.6,0.4) (1,0)};
            \addplot[color=black, dashed,thick] coordinates {(0,0.5) (1,0.5)};
            \addplot[color=black, dashed, ultra thick] coordinates {(0,0.6) (1,0.6)};
            \addplot[color=black, dashed, ultra thick] coordinates {(0,0.4) (1,0.4)};

            \draw[<->, color=black,thick] (axis cs:0.75,0.4) -- (axis cs:.75,0.5);
            \node[anchor=west] at (axis cs:.75,0.45) {$\epsilon$};

            \node at (axis cs:0.25,0.25) {$A=\risky$};
            \node at (axis cs:0.75,0.75) {$A=\safe$};
            \node[anchor=west] at (axis cs:1.1,.8) {$R=X_m$};
            \node[anchor=west] at (axis cs:1.1,0.5) {$R=\withheld$};
            \node[anchor=west] at (axis cs:1.1,0.2) {$R=X_m$};
        \end{axis}
        \end{tikzpicture}

        \caption{
            Strategically withholding predictions around $M = \sfrac{1}{2}$ reduces the region in which recommendation dependence distorts decisions, and improves expected loss for the principal.}
        \label{fig:gap}
        \end{subfigure}

        \caption{Distribution of outcome (top left), optimal decision (top right), recommendation-dependent decision (bottom left), and recommendation-dependent decision with withheld machine prediction (bottom right) in \autoref{exm2}.}

        \label{fig:ex-continuous}
    \end{figure}

    \section{Proofs}
    \label{apx:proofs}

        \begin{proof}[Proof of \autoref{prop:PT}]
    Writing out this reference-dependent loss with loss aversion for the specific loss functions, we find that
    \begin{align*}
        \ell^{\text{PT}}(Y,A,R)
        &= \lambda [\ell(Y,A) - \ell(Y,R)]_+ -  \: [\ell(Y,A) - \ell(Y,R)]_-
        \\
        &= \ell(Y,A) - \ell(Y,R) + (\lambda - 1) [\ell(Y,A) - \ell(Y,R)]_+
        \\
        &=
        \ell(Y,A) - \ell(Y,R)
        +
        \begin{cases}
            (\lambda - 1) c_I, & Y = \good, A=\safe, R = \risky, \\
            (\lambda - 1) c_{II}, & Y = \bad, A=\risky, R = \safe.
        \end{cases}
    \end{align*}
    Since $\ell(Y,R)$ is not affected by the decision-maker's choice, their preferences are as if they are minimizing expected loss with loss function
    \[
        \ell^*(Y,A,R) = \ell(Y,A)
        +
        \begin{cases}
            (\lambda - 1) c_I, & Y = \good, A=\safe, R = \risky, \\
            (\lambda - 1) c_{II}, & Y = \bad, A=\risky, R = \safe,
        \end{cases}
    \]
    as in \eqref{eqn:Recdeppref} with $\Delta_I = (\lambda - 1) c_I, \Delta_{II} = (\lambda - 1) c_{II}$.
    \end{proof}

    \begin{proof}[Proof of \autoref{prop:Costlydefault}]
    \allowdisplaybreaks
        Enumerating all combinations of recommendation, action, outcome, we obtain:
        \begin{align*}
            \ell^\text{default}(y,a,r)
            &=
            \begin{cases}
                \Ind{y=\bad} c
                +
                \begin{cases}
                    c_{II} - c, & y = \bad, a = \risky \\
                    c + c_{I}, & y = \good, a = \safe \\
                    0, & \text{otherwise}
                \end{cases},
                & r = \risky \\
                \Ind{y=\good} c
                +
                \begin{cases}
                    c + c_{II}, & y = \bad, a = \risky \\
                    c_{I} - c, & y = \good, a = \safe \\
                    0, & \text{otherwise}
                \end{cases},
                & r = \safe            
            \end{cases}
            \\
            &=
            \underbrace{
            \begin{cases}
                c, & y=\bad, r = \risky \\
                c, & y=\good, r = \safe \\
                0, &\text{otherwise}
            \end{cases}}_{= \ell_0(y,r)}
            +
            \begin{cases}
            \frac{c_{II} - c}{c_{II}}
            \cdot
                \begin{cases}
                    c_{II}, & y = \bad, a = \risky \\
                    \frac{c_{II} (c + c_{I})}{c_{II} - c}, & y = \good, a = \safe \\
                    0, & \text{otherwise}
                \end{cases},
                & r = \risky \\
                \frac{c_{I} - c}{c_I}
                \cdot
                \begin{cases}
                    \frac{c_I(c + c_{II})}{c_{I} - c}, & y = \bad, a = \risky \\
                    c_{I}, & y = \good, a = \safe \\
                    0, & \text{otherwise}
                \end{cases},
                & r = \safe          
            \end{cases}
            \\
            &=
            \ell_0(y,r)
            +
            \underbrace{\begin{cases}
                c_{II}, & y = \bad, a = \risky = r \\
                c_{I} + c \frac{c_{I} + c_{II}}{c_{II} - c}, & y = \good, a = \safe \neq \risky = r \\
                c_{II} + c \frac{c_{I} + c_{II}}{c_{I} - c}, & y = \bad, a = \risky \neq \safe = r \\
                c_{I}, & y = \good, a = \safe = r \\
                0, & \text{otherwise}
            \end{cases}}_{
            \mathclap{
                =
                \begin{cases}
                    c_{I}, & y = \good, a = \safe  \\
                    c_{II}, & y = \bad, a = \risky  \\
                    0, & \text{otherwise}
                \end{cases}
                +
                \begin{cases}
                    c \frac{c_{I} + c_{II}}{c_{II} - c}, & y = \good, a = \safe \neq \risky = r \\
                    c \frac{c_{I} + c_{II}}{c_{I} - c}, & y = \bad, a = \risky \neq \safe = r \\
                    0, & \text{otherwise}
                \end{cases}
            }
            }
            \cdot
            \underbrace{\begin{cases}
            \frac{c_{II} - c}{c_{II}},
                & r = \risky \\
                \frac{c_{I} - c}{c_I},
                & r = \safe          
            \end{cases}}_{=c_0(r)}
            \\
            &=
            \ell_0(y,r)
            +
            c_0(r)
            \cdot
            \left(
            \ell(y,a)
            +
            \begin{cases}
                \Delta_I, & y = \good, a = \safe \neq \risky = r \\
               \Delta_{II}, & y = \bad, a = \risky \neq \safe = r \\
                0, & \text{otherwise}
            \end{cases}
            \right)
        \end{align*}
        for $\Delta_I = c \frac{c_{I} + c_{II}}{c_{II} - c}, \Delta_{II} =  c \frac{c_{I} + c_{II}}{c_{I} - c}$.
        Since an agent faced with a recommendation $R{=}r$ has no control over $\ell_0(Y,r)$ and $c_0(r) > 0$ is a constant, the agent minimizing expected loss takes decisions \emph{as if} minimizing expected loss with the loss function
        \begin{align*}
            \ell^*(y,a,r)
            &=
           \ell(y,a)
            +
            \begin{cases}
                \Delta_I, & y = \good, a = \safe \neq \risky = r \\
               \Delta_{II}, & y = \bad, a = \risky \neq \safe = r \\
                0, & \text{otherwise}
            \end{cases}.
            \qedhere
        \end{align*}
    \end{proof}

    \begin{proof}[Proof of \autoref{prop:Minimaxaudits}]

        Assume, as in the statement of the proposition, that
        \begin{align*}
            \P(A^*=R|Y{=}\good,R{=}\risky \text{ or } Y{=}\bad,R{=}\safe) &\geq \frac{1 + \eta}{2}, 
            \\
            \P(Y{=}\good,R{=}\risky \text{ or } Y{=}\bad,R{=}\safe) & \geq \frac{1}{2}.
        \end{align*}
        Take a policy $\pi$ as given and write $A_\pi$ for the decision of an attentive type given $X_h,R$ and the audit policy $\pi$.
        Then the expected return to the observer is
        \begin{align*}
        &\sum_{y,a,r}
            \pi(y,a,r)
            \P(Y{=}y,R{=}r)
            \Big(
                \frac{q}{2} \overbrace{(1-\cobs)}^{\mathclap{\text{net benefit of auditing $\theta{=}\text{negligent}$ types}}}
                -
                (1-q) \underbrace{\cobs \P(A_\pi{=}a|Y{=}y,R{=}r)}_{\mathclap{\text{cost from auditing a $\theta{=}\text{attentive}$ type}}}
            \Big)
            \\
            &=
            (1-q) \cobs
            \sum_{\pi(y,a,r) = 1}
            \P(Y{=}y,R{=}r)
            \Big(
                \frac{q (1-\cobs)}{2 (1-q) \cobs}
                -
                \P(A_\pi{=}a|Y{=}y,R{=}r)
            \Big).
        \end{align*}
        Consider now choices for the policy that punishes non-aligned mistakes,
        \[\pi(y,r,a) = \Ind{y {=} \bad, r {=} \safe, a {=} \risky} + \Ind{y {=} \good, r {=} \risky, a {=} \safe}.\]
        For choices $A_\pi$ by the attentive agent following this policy, we have that
        \begin{align*}
            \P(A_\pi=R|R{=}r,Y{=}y)
            &=
            \E[\underbrace{\P(A_\pi=r|R{=}r,X_h)}_{\geq \P(A^*=r|R{=}r,X_h)}|R{=}r,Y{=}y]
            \geq
            \P(A^*=R|R{=}r,Y{=}y)
        \end{align*}
        since punishing mistakes against the recommendation only increases adherence, as in the proof of \autoref{rem:Adherence}.
        Hence, the expected utility of the observer is lower bounded by
        \begin{align}
            &(1-q) \cobs
            \sum_{\mathclap{(y,r,a) \in \{ 
            (\bad, \safe, \risky),
            (\good, \risky, \safe)
             \}}}
            \P(Y{=}y,R{=}r)
            \Big(
                \frac{q (1-\cobs)}{2 (1-q) \cobs}
                -
                \P(A^*{=}a|Y{=}y,R{=}r)
            \Big)
                        \notag
            \\
            &=
            (1-q) \cobs
            \overbrace{\P((Y,R) \in \{ (\good,\risky),(\bad,\safe) \})}^{\geq \frac{1}{2}}
            \\
                                    \notag
            &\phantom{=}
            \Big(
                \frac{q (1-\cobs)}{2 (1-q) \cobs}
                -
                \underbrace{\P(A^*{\neq}R|(Y,R) \in \{ (\good,\risky),(\bad,\safe) \})}_{\leq \frac{1 - \eta}{2}}
            \Big)
                        \notag
            \\
            &\geq
            (1-q) \cobs
            \left(\frac{q (1-\cobs)}{(1-q) \cobs} - \frac{1 + \eta}{2}\right)
            / 4
            =
            (1-q) \cobs
            \Big(\eta - \underbrace{\frac{\cobs - q}{(1-q) \cobs}}_{> 0} \Big)
            / 4.
            \label{eqn:lowerlower}
        \end{align}
        
        To establish that this and only this policy is maximin optimal,
        we construct a distribution for which this policy is optimal,
        while any other policy achieves utility strictly below the bound.
        To this end, consider the case of $c_I = c_{II}$, and fix $\delta > 0$ such that $\frac{2 \delta}{1 + \delta^2} = \eta$.
        Let $R \sim \text{Uniform}(\{\risky,\safe\})$.
        The agent's private information is $X_h \sim \text{Bernoulli}(\frac{1 + \delta}{2})$.
        We assume that
        \begin{align*}
            \P(Y = \bad | R{=}r, X_h{=}x_h)
            &=
            \begin{cases}
                \frac{1 + \delta}{2}, &
                r = \safe, x_h = 1, \\
                \frac{1 - \delta}{2}, &
                r = \safe, x_h = 0, \\
                \frac{1 - \delta}{2}, &
                r = \risky, x_h = 1, \\
                \frac{1 + \delta}{2}, &
                r = \risky, x_h = 0. \\
            \end{cases}
        \end{align*}
        By \autoref{rem:Thresholds}, the optimal, expected-loss-minimizing decision by the agent is then
        \begin{align*}
            A^* = \begin{cases}
                R, & X_h = 1, \\
                R^\dagger, & X_h = 0,
            \end{cases}
        \end{align*}
        where $R^\dagger$ is the opposite action of the recommendation $R$ (that is, $\{R, R^\dagger\} = \{\risky,\safe\}$).
        We furthermore assume that $\chum$ is small enough to ensure that $A_\pi = A^*$ for all policies $\pi$ (that is, $\frac{\chum}{2+\chum} < \delta$), and we write $A = A_\pi = A^*$ for simplicity.
        Then we have that
        \begin{align*}
            \P(A{=}R|R) &= \frac{1 + \delta}{2},
            &
            \P(Y{=}\bad|R{=}\safe) &= \left(\frac{1 + \delta}{2}\right)^2 + \left(\frac{1 - \delta}{2}\right)^2
            = \frac{1 + \delta^2}{2}
            = \P(Y{=}\good|R{=}\risky).
        \end{align*}
        As a consequence, since
        \begin{align*}
            &\P(A{=}a|Y{=}y,R{=}r)
            \\
            &= \frac{\P(Y{=}y|A{=}a,R{=}r) \: \P(A{=}a|R{=}r)}{\P(Y{=}y|A{=}\risky,R{=}r) \: \P(A{=}\risky|R{=}r) + \P(Y{=}y|A{=}\safe,R{=}r) \: \P(A{=}\safe|R{=}r)}
        \end{align*}
        we find that
        \begin{align*}
            \P(A{=}\risky|Y{=}\bad,R{=}\safe)
            = \P(A{=}\safe|Y{=}\good,R{=}\risky)
            &=
            \frac{(1-\delta)^2}{(1-\delta)^2 + (1+\delta)^2}
            = \frac{1 - \eta}{2}
            \\
            \P(A{=}\risky|Y{=}\bad,R{=}\risky)
            = \P(A{=}\safe|Y{=}\good,R{=}\safe)
            &=
            \frac{(1-\delta) (1+\delta)}{(1-\delta) (1+\delta) + (1+\delta) (1-\delta)}
            = \frac{1}{2}
            \\
            \P(A{=}\risky|Y{=}\good,R{=}\safe)
            = \P(A{=}\safe|Y{=}\bad,R{=}\risky)
            &=
            \frac{(1+\delta) (1-\delta)}{(1+\delta) (1-\delta) + (1-\delta) (1+\delta)}
            = \frac{1}{2}
            \\
            \P(A{=}\risky|Y{=}\good,R{=}\risky)
            = \P(A{=}\safe|Y{=}\bad,R{=}\safe)
            &=
            \frac{(1+\delta)^2}{(1+\delta)^2 + (1-\delta)^2}
            = \frac{1 + \eta}{2},
        \end{align*}
        where we have used that
        \begin{align*}
            \frac{(1-\delta)^2}{(1-\delta)^2 + (1+\delta)^2}
            - \frac{1}{2}
            =
            \frac{1 + \delta^2 - 2 \delta}{2 + 2 \delta^2}
            - \frac{1}{2}
            &= - \frac{2 \delta}{2 + 2 \delta^2} = - \frac{\eta}{2},
            \\
            \frac{(1+\delta)^2}{(1-\delta)^2 + (1+\delta)^2}
            - \frac{1}{2}
            =
            \frac{1 + \delta^2 + 2 \delta}{2 + 2 \delta^2}
            - \frac{1}{2}
            &= \frac{2 \delta}{2 + 2 \delta^2} = \frac{\eta}{2}.
        \end{align*}
        In particular, it follows that
        \begin{align*}
            \P(A=R|Y{=}\bad,R{=}\safe \text{ or } Y{=}\good,R{=}\risky) &= \frac{1 +\eta}{2},
            \\
            \P(Y{=}\bad,R{=}\safe \text{ or } Y{=}\good,R{=}\risky)
            &=
            \frac{1+\delta^2}{2}
            \geq \frac{1}{2}
        \end{align*}
        and thus the observer's assumptions are fulfilled for this distribution.
        Now, the expected utility from a policy $\pi$ is
        \begin{align*}
            &(1-q) \cobs
            \sum_{y,a,r}
            \pi(y,a,r)
            \P(Y{=}y,R{=}r)
            \Big(
                \frac{q (1-\cobs)}{2 (1-q) \cobs}
                -
                (\P(A{=}a|Y{=}y,R{=}r))
            \Big)
            \\
            &=
            \frac{(1-q) \cobs}{2}
            \sum_{y,a,r}
            \pi(y,a,r)
            \P(Y{=}y,R{=}r)
            \Big(
                1 - 2 \P(A{=}a|Y{=}y,R{=}r) - 
                \underbrace{\frac{\cobs - q}{(1-q) \cobs}}_{= k > 0}
            \Big)
            \\
            &=
            \frac{(1-q) \cobs}{2}
            \Bigg(
            (\pi(\bad,\risky,\safe) + \pi(\good,\safe,\risky))
            \frac{1+\delta^2}{4} (\eta - k)
            \\
            &\phantom{=\frac{(1-q) \cobs}{2}
            \Bigg(}+
            (\pi(\bad,\risky,\risky) + \pi(\good,\safe,\safe) + \pi(\good,\risky,\safe) + \pi(\bad,\safe,\risky)) \frac{1}{4} (-k)
            \\
            &\phantom{=\frac{(1-q) \cobs}{2}
            \Bigg(}+
            (\pi(\good,\risky,\risky) + \pi(\bad,\safe,\safe)) \frac{1-\delta^2}{4} (-\eta-k)
            \Bigg).
        \end{align*}
        Since $\frac{1+\delta^2}{4} \in (\sfrac{1}{4}, \sfrac{1}{2})$,
        the \emph{only} choice of $\pi$ for which the expected utility is at least that of the lower bound from \autoref{eqn:lowerlower}
        is to set $\pi = \pi^*$ with $\pi^*(y,r,a) = \Ind{y {=} \bad, a {=} \risky, r {=} \safe} + \Ind{y {=} \good, a {=} \safe, r {=} \risky}$.
        Hence, the worst-case expected utility from any policy $\pi \neq \pi^*$ is below the worst-case expected utility for $\pi^*$.
        In other words, the policy $\pi^*$ is maximin optimal.
    \end{proof}
    
    \begin{proof}[Proof of \autoref{rem:Thresholds}]
        Given $R=r$ and $X_h=x_h$, the agent minimizes
        \begin{align*}
            \E[\ell(Y,a)|X_h=x_h,R=r]
            =
            &\Ind{a = \risky}
            \P(Y{=}\bad|X_h=x_h,R=r) c_{II}(r) 
            \\
            &+ (1-\Ind{a = \risky}) (1-\P(Y{=}\bad|X_h=x_h,R=r)) c_{I}(r),
        \end{align*}
        which is achieved by choosing $a=\risky$ if and only $\P(Y{=}\bad|X_h=x_h,R=r) \leq \frac{c_{I}(r)}{c_{I}(r) + c_{II}(r)}$.
    \end{proof}
        
    \begin{proof}[Proof of \autoref{rem:Adherence}]
        We have that
        \begin{align*}
            \P(A=R|R=\risky)
            &=
            \P(A=\risky|R=\risky)
            \\
            &=
            \P\left(
                \P(Y{=}\bad|X_h,R{=} \risky)
                \leq p^\risky
            \middle| R=\risky\right)
            \\
            &=
            \P\left(
                \P(Y{=}\bad|X_h,R{=} \risky)
                \leq \frac{c_I + \Delta_I}{c_I + c_{II} + \Delta_I}
            \middle| R=\risky\right)
        \end{align*}
        where $\P(Y{=}\bad|X_h,R{=} \risky)$ is unaffected by $\Delta_I$ and $\frac{c_I + \Delta_I}{c_I + c_{II} + \Delta_I}$ is monotonically increasing in $\Delta_I$, which means that $\P(A=R|R=\risky)$ can not decrease as $\Delta_I$ increases.
        The result for $\P(A=R|R=\safe)$ follows similarly.

        Relative to the optimal agent decision
        \[
            A^* =
            \begin{cases}
                \risky, & \P(Y{=}\bad|X_h,R) \leq p^*, \\
                \safe, & \P(Y{=}\bad|X_h,R) > p^*,
            \end{cases}
        \]
        that minimizes expected loss for the principal (and is not affected by $\Delta_I,\Delta_{II}$),
        the principal experiences additional expected loss
        \begin{equation}
            \label{eqn:extraloss}
            \begin{aligned}
            &\E[\ell(Y,A)] - \E[\ell(Y,A^*)]
            \\
            &=
            \E\Bigg[
                \mathbbm{1}\left(
                \frac{c_I}{c_I {+} c_{II} {+} \Delta_{II}} <
                \P(Y{=}\bad|X_h,R{=}\safe) \leq \frac{c_I}{c_I {+} c_{II}}\right)
                \\
                &\mathrel{{}\phantom{=\E\Bigg[}{}}
                \underbrace{(c_{I} \P(Y{=}\good|X_h,R{=}\safe)
                -
                c_{II} \P(Y{=}\bad|X_h,R{=}\safe))}_{\geq 0}
                \Bigg|
                R {=} \safe\Bigg]
            \: \P(R {=} \safe)
            \\
            &\phantom{{}={}}
            +
            \E\Bigg[
                \mathbbm{1}\left(
                \frac{c_{I}}{c_I {+} c_{II}} <
                \P(Y{=}\bad|X_h,R{=}\risky) \leq \frac{c_{I}+\Delta_I}{c_I {+} c_{II} {+} \Delta_I}\right)
                \\
                &\mathrel{{}\phantom{=\E\Bigg[}{}}
                \underbrace{(c_{II} \P(Y{=}\bad|X_h,R{=}\risky)
                -
                c_{I} \P(Y{=}\good|X_h,R{=}\risky))}_{\geq 0}
                \Bigg|
                R {=} \risky\Bigg]
            \: \P(R {=} \risky)
            \end{aligned}
        \end{equation}
        where the indicator functions are picking up more cases as $\Delta_I,\Delta_{II}$ increase, thus increasing the additional expected loss.

        For the result on large $\Delta_I,\Delta_{II}$, assuming that $c_I,c_{II} > 0$, we have that
        \begin{align*}
            &\P(A \neq R, \ell(Y,A) > 0)
            =
            \P(A = \risky, R = \safe, Y = \bad)
            +
            \P(A = \safe, R = \risky, Y = \good)
            \\
            &=
            \P\left(
                \P(Y{=}\bad|X_h,R{=}\safe) \leq \frac{c_I}{c_I + c_{II} + \Delta_{II}},
                R = \safe,
                Y = \bad\right)
            \\
            &\phantom{{}={}}
            +
            \P\left(
                \P(Y{=}\bad|X_h,R{=}\risky) > \frac{c_I + \Delta_I}{c_I + c_{II} + \Delta_I},
                R = \risky,
                Y = \good\right)
            \\
            &=
            \P\left(
                \P(Y{=}\bad|X_h,R{=}\safe) \leq \frac{c_I}{c_I + c_{II} + \Delta_{II}},
                Y = \bad
                \middle|
                R {=} \safe\right)
            \: \P(R {=} \safe)
            \\
            &\phantom{{}={}}
            +
            \P\left(
                \P(Y{=}\good|X_h,R{=}\risky) \leq \frac{c_{II}}{c_I + c_{II} + \Delta_I},
                Y = \good
                \middle|
                R {=} \risky\right)
            \: \P(R {=} \risky)
            \\
            &=
            \E\left[
                \mathbbm{1}\left(\P(Y{=}\bad|X_h,R{=}\safe) \leq \frac{c_I}{c_I {+} c_{II} {+} \Delta_{II}}\right)
                \P(Y{=}\bad|X_h,R{=}\safe)
                \middle|
                R {=} \safe\right]
            \: \P(R {=} \safe)
            \\
            &\phantom{{}={}}
            +
            \E\left[
                \mathbbm{1}\left(\P(Y{=}\good|X_h,R{=}\risky) \leq \frac{c_{II}}{c_I {+} c_{II} {+} \Delta_I}\right)
                \P(Y{=}\good|X_h,R{=}\risky)
                \middle|
                R {=} \risky\right]
            \: \P(R {=} \risky)
            \\
            &\leq
            \frac{c_I}{c_I {+} c_{II} {+} \Delta_{II}} \P(R {=} \safe)
            +
            \frac{c_{II}}{c_I {+} c_{II} {+} \Delta_I} \P(R {=} \risky)
            \rightarrow 0
        \end{align*}
        as $\Delta_I,\Delta_{II} \rightarrow \infty$.
    \end{proof}

        \begin{proof}[Proof of \autoref{prop:Badinfo}]
        We construct $X_m,X_h^{(1)},X_h^{(2)}$ that take two values each, $0$ and $1$.
        For some sufficiently small $\varepsilon > 0$, consider the following independent distributions:
        \begin{align*}
            X_m, X_h^{(1)} &\sim \text{Bernoulli}(\sfrac{1}{2})
            &
            X_h^{(2)} &\sim \text{Bernoulli}(\varepsilon)
        \end{align*}
        Furthermore, let
        \begin{equation}
        \label{eqn:Informed}
            \P(Y = \bad | X_m,X_h^{(1)},X_h^{(2)})
            =
            \begin{cases}
                \frac{c_{I}}{c_{I} + c_{II} + 2 \Delta_{II}} = p^{--}, 
                &X_m = X_h^{(1)}, X_h^{(2)} = 0, \\
                \frac{c_{I}}{c_{I} + c_{II} + \Delta_{II} / 2} = p^{-},
                &X_m = X_h^{(1)}, X_h^{(2)} = 1, \\
                \frac{c_{I} + \Delta_I / 2}{c_{I} + c_{II} + \Delta_{I} / 2} = p^{+},
                &X_m \neq X_h^{(1)}, X_h^{(2)} = 1, \\
                \frac{c_{I} + 2 \Delta_I}{c_{I} + c_{II} + 2 \Delta_{I}} = p^{++},
                &X_m \neq X_h^{(1)}, X_h^{(2)} = 0.
            \end{cases}
        \end{equation}
        Writing $\bar{p}^- = (1-\varepsilon) p^{--} + \varepsilon p^-, \bar{p}^+ = (1-\varepsilon) p^{++} + \varepsilon p^+$, we have
        \begin{align*}
            \P(Y = \bad | X_m,X_h^{(1)})
            =
            \begin{cases}
            \bar{p}^-, & X_m = X_h^{(1)}, \\
            \bar{p}^+, & X_m \neq X_h^{(1)}.
            \end{cases}
        \end{align*}
        For $\varepsilon >0$ sufficiently small, we find that
        \begin{equation}
        \label{eqn:Uninformed}
            \lefteqn{\underbrace{\phantom{p^{--} \leq \bar{p}^- \leq p^{\safe} \leq p^{-} \leq p^*}}_{\text{strict inequalities for $\Delta_{II} > 0$}}}
            p^{--} \leq \bar{p}^- \leq p^{\safe} \leq p^{-} \leq
            \overbrace{p^* \leq p^{+} \leq p^{\risky} \leq \bar{p}^+ \leq p^{++}}^{\text{strict inequalities for $\Delta_{I} > 0$}}.
        \end{equation}
        For any recommendation policy that varies over $X_m \in \{0,1\}$, the more informed agent takes decisions based on $\P(Y = \bad | X_m,X_h^{(1)},X_h^{(2)})$ in \eqref{eqn:Informed}, while the less informed decision-maker takes decisions based on $\P(Y = \bad | X_m,X_h^{(1)},X_h^{(2)})$ in \eqref{eqn:Uninformed}.

        Assume without loss of generality that $\Delta_{II} > 0$ and that $\rec(1) = \safe$. (The same argument applies for $\Delta_{I} > 0$ or $\rec(0) = \safe$ or both.)
        A first-best decision (that minimizes expected loss) is to choose the risky action if and only if $X_m = X_h^{(1)}$.
        This decision is also optimal for the less informed decision-maker, since $X_m$ is revealed by the recommendation and
        \begin{equation*}
            \P(Y = \bad | X_m, X_h^{(1)})=
            \begin{cases}
            \bar{p}^- \geq \min\{p^\risky,p^\safe\}, & X_m = X_h^{(1)}, \\
            \bar{p}^+ \leq \max\{p^\risky,p^\safe\}, & X_m \neq X_h^{(1)}.
            \end{cases}
        \end{equation*}
        However, the more informed decision-maker takes the inefficient action $A = \safe$ for $X_m = X_h^{(1)} = X_h^{(2)} = 1$ (which happens with positive probability $\varepsilon / 4$) since
        \begin{align*}
            \P(Y = \bad | X_m=1,X_h^{(1)}=1,X_h^{(2)}{=}1)
            =
            p^{-} \in (p^{\safe}, p^*).
        \end{align*}
        As a consequence, the choices by the informed decision-maker achieves strictly higher expected loss for the principal.
    \end{proof}

    \begin{proof}[Proof of \autoref{rem:Simpleimprovement}]
        Writing $p(x_h,r) = \P(Y{=}\bad|X_h{=}x_h,R{=}r)$ as in the main text,
        we have that
        \begin{align*}
            & E[\ell(Y,R)]
            -
            \E[\ell(Y,A)]
            \\
            &=
            \P(R{=}\risky) \:
            \E[
                \Ind{A{=}\safe}
                (\ell(Y,\risky) - \ell(Y,\safe))
            |R{=}\risky]
            \\
            &\phantom{=}+
            \P(R{=}\safe) \:
            \E[
                \Ind{A{=}\risky}
                (\ell(Y,\safe) - \ell(Y,\risky))
            |R{=}\safe]
            \\
            &=
            \P(R{=}\risky) \:
            \E[
                \Ind{p(X_h,\risky) > p^\risky}
                \underbrace{(p(X_h,\risky) c_{II} - (1-p(X_h,\risky)) c_I )}_{=(c_I + c_II) p(X_h,\risky) - c_I}
            |R{=}\risky]
            \\
            &\phantom{=}+
            \P(R{=}\safe) \:
            \E[
                \Ind{p(X_h,\safe) \leq p^\safe}
                ((1-p(X_h,\safe)) c_I - p(X_h,\safe) c_{II})
            |R{=}\safe]
            \\
            &=
            (c_I + c_{II}) \P(R{=}\risky) \:
            \E[
                \Ind{p(X_h,\risky) > p^\risky}
                \underbrace{(p(X_h,\risky) - p^*)}_{\geq p^\risky - p^* \geq 0}
            |R{=}\risky]
            \\
            &\phantom{=}+
            \P(R{=}\safe) \:
            \E[
                \Ind{p(X_h,\safe) \leq p^\safe}
                \underbrace{(p^* - p(X_h,\risky))}_{\geq p^* - p^\safe \geq 0}
            |R{=}\safe]
            \geq 0.
            \qedhere
        \end{align*}
    \end{proof}

    \begin{proof}[Proof of \autoref{prop:Inefficient}]
        Consider an arbitrary distribution over $X_m$ as well as $X_h \sim \text{Uniform}(\{-1,+1\})$ independently of $X_m$ with
        \begin{align*}
            \P(Y=\bad|X_m,X_h)
            =
            \begin{cases}
                \frac{c_I + \Delta_I / 2}{c_I + c_{II}}, & X_h = +1, \\
                \frac{c_I}{c_I + c_{II} + \Delta_{II} / 2}, & X_h = -1.
            \end{cases}
        \end{align*}
        Following \autoref{rem:Thresholds}, the first-best optimal decision is
        \begin{align*}
            A^*
            =
            \begin{cases}
                \safe, & X_h = +1, \\
                \risky, & X_h = -1
            \end{cases}
        \end{align*}
        since $\frac{c_I + \Delta_I / 2}{c_I + c_{II}} > p^* = \frac{c_I}{c_I + c_{II}} > \frac{c_I}{c_I + c_{II} + \Delta_{II} / 2}$.
        Without access to a machine recommendation, the agent takes this optimal decision, $A^* = A_0$.
        For any machine recommendation, however, decisions are inefficient since they distort a positive fraction of decisions.
    For example, assume that $\P(R = \safe) > 0$.
    Then, for $R = \safe, X_h = -1$ (which happens with positive probability $\P(R = \safe)/2$), we have that
    $A = \safe$ by \autoref{rem:Thresholds}, since now $\frac{c_I + \Delta_I / 2}{c_I + c_{II}} < \frac{c_I + \Delta_I}{c_I + c_{II}}$.
    From the perspective of the principal, this leads to an efficiency loss of at least $\frac{\Delta_I / 2}{c_I + c_{II}} c_{II} \P(R = \safe) / 2$ from excess type-II errors.
    \end{proof}

    \begin{proof}[Proof of \autoref{prop:Complementarity}]
        For the comparison to machine decisions, 
        consider recommending the optimal machine decision,
        \[
            R =
            \begin{cases}
                \risky, &\P(Y{=}\bad|X_m) \leq p^* = \frac{c_I}{c_I + c_{II}}, \\
                \safe, &\P(Y{=}\bad|X_m) > p^*. \\
            \end{cases}
        \]
        For the action $A$ chosen by the agent to be different from the recommendation, we must have that
        \begin{align*}
            \P(Y{=}\bad|X_h,R{=}\risky) &\geq \frac{c_I + \Delta_I}{c_I + c_{II} + \Delta_I} \geq p^*
            &
            &(\safe  = A \neq R =\risky),
            \\
            \P(Y{=}\bad|X_h,R{=}\safe) &\leq \frac{c_I}{c_I + c_{II} + \Delta_{II}} \leq p^*
             &
            &(\risky  = A \neq R =\safe),
        \end{align*}
        and both cases can only improve over implementing $R$ directly.
        Specifically,
        it follows that
        \begin{align*}
            \P(Y{=}\bad|\safe  {=} A {\neq} R {=}\risky)
            &=
            \E[\P(Y{=}\bad|X_h,R{=}\risky)|\safe  {=} A {\neq} R {=}\risky] \geq p^*,
            \\
            \P(Y{=}\bad|\risky  {=} A {\neq} R {=}\safe)
            &=
            \E[\P(Y{=}\bad|X_h,R{=}\safe) |\risky  {=} A {\neq} R {=}\safe]
            \leq p^*
        \end{align*}
        and thus
        \begin{align*}
            \E[\ell(Y,A)]
            &=
            \E[\ell(Y,A) \Ind{A{=}R}] + \E[\ell(Y,A) \Ind{\safe  {=} A {\neq} R {=}\risky}] + \E[\ell(Y,A) \Ind{\risky  {=} A {\neq} R {=}\safe}]
            \\
            &\leq
            \E[\ell(Y,A) \Ind{A{=}R}]
            +  c_I (1-p^*) \E[\Ind{\safe  {=} A {\neq} R {=}\risky}] + c_{II} p^*
            \E[\Ind{\risky  {=} A {\neq} R {=}\safe}]
            \\
            &=
            \E[\ell(Y,A) \Ind{A{=}R}]
            +  c_{II} p^* \E[\Ind{\safe  {=} A {\neq} R {=}\risky}] + c_{I} (1-p^*)
            \E[\Ind{\risky  {=} A {\neq} R {=}\safe}]
            \\
            &\leq
            \E[\ell(Y,R) \Ind{A{=}R}] + \E[\ell(Y,R) \Ind{\safe  {=} A {\neq} R {=}\risky}] + \E[\ell(Y,R) \Ind{\risky  {=} A {\neq} R {=}\safe}]
            \\
            &= \E[\ell(Y,R)] = \E[\min_a \E[\ell(Y,a)|X_m]].
        \end{align*}
        
        For the comparison to human decisions, 
        consider the recommendation $R \equiv \neutral$, which will lead to the same decision as if the human is acting by themselves.
        Hence, $\E[\ell(Y,A)] \leq \E[\min_a \E[\ell(Y,a)|X_h]]$ for this recommendation.

        For the comparison to optimal two-level recommendations decisions, 
        consider the two-level recommendation $R_-$, which will lead to decisions $A = A_-$.
        Hence, $\E[\ell(Y,A)] \leq \E[\ell(Y,A_-)]$ for this choice of recommendation.        

        Putting all three parts together,
        each part shows that there is a recommendation policy for which the inequality holds.
        Consider now a choice of recommendation policy $\rec$ that minimizes $\E[\ell(Y,A)]$.
        Then, for this choice,
        actions given this recommendation policy do (weakly) better than each of these three policies, and thus must fulfill the inequality.
    \end{proof}

    \begin{proof}[Proof of \autoref{prop:Triageminimax}]
        The result follows from \autoref{prop:Triageminimaxgeneral} by noting that $\overline{\rec}^* = \overline{\rec}$ in \eqref{eqn:Recstar}.
        Indeed, consider \textit{any} fixed recommendation algorithm $\rec$.
        Then 
        \begin{align*}
            &\overline{\E}[\Ind{\rec(X_m) {\neq} \neutral} \ell(Y,\rec(X_m)) + \Ind{\rec(X_m) {=} \neutral} \ell(Y,A_0)]
            \\
            &=
            \overline{\E}[\Ind{\rec(X_m) {\neq} \neutral} \ell(Y,\rec(X_m)) + \Ind{\rec(X_m) {=} \neutral} \E[\ell(Y,A_0)|X_m]]
            \\
            &\leq
            \overline{\E}[\Ind{\rec(X_m) {\neq} \neutral} \ell(Y,\rec(X_m)) + \Ind{\rec(X_m) {=} \neutral} \E[\ell(Y,A^\dagger_0)|X_m]]
            \\
            &=
            \overline{\E}[\Ind{\rec(X_m) {\neq} \neutral} \ell(Y,\rec(X_m)) + \Ind{\rec(X_m) {=} \neutral} \ell(Y,A^\dagger_0)].
        \end{align*}
        In particular, since the first expression is minimized over $\rec$ by $\overline{\rec}$ (and the last by $\overline{\rec}^\dagger$) per the proof of \autoref{prop:Triageminimaxgeneral},
        \begin{align*}
            &\overline{\E}[\Ind{\overline{\rec}(X_m) {\neq} \neutral} \ell(Y,\overline{\rec}(X_m)) + \Ind{\overline{\rec}(X_m) {=} \neutral} \ell(Y,A_0)]
            \\
            &\leq
            \overline{\E}[\Ind{\overline{\rec}^\dagger(X_m) {\neq} \neutral} \ell(Y,\overline{\rec}^\dagger(X_m)) + \Ind{\overline{\rec}^\dagger(X_m) {=} \neutral} \E[\ell(Y,A_0)|X_m]]
            \\
            &\leq
            \overline{\E}[\Ind{\overline{\rec}^\dagger(X_m) {\neq} \neutral} \ell(Y,\overline{\rec}^\dagger(X_m)) + \Ind{\overline{\rec}^\dagger(X_m) {=} \neutral} \E[\ell(Y,A^\dagger_0)|X_m]].
        \end{align*}
        It follows that $\overline{\rec}^* = \overline{\rec}$.
    \end{proof}

    \begin{proof}[Proof of \autoref{prop:Feasiblecomplementarity}]
        The proof follows from the more general result in \autoref{prop:Generalfeasiblecomplementarity}.
    \end{proof}

    \begin{proof}[Proof of \autoref{prop:Unobservedminimax}]
        We follow the structure of the proof of \autoref{prop:Triageminimaxgeneral}.
        
        First, for any recommendation policy $\rec: \mathcal{X}_m \rightarrow \mathcal{R} = \{\risky,\neutral,\safe\}$ consider the decision $\overline{A}'$ given by $\overline{A}' = R$ for $R = \rec(X_m) \neq \neutral$ and $\overline{A}' = A_0$ otherwise, which is a feasible choice for the agent.
        We therefore have that
        \begin{align*}
            \E[\ell(Y,A)] \leq \E[\ell^*(Y,A,R)] \leq \E[\ell^*(Y,\overline{A}',R)] = \E[\ell(Y,\overline{A}')]
        \end{align*}
        since $\ell(Y,A) \leq \ell^*(Y,A,R)$ and $\ell^*(Y,\overline{A}',R) = \ell(Y,\overline{A}')$, where the latter holds because $\overline{A}'$ never represents an action that goes against a recommendation. 
        Furthermore,
        \begin{align*}
            \E[\ell(Y,\overline{A}')]
            &=
            \E[\ell(Y,\risky) \Ind{A_0{=}\risky, R \neq \safe}]
            +
            \E[\ell(Y,\safe) \Ind{A_0{=}\risky, R = \safe}]
            \\
            &\phantom{=}+
            \E[\ell(Y,\risky) \Ind{A_0{=}\safe, R = \risky}]
            +
            \E[\ell(Y,\safe) \Ind{A_0{=}\safe, R \neq \risky}].
        \end{align*}
        Assume now that $R \neq \risky$, as in the case of the proposed recommendation algorithm.
        Then
        \begin{align*}
            \E[\ell(Y,\overline{A}')]
            &=
            \overbrace{\E[\ell(Y,\risky) \Ind{A_0{=}\risky, R {=} \neutral}]
            +
            \E[\ell(Y,\safe) \Ind{A_0{=}\risky, R {=} \safe}]}^{\text{observed for $\P \in \mathcal{P}'(\overline{\P}')$}}
            \\
            &\phantom{=}+
            \underbrace{\E[\ell(Y,\safe) \Ind{A_0{=}\safe}]}_{\text{unidentified for $\P \in \mathcal{P}'(\overline{\P}')$}}.
        \end{align*}
        Since the agent chose $A_0 {=} \safe$ for the latter instances when the risky action was also available,
        we know that $\P(Y{=}\bad|A_0{=}\safe) \geq p^* = \frac{c_I}{c_I + c_{II}}$
        by revealed preference. (We discuss below the case where the agent may make arbitrary mistakes because they never observed $Y$ in this case.)
        As a consequence,
        \begin{align*}
            \E[\ell(Y,\safe) \Ind{A_0{=}\safe}]
            =
            c_{I} (1-\P(Y{=}\bad|A_0{=}\safe)) \P(A_0{=}\safe)
            \leq \frac{c_I c_{II}}{c_I + c_{II}} \underbrace{\P(A_0{=}\safe)}_{\text{observed}}.
        \end{align*}
        As a consequence, plugging in the recommendation $R = \overline{\rec}'(X_m)$ from the proposition, we obtain the upper bound
        \begin{align*}
            \E[\ell(Y,A)]
            &\leq
            \overline{\E}'[\ell(Y',\risky) \Ind{\overline{\rec}'(X_m) {=} \neutral} + \ell(Y',\safe) \Ind{\overline{\rec}'(X_m) {=} \risky}|A_0{=}\risky]
            \:
            \overline{\P}'(A_0{=}\risky)
            \\
            &\phantom{=}+
            \frac{c_I c_{II}}{c_I + c_{II}}
            \overline{\P}'(A_0{=}\safe) = \overline{L}'(\overline{\P}').
        \end{align*}
        
        Second, to establish a lower bound, consider the distribution $\P^* \in \mathcal{P}'(\overline{\P}')$ that extends $\overline{\P}'$ by setting $X_h = A_0$, $\P^*(Y{=}\bad|X_m,X_h{=}\safe) \equiv p^*$.
        On that distribution, the first-best decision $A^*$, which selects $A^* = \risky$ if and only if $\P^*(Y{=}\bad|X_m,X_h) \leq p^*$,
        achieves loss
        \begin{align*}
            \E^*[\ell(Y,A^*)]
            &=
            \E^*[\ell(Y,A^*) | X_h{=}\risky] \P^*(X_h{=}\risky)
            +
            \E^*[\ell(Y,A^*) | X_h{=}\safe] \P^*(X_h{=}\safe)
            \\
            &=
            \E^*[\ell(Y,\risky) \Ind{\P^*(Y{=}\bad|X_m,X_h{=}\risky) \leq p^*}
            \\
            &\phantom{=\E^*[}
            +
            \ell(Y,\safe) \Ind{\P^*(Y{=}\bad|X_m,X_h{=}\risky) > p^*} | X_h{=}\risky] \P^*(X_h{=}\risky)
            \\
            &\phantom{=}
            +
            c_{II} \underbrace{\P^*(Y{=}\bad|X_h{=}\safe)}_{=p^* =\frac{c_I}{c_I + c_{II}}} \P^*(X_h{=}\safe)
            \\
            &=
            \overline{\E}'[\ell(Y',\risky) \Ind{\overline{\P}'(Y'{=}\bad|X_m,A_0{=}\risky) \leq p^*}
            \\
            &\phantom{=\E^*[}
            +
            \ell(Y',\safe) \Ind{\overline{\P}'(Y'{=}\bad|X_m,A_0{=}\risky) > p^*} | A_0{=}\risky] \overline{\P}'(A_0{=}\risky)
            \\
            &\phantom{=}
            +
            \frac{c_I c_{II}}{c_I + c_{II}} \overline{\P}'(A_0{=}\safe)
            \\
            &=
            \overline{\E}'[\ell(Y',\risky) \Ind{\overline{\rec}'(X_m) {=} \neutral} + \ell(Y',\safe) \Ind{\overline{\rec}'(X_m) {=} \safe}|A_0{=}\safe]
            \:
            \overline{\P}'(A_0{=}\safe)
            \\
            &\phantom{=}+
            \frac{c_I c_{II}}{c_I + c_{II}}
            \overline{\P}'(A_0{=}\safe) = \overline{L}'(\overline{\P}').
        \end{align*}
        As a consequence, minimax loss is $\overline{L}'(\overline{\P}')$, which is guaranteed by the recommendation policy $\overline{\rec}'$.

        The above argument assumes that the human decision-maker takes a non-dominated baseline decision even for the part of the distribution where $Y$ is not observed, implying that $\E[\ell(Y,\safe)|A_0{=}\safe] \leq \E[\ell(Y,\risky)|A_0{=}\safe]$ and thus $\E[\ell(Y,\safe)|A_0{=}\safe] \leq \frac{c_I c_{II}}{c_I + c_{II}}$.
        But the same result goes through even when the agent may have wrong assumptions about $\P(Y{=}\bad|X_m,A_0{=}\safe)$.
        Specifically, assume that the agent assumes that $\P(Y{=}\bad|X_m,A_0{=}\safe) = 1$, but really $\P(Y{=}\bad|X_m,A_0{=}\safe)=0$, then the agent always chooses $A{=}\safe$ no matter the recommendations, leading to a loss of $\ell(\good,\safe) = c_I$ to the principal.
        (We could also consider the case where the agent assumes $\P(Y{=}\bad|X_m,A_0{=}\safe) = 0$ where really $\P(Y{=}\bad|X_m,A_0{=}\safe) = 1$, but that would render the baseline decision of $A_0{=}\safe$ inconsistent with the agent's belief.)
        The worst-case loss for the instances with $A_0{=}\safe$ under any recommendation algorithm is then $\E[\ell(Y,\safe)|A_0{=}\safe] = c_I$, and the result in the proposition goes through with
        \begin{align*}
            \overline{L}'(\overline{\P}') &=
            \overline{\E}'[\ell(Y',\risky) \Ind{\overline{\rec}'(X_m) {=} \neutral} + \ell(Y',\safe) \Ind{\overline{\rec}'(X_m) {=} \safe}|A_0{=}\safe]
            \:
            \overline{\P}'(A_0{=}\safe)
            \\
            &\phantom{=}+
            c_I
            \overline{\P}'(A_0{=}\safe)
        \end{align*}
        as the minimax expected loss guaranteed by $\overline{\rec}'$.
        (If we allow for baseline choices that are inconsistent with the agent's belief, then the bound would similarly include $\max\{c_I,c_{II}\}$ instead of $c_I$.)
    \end{proof}

    \begin{proof}[Proof of \autoref{prop:Onesidedminimax}]
        $\overline{\rec}''$ falls within the more general family of policies detailed by \autoref{prop:Onesidedminimaxgeneral}, where we set $p^\low = p^*, p^\high \geq 1$.
    \end{proof}
    
    \begin{proof}[Proof of \autoref{prop:Delete}]
        For the comparison to the machine decision, note that the optimal machine-only decision (assuming ties are broken in favor of the risky decision) is given by
        \[
            \argmin_a \E[\ell(Y,a)|X_m]
            \ni
            A^*
            =
            \begin{cases}
                \risky, & \widehat{Y} \leq p^* = \frac{c_I}{c_I + c_{II}}, \\
                \safe, & \widehat{Y} > p^*.
            \end{cases}
        \]
        Similarly to the proof of \autoref{prop:Complementarity},
        for the action $A$ chosen by the agent to be different from $A^*$, we must have that
        \begin{align*}
            &\P(Y{=}\bad|X_h,A^*{=}\risky)
            =
            \P(Y{=}\bad|X_h,\widehat{Y} \leq p^*)
            \\
            &\geq
            \frac{c_I + \delta_I(\widehat{Y})}{c_I + c_{II} + \delta_I(\widehat{Y}) + \delta_{II}(\widehat{Y})}
            \geq
            \frac{c_I + \delta_I(p^*)}{c_I + c_{II} + \delta_I(p^*) + \delta_{II}(p^*)} = p^*
            \:\:\:
            (\safe  = A \neq A^* =\risky),
            \\
                        &\P(Y{=}\bad|X_h,A^*{=}\risky)
            =
            \P(Y{=}\bad|X_h,\widehat{Y} > p^*)
            \\
            &\leq
            \frac{c_I + \delta_I(\widehat{Y})}{c_I + c_{II} + \delta_I(\widehat{Y}) + \delta_{II}(\widehat{Y})}
            \leq
            \frac{c_I + \delta_I(p^*)}{c_I + c_{II} + \delta_I(p^*) + \delta_{II}(p^*)} = p^*
            \:\:\:
            (\risky  = A \neq A^* =\safe),
        \end{align*}
        where we have used monotonicity and that $p^*$ is recommendation-neutral.
        Hence,
        \begin{align*}
            \P(Y{=}\bad|\safe  {=} A {\neq} A^* {=}\risky)
            &=
            \E[\P(Y{=}\bad|X_h,A^*{=}\risky)|\safe  {=} A {\neq} A^* {=}\risky] \geq p^*,
            \\
            \P(Y{=}\bad|\risky  {=} A {\neq} A^* {=}\safe)
            &=
            \E[\P(Y{=}\bad|X_h,A^*{=}\safe) |\risky  {=} A {\neq} A^* {=}\safe]
            \leq p^*
        \end{align*}
        and thus
        \begin{align*}
            \E[\ell(Y,A)]
            &=
            \E[\ell(Y,A) \Ind{A{=}A^*}] + \E[\ell(Y,A) \Ind{\safe  {=} A {\neq} A^* {=}\risky}] + \E[\ell(Y,A) \Ind{\risky  {=} A {\neq} A^* {=}\safe}]
            \\
            &\leq
            \E[\ell(Y,A) \Ind{A{=}A^*}]
            +  c_I (1-p^*) \E[\Ind{\safe  {=} A {\neq} A^* {=}\risky}] + c_{II} p^*
            \E[\Ind{\risky  {=} A {\neq} A^* {=}\safe}]
            \\
            &=
            \E[\ell(Y,A) \Ind{A{=}A^*}]
            +  c_{II} p^* \E[\Ind{\safe  {=} A {\neq} A^* {=}\risky}] + c_{I} (1-p^*)
            \E[\Ind{\risky  {=} A {\neq} A^* {=}\safe}]
            \\
            &\leq
            \E[\ell(Y,A^*) \Ind{A{=}A^*}] + \E[\ell(Y,A^*) \Ind{\safe  {=} A {\neq} A^* {=}\risky}] + \E[\ell(Y,A^*) \Ind{\risky  {=} A {\neq} A^* {=}\safe}]
            \\
            &= \E[\ell(Y,A^*)] = \E[\min_a \E[\ell(Y,a)|X_m]].
        \end{align*}

        For the comparison to human decisions,
        choosing $p^\low \equiv 1, p^\high \equiv 0$ means that the score is always withheld and interpreted as $\P(Y{=}\bad)$.
        Since $\P(Y{=}\bad)$ is recommendation-neutral and does not contain any new information, it does not affect the final action, so it leads to the same decision as the human-only decision.
        Hence, $\E[\ell(Y,A)] \leq \E[\min_a \E[\ell(Y,a)|X_h]]$ for this recommendation.

        For the comparison to not withholding the risk score, note that always sending $\P(Y{=}\bad|X_m)$ is a special case with $p^\low = p^* = p^\high$, so optimal choices of $p^\low, p^\high$ (that minimize $\E[\ell(Y,A)]$) are guaranteed to weakly improve over that baseline as well.
    \end{proof}

    \begin{proof}[Proof of \autoref{prop:refdep}]
            For fixed $x_h,r$, write $p = \P(Y{=}\bad|X_h{=}x_h,R{=}r)$.
            For 2., we have that:
            \begin{align*}
                \E[\ell(Y,a|L_I,L_{II})|X_h{=}x_h,R{=}r]
                =
                \begin{cases}
                    \begin{cases}
                    \smash{\overbrace{p ( - c_I (1-p))
                    +
                    (1-p) \lambda (c_I - c_I (1-p))}^{=p(1-p) c_I (\lambda - 1)}}, &a {=} \safe \\
                    p \lambda c_{II}
                    - c_I (1-p), &a {=} \risky
                    \end{cases}
                    & r {=} \safe \\
                    \begin{cases}
                    (1-p) \lambda c_I - c_{II} p, &a {=} \safe \\
                    \smash{\underbrace{(1-p) ( - c_{II} p)
                    +
                    p \lambda (c_{II} - c_{II} p)}_{=p(1-p) c_{II} (\lambda - 1)}}
                        , &a {=} \risky
                    \end{cases}
                    & r {=} \risky
                \end{cases}
            \end{align*}
            For 3., we directly have that:
            \begin{align*}
                \E[\ell(Y,a|L_I,L_{II})|X_h{=}x_h,R{=}r]
                =
                \begin{cases}
                    \begin{cases}
                    p(1-p) c_I (\lambda - 1), &a {=} \safe \\
                    p \lambda c_{II}
                    - c_I (1-p), &a {=} \risky
                    \end{cases}
                    & r {=} \safe \\
                    \begin{cases}
                    (1-p) \lambda c_I - c_{II} p, &a {=} \safe \\
                    p(1-p) c_{II} (\lambda - 1)
                        , &a {=} \risky
                    \end{cases}
                    & r {=} \risky
                \end{cases}
            \end{align*}
            In particular, the two expected losses are the same.

            The optimal decision of the agent is to choose the risky action whenever $p \leq p^r$ for decision thresholds $p^r$ that differ by the recommendation, and are defined by the indifference conditions
            \begin{align*}
                p^\safe (1-p^\safe) c_I (\lambda - 1) &= p^\safe \lambda c_{II}
                    - c_I (1-p^\safe),
                \\
                p^\risky (1-p^\risky) c_{II} (\lambda - 1) &= (1-p^\risky) \lambda c_{I}
                    - c_{II} p^\risky.
            \end{align*}
            Solving the quadratic explicitly, we obtain solutions $p^\safe = p^\safe(\lambda), p^\risky = p^\risky(\lambda)$ that are monotonically decreasing for $p^\safe(\lambda)$ and increasing for $p^\risky(\lambda)$, with $p^\safe(1) = \frac{c_I}{c_I + c_{II}} = p^\risky(1)$.
            Substituting $p^\safe = \frac{c_I}{c_I + c_{II} + \Delta_{II}}, p^\risky = \frac{c_I + \Delta_I}{c_I + c_{II} + \Delta_{I}}$, we find the claimed solutions.
            In the special case where $c_I = 1 = c_{II}$,
            $p^\risky(\lambda) = \frac{1}{1 + \sqrt{\lambda}}, p^\safe(\lambda) = \frac{\sqrt{\lambda}}{1 + \sqrt{\lambda}}$,
            so $\Delta_I(\lambda) = \sqrt{\lambda} - 1 = \Delta_{II}(\lambda)$.
        \end{proof}

    \newcommand{\tH}{\tilde{H}}
    \newcommand{\tM}{\tilde{M}}
    \renewcommand{\th}{\tilde{h}}
    \newcommand{\tm}{\tilde{m}}
    \newcommand{\tr}{\tilde{r}}

    For the following proofs that rely on Assumptions~\ref{asm:independent}--\ref{asm:monotonicity}, we note that we can consider all statements to be a.s.\ conditional on $Z_0$ (and omitting $Z_0$ in our notation), since principal and agent have access to $Z_0$ and all policies are allowed to depend on its realization.
    Furthermore, writing $\tH = \phi_h(Z_h;Z_0)$ and $\tM = \phi_m(Z_m;Z_0)$, we obtain the representation
    \begin{align*}
        \P(Y{=}\bad|X_h,X_m) &= \P(Y{=}\bad|\tH,\tM),
        &
        \P(Y{=}\bad|\tH{=}\th,\tM{=}\tm)
        &= \phi(\th,\tm)
    \end{align*}
    for $\th$ and $\tm$ in the support of $\tH$ and $\tM$, respectively,
    with $f$ monotonically increasing in both (scalar) arguments and $\tH$ independent of $\tM$ and $R$ (conditonal on $Z_0$).
    This notation improves the readability of the following proofs, and we maintain it throughout.

    \begin{proof}[Proof of \autoref{prop:Agent-Threshold}]
        The optimal agent action is almost surely given by those in \autoref{rem:Thresholds} (where ties are broken in favor of the risky decision), so our goal is to show that there are functions $h^\risky,h^\safe$ such that
        for all $\th$ in the support of $\tH$ and all $\tr \in \{\risky,\safe\}$,
        \begin{align*}
            \P(Y{=}\bad|\tH{=}\th,R{=}\tr) &\leq p^{\tr}
            &
            &\Longleftrightarrow
            &
            \P(Y{=}\bad|\tH{=}\th) &\leq h^{\tr}.
        \end{align*}
        Here, we invoke the notation from above this proof, and assume that $\P(R = \risky), \P(R = \safe) > 0$, since the case where $\P(R = \risky) = 0$ or $\P(R = \safe) = 0$ is trivial.
        Note first that almost surely
        \[
            \P(Y{=}\bad|\tH{=}\th,R{=}\tr) =
            \E[\phi(\tH,\tM)|\tH{=}\th,R{=}\tr]
            =
            \E[\phi(\th,\tM)|R{=}\tr],
        \]
        and the right-hand side is monotonically increasing in $\th$ by independence and monotonicity of $f$, and the same holds for
        \begin{align*}
            &\P(Y{=}\bad|\tH{=}\th)
            = \E[\phi(\tH,\tM)|\tH{=}\th] = \E[\phi(\th,\tM)]
            \\
            &=
            \E[\phi(\th,\tM)|R{=}\risky] \: \P(R{=}\risky)
            +
            \E[\phi(\th,\tM)|R{=}\safe] \: \P(R{=}\safe)
        \end{align*}
        where we have used independence of $\tH$ and $R$ (a.s.\ conditional on $Z_0$, which is implicit here).
        As a consequence, for all $\th_1,\th_2$ in the support of $\tH$, all $\tr \in \{\risky,\safe\}$, and all $\varepsilon > 0$,
        \begin{align*}
            &&
            \E[\phi(\th_1,\tM)|R{=}\tr] + \varepsilon
            &\leq \E[\phi(\th_2,\tM)|R{=}\tr]
            \\
            &\Longrightarrow 
            &
            \E[\phi(\th_1,\tM)] + \varepsilon \: \P(R{=}\tr)
            &\leq \E[\phi(\th_2,\tM)]
        \end{align*}
        since from the left it also follows that $\th_1 < \th_2$ and thus $\P(Y{=}\bad|\tH{=}\th_1,R{=}\tr')
        \leq \P(Y{=}\bad|\tH{=}\th_2,R{=}\tr')$ for the other $\tr' \neq \tr$ (while the opposite implication does not generally hold).
        Let now
        \begin{equation}
            \label{eqn:Cutoffdef}
            h^{\tr} = \textstyle\sup_{\:\th \text{ in the support of } \tH; \: \E[\phi(\th,\tM)|R{=}\tr] \leq p^{\tr}\:} \E[\phi(\th,\tM)]
        \end{equation}
        where we define the supremum over the empty set as 0.
        We have that
        \begin{align*}
            \underbrace{\P(Y{=}\bad|\tH{=}\th,R{=}\tr)}_{=\E[\phi(\th,\tM)|R{=}\tr]} &\leq p^{\tr}
            &
            &\Longrightarrow
            &
            \underbrace{\P(Y{=}\bad|\tH{=}\th)}_{=\E[\phi(\th,\tM)} &\leq h^{\tr}.
        \end{align*}
        by the definition of $h^{\tr}$
        and
        \begin{align*}
            &&
            &\P(Y{=}\bad|\tH{=}\th,R{=}\tr) > p^{\tr}
            \\
            &\Longrightarrow
            &
            &
            \exists \varepsilon > 0:
            \:
            \E[\phi(\th,\tM)|R{=}\tr] \geq \E[\phi(\th',\tM)] + \varepsilon
            \\
            &&
            &\phantom{\exists \varepsilon > 0:
            \:}\forall \th' \text{ in the support of } \tH \text{ with } \E[\phi(\th',\tM)|R{=}\tr] \leq p^{\tr}
            \\
            &\Longrightarrow
            &
            &
            \exists \varepsilon > 0:
            \:
            \E[\phi(\th,\tM)] \geq \E[\phi(\th',\tM)] + \varepsilon
            \\
            &&
            &\phantom{\exists \varepsilon > 0:
            \:}\forall \th' \text{ in the support of } \tH \text{ with } \E[\phi(\th',\tM)|R{=}\tr] \leq p^{\tr}
            \\
            &\Longrightarrow
            &
            &\P(Y{=}\bad|\tH{=}\th) > h^{\tr}.
        \end{align*}
        Hence
        \begin{align*}
            \P(Y{=}\bad|\tH{=}\th,R{=}\tr) &\leq p^{\tr}
            &
            &\Longleftrightarrow
            &
            \P(Y{=}\bad|\tH{=}\th) &\leq h^{\tr}.
            \qedhere
        \end{align*}
    \end{proof}

    \begin{proof}[Proof of \autoref{prop:Agent-Thresholdordering}]

        We employ the simplified notation above the proof of \autoref{prop:Agent-Threshold} (and condition on $Z_0$ throughout).
        Consider decision thresholds
        \begin{equation}
            \label{eqn:Cutoffdef2}
            \begin{aligned}
            h^{\safe}_0(p,m) &= \textstyle\sup_{\:\th \text{ in the support of } \tH; \: \E[\phi(\th,\tM)|\P(Y{=}\bad|X_m) > m] \leq p\:} \E[\phi(\th,\tM)],
            \\
            h^{\risky}_0(p,m) &= \textstyle\sup_{\:\th \text{ in the support of } \tH; \: \E[\phi(\th,\tM)|\P(Y{=}\bad|X_m) \leq m] \leq p\:} \E[\phi(\th,\tM)]
            \end{aligned}
        \end{equation}
        defined by \eqref{eqn:Cutoffdef} in the proof of \autoref{prop:Agent-Threshold} for the threshold recommendations from \eqref{eqn:Principal-Threshold},
        where the supremum over the empty set is again 1.
        In addition, define the analogous threshold
        \begin{equation}
            h^*_0(p) = \textstyle\sup_{\:\th \text{ in the support of } \tH; \: \E[\phi(\th,\tM)] \leq p\:} \E[\phi(\th,\tM)]
        \end{equation}
        for decisions that do not use machine input.
        By the proof \autoref{prop:Agent-Threshold},
        we obtain thresholds
        with
        \begin{align*}
            \P(Y{=}\bad|\tH{=}\th,\P(Y{=}\bad|\tM){\leq} m)
            &\leq p^{\risky}
            &
            &\Longleftrightarrow
            &
            \P(Y{=}\bad|\tH{=}\th) &\leq h^{\risky}_0(p^{\risky},m),
            \\
            \P(Y{=}\bad|\tH{=}\th,\P(Y{=}\bad|\tM) {>} m)
            &\leq p^{\safe}
            &
            &\Longleftrightarrow
            &
            \P(Y{=}\bad|\tH{=}\th) &\leq h^{\safe}_0(p^{\safe},m)
            \\
            \P(Y{=}\bad|\tH{=}\th)
            &\leq p^*
            &
            &\Longleftrightarrow
            &
            \P(Y{=}\bad|\tH{=}\th) &\leq h^*_0(p^*).
        \end{align*}
        By construction, the $h^{\safe}_0(p,m), h^{\safe}_0(p,m), h^*_0(p)$ are monotonically increasing in $p$.
        By monotonicity of $f$ and independence of $\tH$ and $\tM$, we also have that
        \begin{align*}
            \P(Y{=}\bad| \tilde{H}=\th, \P(Y{=}\bad|\tilde{M}) {>} m)
            &=
            \E[
                \phi(\th,\tM)
                |
                \E[\phi(\tH,\tM)|\tM] {>} m
            ],
            \\
            \P(Y{=}\bad| \tilde{H}=\th, \P(Y{=}\bad|\tilde{M}) {\leq} m)
            &=
            \E[
                \phi(\th,\tM)
                |
                \E[\phi(\tH,\tM)|\tM] {\leq} m
            ]
        \end{align*}
        are monotonically increasing in $X_m$,
        and
        $h^{\safe}_0(p,m), h^{\risky}_0(p,m)$ thus monotonically decreasing in $X_m$.
        Finally,
        $h^{\safe}_0(p,1) = h^*_0(p)$ and $h^{\risky}_0(p,0) \geq h^*_0(p)$.
        As a consequence, using $p^{\risky} \geq p^* \geq p^{\safe}$,
        \[
            h^{\risky}_0(p^{\risky},m)
            \geq
            h^{\risky}_0(p^*,m)
            \geq
            h^{\risky}_0(p,0)
            \geq
            h^*_0(p^*)
            =
            h^{\safe}_0(p^*,1)
            \geq
            h^{\safe}_0(p^*,m)
            \geq
            h^{\safe}_0(p^{\safe},m).
        \]

        As a last step, we now need to transform thresholds
        $h^{\risky}_0(p^{\risky},m) \geq  h^*_0(p^*) \geq h^{\safe}_0(p^{\safe},m)$
        into thresholds
        $h^{\risky}(p^{\risky},m) \geq  h^*(p^*) \geq h^{\safe}(p^{\safe},m)$
        with $h^*(p^*) = p^*$ (where we note that $h^*_0(p^*) \leq p^*$, but the inequality can be strict).
        To this end for $h \in [0,1]$ let
        \[
            h^{-1}(h)
            =
            \begin{cases}
                p^*, &
                h \leq p^* < p \text{ for all } p \text{ with } h_0^*(p) > h,
                \\
                h,
                & \text{otherwise}.
            \end{cases}
        \]
        for which $h^{-1}(h_0^*(p^*)) = p^*$ since $h_0^*(p^*) \leq p^*$ and whenever $h_0^*(p) > h_0^*(p^*)$ we must have that $p > p^*$ by monotonicity of $h_0^*$.
        
        First, $h^{-1}$ is monotonically increasing on $[0,1]$.
        Indeed, this is straightforward for any $h_1,h_2$ that either both fulfill or both do not fulfill the condition in the first line.
        For the remaining case, assume that $h_1 \leq p^* < p$ for all $p$ with $h_0^*(p) > h_1$ (which implies $h^{-1}(h_1) = p^*$), while $h_2 > p^*$ or there is some $p_2 \leq p$ with $h_0^*(p_2) > h_2$ (both of which imply $h^{-1}(h_2) = h_2$).
        If $h_2 > p^*$ then $h_2 > p^* \geq h_1$ and $h^{-1}(h_2) > p^* = h^{-1}(h_1)$, so monotonicity holds.
        If $h_2 \leq p^*$ and such a $p_2$ exists then we must have $h_0^*(p_2) \leq h_1$; with $h_0^*(p_2) > h_2$ this implies $h_1 > h_2$ and $h^{-1}(h_1) = p^* \geq h_2 = h^{-1}(h_2)$, so monotonicity holds again.
        
        Second,
        \begin{align*}
            \P(Y{=}\bad|\tH{=}\th) &\leq h
            &
            &\Longleftrightarrow
            &
            \P(Y{=}\bad|\tH{=}\th) &\leq h^{-1}(h),
        \end{align*}
        where $\Longrightarrow$ follows from $h^{-1}(h) \geq h$.
        For $\Longleftarrow$ we note that $h^*_0(h^{-1}(h)) \leq h$, which holds for $h^{-1}(h) = h$ by $h^*_0(p) \leq p$
        and otherwise since
        $h^*_0(p^*) \leq h$ for all $h \leq p^*$ such that $h_0^*(p) > h$ implies that $p > p^*$, since in this case for all $p \leq p^*$ it follows that $h_0^*(p) \leq h$.
        Hence, from $\P(Y{=}\bad|\tH{=}\th) \leq h^{-1}(h)$ we obtain $\P(Y{=}\bad|\tH{=}\th) \leq h_0^*(h^{-1}(h))$ by the properties of $h_0^*$ and thus $\P(Y{=}\bad|\tH{=}\th) \leq h$ from $h_0^*(h^{-1}(h)) \leq h$.

        As a consequence of these properties of $h^{-1}$ along with those of $h^{\risky}_0, h^*_0, h^{\safe}_0$, we can define
        \begin{align*}
            h^{\risky}(p^{\risky},m)
            &=
            h^{-1}(h^{\risky}_0(p^{\risky},m)),
            &
            h^{\safe}(p^{\safe},m)
            &=
            h^{-1}(h^{\safe}_0(p^{\safe},m))
        \end{align*}
        for which
        \begin{align*}
            &\P(Y{=}\bad|\tH{=}\th,\P(Y{=}\bad|\tM){\leq} m)
            \leq p^{\risky}
            \\
            &\Longleftrightarrow
            \P(Y{=}\bad|\tH{=}\th) \leq h^{\risky}_0(p^{\risky},m)
            \Longleftrightarrow
            \P(Y{=}\bad|\tH{=}\th) \leq h^{\risky}(p^{\risky},m),
            \\
            &\P(Y{=}\bad|\tH{=}\th,\P(Y{=}\bad|\tM) {>} m)
            \leq p^{\safe}
            \\
            &\Longleftrightarrow
            \P(Y{=}\bad|\tH{=}\th) \leq h^{\safe}_0(p^{\safe},m)
            \Longleftrightarrow
            \P(Y{=}\bad|\tH{=}\th) \leq h^{\safe}(p^{\safe},m)
        \end{align*}
        and
            $h^{\risky}(p^{\risky},m) \geq h^*(p^*) \geq h^{\safe}(p^{\safe},m)$
        by monotonicity.
    \end{proof}

    \begin{proof}[Proof of \autoref{prop:Agent-Thresholdcomparative}]

        For a fixed threshold $X_m$, the thresholds constructed in the proof of \autoref{prop:Agent-Thresholdordering} are monotonically increasing in $p^{\risky}$ and $p^{\safe}$, respectively.
        Since $p^{\risky} = \frac{c_I + \Delta_I}{c_I + c_{II} + \Delta_I}$ is monotonically increasing in $\Delta_I$
        and $p^{\safe} = \frac{c_I}{c_I + c_{II} + \Delta_{II}}$ is monotonically decreasing in $\Delta_{II}$,
        these thresholds have the desired properties.

        Similarly, for fixed thresholds $p^{\risky}$ and $p^{\safe}$, the thresholds constructed in the proof of \autoref{prop:Agent-Thresholdordering} are similarly monotonically decreasing in $X_m$, since monotonicity holds for $h^{\safe}_0(p,m), h^{\risky}_0(p,m)$ by construction.

    \end{proof}

    \begin{proof}[Proof of \autoref{prop:DecisionvsRec}]
        An instance is provided by \autoref{exm1}.
    \end{proof}

    \begin{proof}[Proof of \autoref{prop:Localmonotonicity}]

        Using the simplified notation from above the proof of \autoref{prop:Agent-Threshold},
        note that we can express (by monotonicity of $\E[\phi(\th,\tM)],\E[\phi(\tH,\tm)]$) the threshold-based policies by the agent and the principal as
        \begin{align*}
            R &=
            \begin{cases}
                \risky, & \tM \leq \bar{m}, \\
                \safe, & \tM > \bar{m},
            \end{cases}
            &
            A &=
            \begin{cases}
                \risky, & \tH \leq \bar{h}^R, \\
                \safe, & \tH > \bar{h}^R.
            \end{cases}
        \end{align*}
        Given thresholds $\bar{m},\bar{h}^{\risky},\bar{h}^{\safe}$,
        expected losses of principal and agent are
        \begin{align*}
            &L(\bar{m},\bar{h}^{\risky},\bar{h}^{\safe})
            =
            \E[\ell(Y,A)]
            \\
            &=
            \E[\Ind{\tM {\leq} \bar{m}, \tH {\leq} \bar{h}^{\risky}} \phi(\tH,\tM)] c_{II}
            + \E[\Ind{\tM {\leq} \bar{m}, \tH {>} \bar{h}^{\risky}} (1{-}\phi(\tH,\tM))] c_I
            \\
            &\phantom{{}={}}
            + \E[\Ind{\tM {>} \bar{m}, \tH {\leq} \bar{h}^{\safe}} \phi(\tH,\tM)] c_{II}
            + \E[\Ind{\tM {>} \bar{m}, \tH {>} \bar{h}^{\safe}} (1{-}\phi(\tH,\tM))] c_I,
            \\
            &L^*(\bar{m},\bar{h}^{\risky},\bar{h}^{\safe})
            =
            \E[\ell^*(Y,A,R)] = L(\bar{m},\bar{h}^{\risky},\bar{h}^{\safe})
            \\
            &\phantom{{}={}}
            +  \E[\Ind{\tM {\leq} \bar{m}, \tH {>} \bar{h}^{\risky}} (1{-}\phi(\tH,\tM))] \Delta_I + \E[\Ind{\tM {>} \bar{m}, \tH {\leq} \bar{h}^{\safe}} \phi(\tH,\tM)] \Delta_{II}.
        \end{align*}
        The optimal agent thresholds $\bar{h}^{\risky}_{\Delta_I}(\bar{m}), \bar{h}^{\safe}_{\Delta_{II}}(\bar{m})$ minimize $L^*(\bar{m},\bar{h}^{\risky},\bar{h}^{\safe})$ given $\bar{m}$,
        which yields 
        the first-order conditions
        \begin{align*}
            \E[\phi(\bar{h}^{\risky},\tM)|\tM \leq \bar{m}] &= \frac{c_I {+} \Delta_I}{c_I {+} c_{II} {+} \Delta_I}
            ,
            &
            \E[\phi(\bar{h}^{\safe},\tM)|\tM > \bar{m}] &= \frac{c_I}{c_I {+} c_{II} {+} \Delta_{II}}
            .
        \end{align*}
        with unique solutions $\bar{h}^{\risky}_{\Delta_I}(\bar{m}) > \bar{h}^{\safe}_{\Delta_{II}}(\bar{m})$ by monotonicity of $f$ and our regularity assumptions,
        which by the implicit function theorem are continuously differentiable in $\bar{m}$
        with
        \begin{align*}
            \frac{\partial}{\partial \bar{m}} \bar{h}^{\risky}_{\Delta_I}(\bar{m}) 
            &
            =
            \mu_M(\bar{m}) \frac{
                \frac{c_I {+} \Delta_I}{c_I {+} c_{II} {+} \Delta_I}
                -
                \phi(\bar{h}^{\risky}_{\Delta_I}(\bar{m}), \bar{m})
            }{
                \E[
                \sfrac{\partial f}{\partial \tilde{h}}(\bar{h}^{\risky}_{\Delta_I}(\bar{m}), \tM) \Ind{\tM \leq \bar{m}}]
            }
            < 0,
            \\
            \frac{\partial}{\partial \Delta_I} \bar{h}^{\risky}_{\Delta_I}(\bar{m})
            &
            =
            \frac{
                \E[(1 - \phi(\bar{h}^{\risky}_{\Delta_I}(\bar{m}), \tM)) \Ind{\tM \leq \bar{m}}]
            }{
                (c_I{+}c_{II}{+}\Delta_I) 
                \E[
                    \sfrac{\partial f}{\partial \tilde{h}}(\bar{h}^{\risky}_{\Delta_I}(\bar{m}), \tM) \Ind{\tM \leq \bar{m}}]
            }
            > 0,
            \\
            \frac{\partial}{\partial \bar{m}} \bar{h}^{\safe}_{\Delta_{II}}(\bar{m})
            &
            =
            \mu_M(\bar{m}) \frac{
                \phi(\bar{h}^{\safe}_{\Delta_{II}}(\bar{m}), \bar{m})
                -
                \frac{c_I}{c_I {+} c_{II} {+} \Delta_{II}}
            }{
                \E[
                \sfrac{\partial f}{\partial \tilde{h}}(\bar{h}^{\safe}_{\Delta_{II}}(\bar{m}), \tM) \Ind{\tM > \bar{m}}]
            }
            < 0,
            \\
            \frac{\partial}{\partial \Delta_{II}} \bar{h}^{\safe}_{\Delta_{II}}(\bar{m}) 
            &
            =
            \frac{
                - \E[\phi(\bar{h}^{\safe}_{\Delta_{II}}(\bar{m}), \tM)) \Ind{\tM > \bar{m}}]
            }{
                (c_I{+}c_{II}{+}\Delta_{II}) 
                \E[
                    \sfrac{\partial f}{\partial \tilde{h}}(\bar{h}^{\safe}_{\Delta_{II}}(\bar{m}), \tM) \Ind{\tM > \bar{m}}]
            }
            < 0.
        \end{align*}
        The optimal principal threshold $\bar{m}^*_{\Delta_I,\Delta_{II}}$ then minimizes $L(\bar{m}, \bar{h}^{\risky}_{\Delta_I}(\bar{m}), \bar{h}^{\safe}_{\Delta_{II}}(\bar{m}))$.

        Writing
        $\frac{\d L}{\d \bar{m}}$
        for the (total) derivative of 
        $L(\bar{m}, \bar{h}^{\risky}_{\Delta_I}(\bar{m}), \bar{h}^{\safe}_{\Delta_{II}}(\bar{m}))$ with respect to $\bar{m}$
        and $\mu_M,\mu_H$ for the density functions of $\tM,\tH$, respectively,
        we have that
        \begin{align*}
            \frac{\d L}{\d \bar{m}}
            &=
            \frac{\partial L}{\partial \bar{m}}
            +
            \frac{\partial \bar{h}^{\risky}}{\partial \bar{m}}
            \frac{\partial L}{\partial \bar{h}^{\risky}} 
            +
            \frac{\partial \bar{h}^{\safe}}{\partial \bar{m}}
            \frac{\partial L}{\partial \bar{h}^{\safe}}
            \\
            &=
            \frac{\partial L}{\partial \bar{m}}
            +
            \frac{\partial \bar{h}^{\risky}}{\partial \bar{m}}
            \overbrace{\frac{\partial L^*}{\partial \bar{h}^{\risky}} }^{=0}
            +
            \Delta_I
            \frac{\partial \bar{h}^{\risky}}{\partial \bar{m}}
            \frac{\partial}{\partial \bar{h}^{\risky}}
            \E[\Ind{\tM {\leq} \bar{m}, \tH {>} \bar{h}^{\risky}_{\Delta_I}(\bar{m})} (1{-}\phi(\tH,\tM))]
            \\
            &\phantom{{}=
            \frac{\partial L}{\partial \bar{m}}
            {}}
            +
            \frac{\partial \bar{h}^{\safe}}{\partial \bar{m}}
            \underbrace{\frac{\partial L^*}{\partial \bar{h}^{\safe}}}_{=0}
            +
            \Delta_{II}
            \frac{\partial \bar{h}^{\safe}}{\partial \bar{m}}
            \frac{\partial}{\partial \bar{h}^{\safe}}
            \E[\Ind{\tM {>} \bar{m}, \tH {\leq} \bar{h}^{\safe}_{\Delta_{II}}(\bar{m})} (1{-}\phi(\tH,\tM))]
            \\
            &=
            \mu_M(\bar{m}) \E[\Ind{\bar{h}^{\safe}_{\Delta_{II}}(\bar{m}) {<} \tH {\leq} \bar{h}^{\risky}_{\Delta_I}(\bar{m})} ((c_I + c_{II}) \phi(\tH,\bar{m}) - c_I)]
            \\
            &\phantom{{}={}}
            -
            \Delta_I \frac{\partial \bar{h}^{\risky}}{\partial \bar{m}}
            \mu_H(\bar{h}^{\risky}_{\Delta_I}(\bar{m})) \E[\Ind{\tM {\leq} \bar{m}} (1{-}\phi(\bar{h}^{\risky}_{\Delta_I}(\bar{m}),\tM))]
            \\
            &\phantom{{}={}}
            +
            \Delta_{II} \frac{\partial \bar{h}^{\safe}
            }{\partial \bar{m}}
            \mu_H(\bar{h}^{\safe}_{\Delta_{II}}(\bar{m})) \E[\Ind{\tM {>} \bar{m}} \phi(\bar{h}^{\safe}_{\Delta_{II}}(\bar{m}),\tM)] 
            =
            F_{\Delta_I,\Delta_{II}}(\bar{m}),
        \end{align*}
        where $F_{\Delta_I,\Delta_{II}}(\bar{m})$ is continuously differentiable in $\bar{m},\Delta_I,\Delta_{II}$
        with
        \begin{align*}
            \frac{\partial}{\partial \Delta_I} F_{0,0}(\bar{m})
            &=
            \overbrace{\mu_M(\bar{m}) \mu_H(\bar{h}^{\risky}(\bar{m}))}^{>0} \overbrace{((c_I + c_{II}) \phi(\bar{h}^{\risky}_{\Delta_I}(\bar{m}),\bar{m}) - c_I)}^{>0}
            \overbrace{\frac{\partial \bar{h}^{\risky}}{\partial \Delta_I}}^{>0}
            \\
            &\phantom{{}={}}
            - 
            \underbrace{\frac{\partial \bar{h}^{\risky}}{\partial \bar{m}}}_{<0}
            \underbrace{\mu_H(\bar{h}^{\risky}_{\Delta_I}(\bar{m})) \E[\Ind{\tM {\leq} \bar{m}} (1{-}\phi(\bar{h}^{\risky}_{\Delta_I}(\bar{m}),\tM))]}_{>0} > 0,
            \\
            \frac{\partial}{\partial \Delta_{II}} F_{0,0}(\bar{m})
            &=
            - \overbrace{\mu_M(\bar{m}) \mu_H(\bar{h}^{\safe}(\bar{m}))}^{>0} \overbrace{((c_I + c_{II}) \phi(\bar{h}^{\safe}_{\Delta_{II}}(\bar{m}),\bar{m}) - c_I)}^{<0}
            \overbrace{\frac{\partial \bar{h}^{\safe}}{\partial \Delta_{II}}}^{<0}
            \\
            &\phantom{{}={}}
            +
            \underbrace{\frac{\partial \bar{h}^{\safe}}{\partial \bar{m}}}_{<0}
            \underbrace{\mu_H(\bar{h}^{\safe}_{\Delta_{II}}(\bar{m})) \E[\Ind{\tM {>} \bar{m}} \phi(\bar{h}^{\safe}_{\Delta_{II}}(\bar{m}),\tM)]}_{>0} < 0,
        \end{align*}

        The optimal threshold $\bar{m}^*_{\Delta_I,\Delta_{II}}$ fulfils the first-order condition $F_{\Delta_I,\Delta_{II}}(\bar{m}^*_{\Delta_I,\Delta_{II}}) = 0$.
        Furthermore, by assumption, the solution at $\Delta_I = 0 = \Delta_{II}$ is unique with $\frac{\partial}{\partial \bar{m}} F_{0,0}(\bar{m}^*_{0,0}) > 0$.
        By the implicit function theorem,
        there is a neighborhood of $\Delta_I = 0 = \Delta_{II}$ in which $\bar{m}^*_{\Delta_I,\Delta_{II}}$ is continuously differentiable in $\Delta_I,\Delta_{II}$ with derivatives
        \begin{align*}
            \frac{\partial}{\partial \Delta_I} \bar{m}^*_{\Delta_I,\Delta_{II}}
            &=
            -
            \frac{\frac{\partial}{\partial \Delta_I} F_{\Delta_I,\Delta_{II}}(\bar{m}^*_{\Delta_I,\Delta_{II}})}{\frac{\partial}{\partial \bar{m}} F_{\Delta_I,\Delta_{II}}(\bar{m}^*_{\Delta_I,\Delta_{II}})},
            &
            \frac{\partial}{\partial \Delta_{II}} \bar{m}^*_{\Delta_I,\Delta_{II}}
            &=
            -
            \frac{\frac{\partial}{\partial \Delta_{II}} F_{\Delta_I,\Delta_{II}}(\bar{m}^*_{\Delta_I,\Delta_{II}})}{\frac{\partial}{\partial \bar{m}} F_{\Delta_I,\Delta_{II}}(\bar{m}^*_{\Delta_I,\Delta_{II}})}.
        \end{align*}
        By continuity of the derivatives, the first one is negative and the second one is positive in a sufficiently small neighborhood of $\Delta_I = 0 = \Delta_{II}$.
    \end{proof}

    \begin{proof}[Proof of \autoref{prop:Thirdmonotonicity}]

        Using the simplified notation from above the proof of \autoref{prop:Agent-Threshold} and following the proof of \autoref{prop:Localmonotonicity},
        note that we can express the threshold-based policies by the agent and the principal as
        \begin{align*}
            R &=
            \begin{cases}
                \risky, & \tM \leq \bar{m}^{\low}, \\
                \neutral & \bar{m}^{\low} < \tM \leq \bar{m}^{\high}, \\
                \safe, & \tM > \bar{m}^{\high},
            \end{cases}
            &
            A &=
            \begin{cases}
                \risky, & \tH \leq \bar{h}^R, \\
                \safe, & \tH > \bar{h}^R.
            \end{cases}
        \end{align*}
        Given thresholds $\bar{m}^{\low},\bar{m}^{\high},\bar{h}^{\risky},\bar{h}^{\neutral},\bar{h}^{\safe}$,
        expected losses of principal and agent are
        \begin{align*}
            &L(\bar{m}^{\low},\bar{m}^{\high},\bar{h}^{\risky},\bar{h}^{\neutral},\bar{h}^{\safe})
            =
            \E[\ell(Y,A)]
            \\
            &=
            \E[\Ind{\tM {\leq} \bar{m}^{\low}, \tH {\leq} \bar{h}^{\risky}} \phi(\tH,\tM)] c_{II}
            + \E[\Ind{\tM {\leq} \bar{m}^{\low}, \tH {>} \bar{h}^{\risky}} (1{-}\phi(\tH,\tM))] c_I
            \\
            &\phantom{{}={}}
            + \E[\Ind{\bar{m}^{\low} {<} \tM {\leq} \bar{m}^{\high}, \tH {\leq} \bar{h}^{\neutral}} \phi(\tH,\tM)] c_{II}
            + \E[\Ind{\bar{m}^{\low} {<} \tM {\leq} \bar{m}^{\high}, \tH {>} \bar{h}^{\neutral}} (1{-}\phi(\tH,\tM))] c_I
            \\
            &\phantom{{}={}}
            + \E[\Ind{\tM {>} \bar{m}^{\high}, \tH {\leq} \bar{h}^{\safe}} \phi(\tH,\tM)] c_{II}
            + \E[\Ind{\tM {>} \bar{m}^{\high}, \tH {>} \bar{h}^{\safe}} (1{-}\phi(\tH,\tM))] c_I,
            \\
            &L^*(\bar{m}^{\low},\bar{m}^{\high},\bar{h}^{\risky},\bar{h}^{\neutral},\bar{h}^{\safe})
            =
            \E[\ell^*(Y,A,R)] = L(\bar{m}^{\low},\bar{m}^{\high},\bar{h}^{\risky},\bar{h}^{\neutral},\bar{h}^{\safe})
            \\
            &\phantom{{}={}}
            +  \E[\Ind{\tM {\leq} \bar{m}^{\low}, \tH {>} \bar{h}^{\risky}} (1{-}\phi(\tH,\tM))] \Delta_I + \E[\Ind{\tM {>} \bar{m}^{\high}, \tH {\leq} \bar{h}^{\safe}} \phi(\tH,\tM)] \Delta_{II}.
        \end{align*}
        The optimal agent thresholds
        $\bar{h}^{\risky}_{\Delta_I}(\bar{m}^{\low}), \bar{h}^{\neutral}(\bar{m}^{\low},\bar{m}^{\high}), \bar{h}^{\safe}_{\Delta_{II}}(\bar{m}^{\high})$
        now are determined uniquely by the first-order conditions
        \begin{align*}
            \E[\phi(\bar{h}^{\risky},\tM)|\tM \leq \bar{m}^{\low}] &= \frac{c_I {+} \Delta_I}{c_I {+} c_{II} {+} \Delta_I},
            \\
            \E[\phi(\bar{h}^{\neutral},\tM)|\bar{m}^{\low} < \tM \leq \bar{m}^{\high}] &= \frac{c_I}{c_I {+} c_{II}},
            \\
            \E[\phi(\bar{h}^{\safe},\tM)|\tM > \bar{m}^{\high}] &= \frac{c_I}{c_I {+} c_{II} {+} \Delta_{II}}
        \end{align*}
        with unique solutions $\bar{h}^{\risky}_{\Delta_I}(\bar{m}^{\low}) > \bar{h}^{\neutral}(\bar{m}^{\low},\bar{m}^{\high}) > \bar{h}^{\safe}_{\Delta_{II}}(\bar{m}^{\high})$ by monotonicity of $f$ and our regularity assumptions,
        which by the implicit function theorem are continuously differentiable in $\bar{m}$
        as in the proof of \autoref{prop:Localmonotonicity} with
        \begin{align*}
            \frac{\partial}{\partial \bar{m}^{\low}} \bar{h}^{\risky}_{\Delta_I}(\bar{m}^{\low}) & < 0,
            &
            \frac{\partial}{\partial \Delta_I} \bar{h}^{\risky}_{\Delta_I}(\bar{m}^{\low}) & > 0,
            \\
            \frac{\partial}{\partial \bar{m}^{\low}} \bar{h}^{\neutral}(\bar{m}^{\low},\bar{m}^{\high}) & < 0,
            &
            \frac{\partial}{\partial \bar{m}^{\high}} \bar{h}^{\neutral}(\bar{m}^{\low},\bar{m}^{\high}) & < 0,
            \\
            \frac{\partial}{\partial \bar{m}^{\high}} \bar{h}^{\safe}_{\Delta_{II}}(\bar{m}^{\high}) & < 0,
            &
            \frac{\partial}{\partial \Delta_{II}} \bar{h}^{\safe}_{\Delta_{II}}(\bar{m}^{\high}) & < 0.
        \end{align*}
        The optimal principal thresholds $\bar{m}^{\high *}_{\Delta_I,\Delta_{II}}, \bar{m}^{\high *}_{\Delta_I,\Delta_{II}}$ then minimize
        \[L(\bar{m}^{\low},\bar{m}^{\high}, \bar{h}^{\risky}_{\Delta_I}(\bar{m}^{\low}), \bar{h}^{\neutral}(\bar{m}^{\low},\bar{m}^{\high}), \bar{h}^{\safe}_{\Delta_{II}}(\bar{m}^{\high})).\]
        Using the same notation as in the proof of \autoref{prop:Localmonotonicity},
        we have that
        \begin{align*}
            \frac{\d L}{\d \bar{m}^{\low}}
            &=
            \frac{\partial L}{\partial \bar{m}^{\low}}
            +
            \frac{\partial \bar{h}^{\risky}}{\partial \bar{m}^{\low}}
            \frac{\partial L}{\partial \bar{h}^{\risky}}
            \\
            &=
            \frac{\partial L}{\partial \bar{m}^{\low}}
            +
            \Delta_I
            \frac{\partial \bar{h}^{\risky}}{\partial \bar{m}^{\low}}
            \frac{\partial}{\partial \bar{h}^{\risky}}
            \E[\Ind{\tM {\leq} \bar{m}^{\low}, \tH {>} \bar{h}^{\risky}_{\Delta_I}(\bar{m}^{\low})} (1{-}\phi(\tH,\tM))]
            \\
            &=
            \mu_M(\bar{m}^{\low}) \E[\Ind{\bar{h}^{\neutral}(\bar{m}^{\low},\bar{m}^{\high}) {<} \tH {\leq} \bar{h}^{\risky}_{\Delta_I}(\bar{m}^{\low})} ((c_I + c_{II}) \phi(\tH,\bar{m}^{\low}) - c_I)]
            \\
            &\phantom{{}={}}
            -
            \Delta_I \frac{\partial \bar{h}^{\risky}}{\partial \bar{m}^{\low}}
            \mu_H(\bar{h}^{\risky}_{\Delta_I}(\bar{m}^{\low})) \E[\Ind{\tM {\leq} \bar{m}^{\low}} (1{-}\phi(\bar{h}^{\risky}_{\Delta_I}(\bar{m}^{\low}),\tM))]
            =
            F^{\low}_{\Delta_I}(\bar{m}^{\low},\bar{m}^{\high}),
            \\
            \frac{\d L}{\d \bar{m}^{\high}}
            &=
            \frac{\partial L}{\partial \bar{m}^{\high}}
            +
            \frac{\partial \bar{h}^{\safe}}{\partial \bar{m}^{\high}}
            \frac{\partial L}{\partial \bar{h}^{\safe}}
            \\
            &=
            \frac{\partial L}{\partial \bar{m}^{\high}}
            +
            \Delta_{II}
            \frac{\partial \bar{h}^{\safe}}{\partial \bar{m}^{\high}}
            \frac{\partial}{\partial \bar{h}^{\safe}}
            \E[\Ind{\tM {>} \bar{m}^{\high}, \tH {\leq} \bar{h}^{\safe}_{\Delta_{II}}(\bar{m}^{\high})} (1{-}\phi(\tH,\tM))]
            \\
            &=
            \mu_M(\bar{m}^{\high}) \E[\Ind{\bar{h}^{\safe}_{\Delta_{II}}(\bar{m}^{\high}) {<} \tH {\leq} \bar{h}^{\neutral}(\bar{m}^{\low},\bar{m}^{\high})} ((c_I + c_{II}) \phi(\tH,\bar{m}^{\high}) - c_I)]
            \\
            &\phantom{{}={}}
            +
            \Delta_{II} \frac{\partial \bar{h}^{\safe}}{\partial \bar{m}^{\high}}
            \mu_H(\bar{h}^{\safe}_{\Delta_{II}}(\bar{m}^{\high})) \E[\Ind{\tM {>} \bar{m}^{\high}} \phi(\bar{h}^{\safe}_{\Delta_{II}}(\bar{m}^{\high}),\tM)]
            =
            F^{\high}_{\Delta_{II}}(\bar{m}^{\low},\bar{m}^{\high}),
        \end{align*}
        where $(F^{\low}_{\Delta_I}(\bar{m}^{\low},\bar{m}^{\high}),F^{\high}_{\Delta_{II}}(\bar{m}^{\low},\bar{m}^{\high}))$ is continuously differentiable in $\bar{m}^{\low},\bar{m}^{\high},\Delta_I,\Delta_{II}$
        with
        \begin{align*}
            \frac{\partial}{\partial \Delta_I} F^{\low}_{0}(\bar{m}^{\low},\bar{m}^{\high})
            &=
            \overbrace{\mu_M(\bar{m}^{\low}) \mu_H(\bar{h}^{\risky}(\bar{m}^{\low}))}^{>0} \overbrace{((c_I + c_{II}) \phi(\bar{h}^{\risky}_{\Delta_I}(\bar{m}^{\low}),\bar{m}^{\low}) - c_I)}^{>0}
            \overbrace{\frac{\partial \bar{h}^{\risky}}{\partial \Delta_I}}^{>0}
            \\
            &\phantom{{}={}}
            -
            \underbrace{\frac{\partial \bar{h}^{\risky}}{\partial \bar{m}^{\low}}}_{<0}
            \underbrace{\mu_H(\bar{h}^{\risky}_{\Delta_I}(\bar{m}^{\low})) \E[\Ind{\tM {\leq} \bar{m}^{\low}} (1{-}\phi(\bar{h}^{\risky}_{\Delta_I}(\bar{m}^{\low}),\tM))]}_{>0} > 0,
            \\
            \frac{\partial}{\partial \Delta_{II}} F^{\high}_{0}(\bar{m}^{\low},\bar{m}^{\high})
            &=
            - \overbrace{\mu_M(\bar{m}^{\high}) \mu_H(\bar{h}^{\safe}(\bar{m}^{\high}))}^{>0} \overbrace{((c_I + c_{II}) \phi(\bar{h}^{\safe}_{\Delta_{II}}(\bar{m}^{\high}),\bar{m}^{\high}) - c_I)}^{<0}
            \overbrace{\frac{\partial \bar{h}^{\safe}}{\partial \Delta_{II}}}^{<0}
            \\
            &\phantom{{}={}}
            +
            \underbrace{\frac{\partial \bar{h}^{\safe}}{\partial \bar{m}^{\high}}}_{<0}
            \underbrace{\mu_H(\bar{h}^{\safe}_{\Delta_{II}}(\bar{m}^{\high})) \E[\Ind{\tM {>} \bar{m}^{\high}} \phi(\bar{h}^{\safe}_{\Delta_{II}}(\bar{m}^{\high}),\tM)]}_{>0} < 0,
        \end{align*}
        As in the proof of \autoref{prop:Localmonotonicity}, the result follows from the implicit function theorom,
        where we note that
        \begin{align*}
            &\begin{psmallmatrix}
                \frac{\partial}{\partial \Delta_I} \bar{m}^{\low *}_{0,0} & \frac{\partial}{\partial \Delta_{II}} \bar{m}^{\low *}_{0,0} \\
                \frac{\partial}{\partial \Delta_I} \bar{m}^{\high *}_{0,0} & \frac{\partial}{\partial \Delta_{II}} \bar{m}^{\high *}_{0,0}
            \end{psmallmatrix}
            \\
            &=
            - \left.\underbrace{\begin{psmallmatrix}
                \frac{\partial}{\partial \bar{m}^{\low}} F^{\low}_{0}(\bar{m}^{\low *}_{0,0},\bar{m}^{\high *}_{0,0})
                &
                \frac{\partial}{\partial \bar{m}^{\high}} F^{\low}_{0}(\bar{m}^{\low *}_{0,0},\bar{m}^{\high *}_{0,0})
                \\
                \frac{\partial}{\partial \bar{m}^{\low}} F^{\high}_{0}(\bar{m}^{\low *}_{0,0},\bar{m}^{\high *}_{0,0})
                &
                \frac{\partial}{\partial \bar{m}^{\high}} F^{\high}_{0}(\bar{m}^{\low *}_{0,0},\bar{m}^{\high *}_{0,0})
            \end{psmallmatrix}}_{
                = \left. \frac{\partial^2 \E[\ell(Y,A)]}{\partial(\bar{m}^{\low},\bar{m}^{\high})' \partial(\bar{m}^{\low},\bar{m}^{\high})} \right|_{\substack{\bar{m}^{\low} = \bar{m}^{\low *}_{0,0}, \bar{m}^{\high} = \bar{m}^{\high *}_{0,0} \\ \Delta_I = 0 = \Delta_{II}}} \succ \mathbbm{O}}
                \right.^{-1}
            \begin{psmallmatrix}
                \frac{\partial}{\partial \Delta_I} F^{\low}_{0}(\bar{m}^{\low *}_{0,0},\bar{m}^{\high *}_{0,0}) & 0 \\
                0 & \frac{\partial}{\partial \Delta_{II}} F^{\high}_{0}(\bar{m}^{\low *}_{0,0},\bar{m}^{\high *}_{0,0})
            \end{psmallmatrix},
        \end{align*}
        which implies $\frac{\partial}{\partial \Delta_I} \bar{m}^{\low *}_{0,0} < 0, \frac{\partial}{\partial \Delta_{II}} \bar{m}^{\high *}_{0,0} > 0$ and extends to a neighborhood of $\Delta_I = 0 = \Delta_{II}$.
    \end{proof}

        \begin{proof}[Proof of \autoref{prop:Triageminimaxgeneral}]
         To show that this solution is minimax optimal, we show first that deploying the recommendation algorithm $\overline{\rec}^*$ guarantees a maximal expected loss $L(\overline{\P})$ across all $\P \in \mathcal{P}(\overline{\P})$ and all $\Delta_I,\Delta_{II}$. Second, we then show that there is a specific distribution $\P^* \in \mathcal{P}(\overline{\P})$ and a specific pair $\Delta^*_I,\Delta^*_{II}$ such that no recommendation algorithm can improve over $L(\overline{\P})$.

        \emph{For the first step}, for any recommendation policy $\rec: \mathcal{X}_m \rightarrow \mathcal{R} = \{\risky,\neutral,\safe\}$ consider the two decisions $\overline{A}, \overline{A}^\dagger$ given by
        \begin{align*}
            \overline{A} &= 
            \begin{cases}
                A_0, & \rec(X_m) {=} \neutral, \\
                \rec(X_m), & \text{otherwise},
            \end{cases}
            &
            \overline{A}^\dagger &= 
            \begin{cases}
                A_0^\dagger, & \rec(X_m) {=} \neutral, \\
                \rec(X_m), & \text{otherwise}.
            \end{cases}
        \end{align*}
        Given $\rec$, both $\overline{A}$ and $\overline{A}^\dagger$ are feasible choices that the agent could take (since the agent observes $R = \rec(X_m)$ and $A_0$ is $X_h$-measurable), so for their actual choice $A$ we must have that
        \begin{align*}
            \E[\ell^*(Y,A,R)] \leq \min\{\E[\ell^*(Y,\overline{A},R)], \E[\ell^*(Y,\overline{A}^\dagger,R)]\}.
        \end{align*}
        At the same time, $\ell^*(Y,\overline{A},R) = \ell(Y,\overline{A})$ and $\ell^*(Y,\overline{A}^\dagger,R) = \ell(Y,\overline{A}^\dagger)$ since the choices $\overline{A}, \overline{A}^\dagger$ never deviate from a recommended action and therefore do not incur any additional losses.
        Also, $\ell(Y,A) \leq \ell^*(Y,A,R)$.
        Finally, $\E[\ell(Y,\overline{A})] = \overline{\E}[\ell(Y,\overline{A})], \E[\ell(Y,\overline{A}^\dagger)] = \overline{\E}[\ell(Y,\overline{A}^\dagger)]$ since $(Y,X_m,A_0)$ are observed under $\overline{\P}$.
        It follows that
        \begin{align*}
            \E[\ell(Y,A)]
            \leq
            \E[\ell^*(Y,A,R)] \leq \min\{\E[\ell^*(Y,\overline{A},R)], \E[\ell^*(Y,\overline{A}^\dagger,R)]\}
            =
            \min\{\overline{\E}[\ell(Y,\overline{A})], \overline{\E}[\ell(Y,\overline{A}^\dagger)]\}.
        \end{align*}
        
        We now apply this inequality with the recommendation $\overline{\rec}^*$ in the proposition.
        Assume first that $\overline{\rec}^* = \overline{\rec}$.
        In this case, the inequality yields (using the first term)
        \begin{align}
        \label{eqn:Simpleupper}
            \E[\ell(Y,A)]
            \leq \overline{\E}[\Ind{\overline{\rec}(X_m) {\neq} \neutral} \ell(Y,\overline{\rec}(X_m)) + \Ind{\overline{\rec}(X_m) {=} \neutral} \ell(Y,A_0)].
        \end{align}
        Assume instead that $\overline{\rec}^* = \overline{\rec}^\dagger$.
        Then the inequality implies (using the second term)
        \begin{align*}
            \E[\ell(Y,A)]
            \leq \overline{\E}[\Ind{\overline{\rec}^\dagger(X_m) {\neq} \neutral} \ell(Y,\overline{\rec}^\dagger(X_m)) + \Ind{\overline{\rec}^\dagger(X_m) {=} \neutral} \ell(Y,A^\dagger_0)].
        \end{align*}
        By definition of $\overline{\rec}^*$ as the choice that minimizes over these two upper bounds, we directly find that
        \begin{equation}
        \label{eqn:Fullupper}
        \begin{aligned}
            \E[\ell(Y,A)]
            \leq L(\overline{\P}) = \min\{
            &\overline{\E}[\Ind{\overline{\rec}(X_m) {\neq} \neutral} \ell(Y,\overline{\rec}(X_m)) + \Ind{\overline{\rec}(X_m) {=} \neutral} \ell(Y,A_0)],
            \\
            &\overline{\E}[\Ind{\overline{\rec}^\dagger(X_m) {\neq} \neutral} \ell(Y,\overline{\rec}^\dagger(X_m)) + \Ind{\overline{\rec}^\dagger(X_m) {=} \neutral} \ell(Y,A^\dagger_0)]\}.
        \end{aligned}
        \end{equation}        
         
        \emph{For the second step}, we now have to show that we cannot generally improve over the bound $L(\overline{\P})$.
        To this end, consider the distribution $\P^*$ over $(Y,X_m,X_h)$ with $X_h = A_0$, which is fully pinned down by $\overline{\P}$ since there is no additional unobserved private information of the agent.

        Consider any fixed recommendation policy $\rec: \mathcal{X}_m \rightarrow \mathcal{R}$.
        Since $\P^*(\P^*(Y{=}\bad|X_m,X_h) \in (0,1)) = 1$ by assumption,
        we also have $\P^*(\P^*(Y{=}\bad|\rec(X_m),X_h) \in (0,1)) = 1$.
        As a result,
        \begin{align*}
            &\lim_{\Delta_{II} \rightarrow \infty}
            \E^*[\ell(Y,A) \Ind{\rec(X_m)=\safe}]
            \\
            &=
            \lim_{\Delta_{II} \rightarrow \infty}
            \E^*[\ell(Y,\risky) \Ind{\rec(X_m){=}\safe, \P^*(Y{=}\bad|\rec(X_m){=}\safe,X_h) \leq \smash{\overbrace{p^\safe {=} c_I / (c_I {+} c_{II} {+} \Delta_{II})}^{\rightarrow 0}}}]
            \\
            &\phantom{=\lim_{\Delta_{II} \rightarrow \infty}}
            + \E^*[\ell(Y,\safe) \Ind{\rec(X_m){=}\safe, \P^*(Y{=}\bad|\rec(X_m){=}\safe,X_h) > p^\safe {=} c_I / (c_I {+} c_{II} {+} \Delta_{II})}]
            \\
            &=
            \E^*[\ell(Y,\safe) \Ind{\rec(X_m){=}\safe}],
            \\
            &\lim_{\Delta_{I} \rightarrow \infty}
            \E^*[\ell(Y,A) \Ind{\rec(X_m)=\risky}]
            \\
            &=
            \lim_{\Delta_I \rightarrow \infty}
            \E^*[\ell(Y,\risky) \Ind{\rec(X_m){=}\risky, \P^*(Y{=}\bad|\rec(X_m){=}\risky,X_h) \leq \smash{\overbrace{p^\risky {=} (c_I {+} \Delta_I) / (c_I {+} c_{II} {+} \Delta_{I})}^{\rightarrow 1}}}]
            \\
            &\phantom{=\lim_{\Delta_I \rightarrow \infty}}
            + \E^*[\ell(Y,\safe) \Ind{\rec(X_m){=}\risky, \P^*(Y{=}\bad|\rec(X_m){=}\risky,X_h) > p^\risky {=} (c_I {+} \Delta_I) / (c_I {+} c_{II} {+} \Delta_{I})}]
            \\
            &=
            \E^*[\ell(Y,\risky) \Ind{\rec(X_m){=}\risky}]
        \end{align*}
        by monotone convergence.
        Hence,
        \begin{align*}
            \sup_{\Delta_I,\Delta_{II} \geq 0}
            \E^*[\ell(Y,A)]
            \geq
            \E^*[\ell(Y,\rec(X_m)) \Ind{\rec(X_m){\neq}\neutral}]
            +
            \min_{a:\mathcal{X}_h \rightarrow \{ \risky,\safe\}}
            \E^*[\ell(Y,a(X_h)) \Ind{\rec(X_m){=}\neutral}].
        \end{align*}
        As a result, the best worst-case loss that can be achieved for the distribution $\P^*$ fulfills
        \begin{align*}
            \min_{\rec: \mathcal{X}_h \rightarrow \mathcal{R}} \sup_{\Delta_I,\Delta_{II} \geq 0}
            \E^*[\ell(Y,A)]
            \geq
            \min_{\substack{
                \rec: \mathcal{X}_h \rightarrow \mathcal{R} \\
                \mathclap{a: \mathcal{X}_h \rightarrow \{ \risky,\safe\}}}
            }
            \E^*[\ell(Y,\rec(X_m)) \Ind{\rec(X_m){\neq}\neutral} +
            \ell(Y,a(X_h)) \Ind{\rec(X_m){=}\neutral}],
        \end{align*}
        where the order in which we choose the recommendation policy $\rec$ and the optimal decision policy $a$ when the neutral recommendation is given does not change the value of the minimum.

        Since $X_h = A_0$, we can assume that $\mathcal{X}_h = \{\risky, \safe\}$, so there are only four choices for $a$: the identity ($a(x_h) = x_h$, meaning we adopt the baseline decision), its complement ($a(x_h) = x_h^\dagger$, meaning we take the opposite of the baseline decision),
        and the two constant decisions ($a(x_h) = \risky$ or $a(x_h) = \safe$).
        When jointly choosing optimal recommendations and actions, we can ignore the constant decisions since in those two cases, we could simply recommend the risky or safe action, respectively, rather than giving a neutral recommendation.
        Hence, we obtain the bound
        \begin{equation}
        \label{eqn:lowerbound}
        \begin{aligned}
            &\min_{\rec: \mathcal{X}_h \rightarrow \mathcal{R}} \sup_{\Delta_I,\Delta_{II} \geq 0}
            \E^*[\ell(Y,A)]
            \\
            &\geq
            \min\{
                \min_{\rec: \mathcal{X}_h \rightarrow \mathcal{R}} \overline{\E}[\ell(Y,\rec(X_m)) \Ind{\rec(X_m){\neq}\neutral}
                + \ell(Y,A_0) \Ind{\rec(X_m){=}\neutral}],
            \\
                &\phantom{\geq\min\{} \min_{\rec: \mathcal{X}_h \rightarrow \mathcal{R}} \overline{\E}[\ell(Y,\rec(X_m)) \Ind{\rec(X_m){\neq}\neutral}
                +\ell(Y,A^\dagger_0) \Ind{\rec(X_m){=}\neutral}]\},
            \end{aligned}
        \end{equation}
        where we have used that $\overline{\P}$ pins down the distribution $\P^*$ by construction.
        
        To close the gap to the bound $L(\overline{\P})$ from \eqref{eqn:Fullupper}, it remains to show that $\overline{\rec}$ and $\overline{\rec}^\dagger$ minimize the respective expected losses.
        To this end, note that for $x_m \in \mathcal{X}_m$ and $r \in \mathcal{R}$
        \begin{equation}
        \label{eqn:pointwisemin}
            \overline{\E}[\ell(Y,r) \Ind{r{\neq}\neutral}
            +
            \ell(Y,A_0) \Ind{r{=}\neutral}|X_m{=}x_m]
            =
            \begin{cases}
                \overline{\E}[\ell(Y,r)|X_m{=}x_m], & r{\neq}\neutral, \\
                \overline{\E}[\ell(Y,A_0)|X_m{=}x_m], & r {=} \neutral.
            \end{cases}
        \end{equation}
        Since $\overline{\rec}_m(x_m)$ minimizes $\overline{\E}[\ell(Y,r)|X_m{=}x_m]$ over $r \in \{\risky, \safe\}$,
        a minimizer of \eqref{eqn:pointwisemin} is
        \begin{align*}
            \begin{cases}
                \overline{\rec}_m(x_m), & \overline{\E}[\ell(Y,\overline{\rec}_m(x_m))|X_m{=}x_m] < \overline{\E}[\ell(Y,A_0)|X_m{=}x_m], \\
                \neutral, & \text{otherwise}
            \end{cases}
            = \overline{\rec}(x_m).
        \end{align*}
        As a result,
        \begin{align*}
            &\min_{\rec: \mathcal{X}_h \rightarrow \mathcal{R}} \overline{\E}[\ell(Y,\rec(X_m)) \Ind{\rec(X_m){\neq}\neutral}
                + \ell(Y,A_0) \Ind{\rec(X_m){=}\neutral}]
            \\
            &=
            \overline{\E}[\ell(Y,\overline{\rec}(X_m)) \Ind{\overline{\rec}(X_m){\neq}\neutral}
                + \ell(Y,A_0) \Ind{\overline{\rec}(X_m){=}\neutral}],
        \end{align*}
        and by the same argument also
        \begin{align*}
            &\min_{\rec: \mathcal{X}_h \rightarrow \mathcal{R}} \overline{\E}[\ell(Y,\rec(X_m)) \Ind{\rec(X_m){\neq}\neutral}
                + \ell(Y,A^\dagger_0) \Ind{\rec(X_m){=}\neutral}]
            \\
            &=
            \overline{\E}[\ell(Y,\overline{\rec}^\dagger(X_m)) \Ind{\overline{\rec}^\dagger(X_m){\neq}\neutral}
                + \ell(Y,A_0) \Ind{\overline{\rec}^\dagger(X_m){=}\neutral}].
        \end{align*}
    Plugging into \eqref{eqn:lowerbound} and combining with \eqref{eqn:Fullupper},
    we obtain that there exists a distribution $\P^* \in \mathcal{P}(\overline{\P})$ such that
    \begin{align*}
        \min_{f: \mathcal{X}_m \rightarrow \mathcal{R}}
        \sup_{
        \substack{
            \P \in \mathcal{P}(\overline{\P}) \\
            \Delta_I,\Delta_{II} \geq 0
        }} \E[\ell(Y,A)]
        \leq
        L(\overline{\P})
        \leq
        \min_{f: \mathcal{X}_m \rightarrow \mathcal{R}}
        \sup_{\Delta_I,\Delta_{II} \geq 0}
        \E^*[\ell(Y,A)].
    \end{align*}
    Hence, $L(\overline{\P})$ is the minimax expected loss, and it can be achieved by deploying the recommendation algorithm from the proposition.
    \end{proof}
    
    \begin{proof}[Proof of \autoref{prop:Generalfeasiblecomplementarity}]
        By \eqref{eqn:Simpleupper} in the proof of \autoref{prop:Triageminimaxgeneral},
        the recommendation algorithm $\overline{\rec}$ guarantees that
        \begin{align*}
            \E[\ell(Y,\overline{A})]
            \leq
            \overline{\E}[\Ind{\overline{\rec}(X_m) {\neq} \neutral} \ell(Y,\overline{\rec}(X_m)) + \Ind{\overline{\rec}(X_m) {=} \neutral} \ell(Y,A_0)].
        \end{align*}
        By \eqref{eqn:Fullupper}, the same holds for the recommendation algorithm $\overline{\rec}^*$, so
        \begin{align*}
            \max\{\E[\ell(Y,\overline{A})],\E[\ell(Y,\overline{A}^\dagger)]\}
            \leq
            \overline{\E}[\Ind{\overline{\rec}(X_m) {\neq} \neutral} \ell(Y,\overline{\rec}(X_m)) + \Ind{\overline{\rec}(X_m) {=} \neutral} \ell(Y,A_0)].
        \end{align*}
        In addition, the definition of $\overline{\rec}$ from \eqref{eqn:Triage} guarantees that
        \begin{align*}
            &\overline{\E}[\Ind{\overline{\rec}(X_m) {\neq} \neutral} \ell(Y,\overline{\rec}(X_m)) + \Ind{\overline{\rec}(X_m) {=} \neutral} \ell(Y,A_0)]
            \\
            &=
            \overline{\E}[\Ind{\overline{\rec}(X_m) {\neq} \neutral} \underbrace{\overline{\E}[\ell(Y,\overline{\rec}(X_m))|X_m]}_{\leq \overline{\E}[\ell(Y,A_0)|X_m]} + \Ind{\overline{\rec}(X_m) {=} \neutral} \underbrace{\overline{\E}[\ell(Y,A_0)|X_m]}_{\leq \overline{\E}[\ell(Y,\overline{\rec}(X_m))|X_m]}]
            \\
            &\leq
            \max\{
                \overline{\E}[\ell(Y,A_0)],
                \overline{\E}[\ell(Y,\overline{\rec}(X_m))]
            \}.
        \end{align*}
        Finally, note that $\overline{\E}[\ell(Y,A_0)] = \E[\ell(Y,A_0)]$ and $\overline{\E}[\ell(Y,\overline{\rec}(X_m))] = \E[\ell(Y,A_0)]$ for all $\P \in \mathcal{P}(\overline{\P})$ since $\overline{\P}$ pins down the distribution of $(Y,X_m,A_0)$.
    \end{proof}

    \begin{proof}[Proof of \autoref{prop:Onesidedminimaxgeneral}]
        In the proof of \autoref{prop:Unobservedminimax}, the worst-case distribution $\P^*$ still yields the same first-best expected loss. Hence, the minimax expected loss bound must be the same, and it can be guaranteed by any algorithm that allows the agent to recover whether $\overline{\P}'(Y'{=}\bad|X_m,A_0{=}\risky) \leq p^*$ and recommends $\safe$ only for instances where $\overline{\P}'(Y'{=}\bad|X_m,A_0{=}\risky) \leq p^*$, such as those in \eqref{eqn:Onesidedtriagegeneral}, by the same bound as in the first part of the proof of \autoref{prop:Unobservedminimax}.
    \end{proof}

\end{document}